\documentclass{article}

\usepackage{arxiv}

\usepackage{researchpack}
\usepackage{subcaption}
\usepackage{natbib}
\usepackage{hyperref}
\usepackage[capitalise,nameinlink]{cleveref}
\usepackage{xspace}
\usepackage{enumitem}
\usepackage{annotate-equations}
\usepackage{rotating}
\usepackage{threeparttablex}
\usepackage{tcolorbox}
\usepackage{pifont}
\usepackage{wrapfig}
\usepackage{tikz}
\usepackage{enumitem}
\usepackage{array}
\usepackage{centernot}
\usepackage{adjustbox}

\usetikzlibrary{positioning,shapes,arrows,calc}

\declaretheorem[name=Theorem]{theorem}
\numberwithin{theorem}{section}
\declaretheorem[name=Lemma, numberlike=theorem]{lemma}

\declaretheorem[name=Corollary, numberlike=theorem]{corollary}
\declaretheorem[name=Remark, numberlike=theorem]{remark}
\declaretheorem[name=Definition, numberlike=theorem]{definition}
\declaretheorem[name=Example, numberlike=theorem]{example}
\declaretheorem[name=Assumption, numberlike=theorem]{assumption}

\crefname{assumption}{Assumption}{Assumptions}
\Crefname{assumption}{Assumption}{Assumptions}

\newcommand{\BK}{\ensuremath{\mathsf{K}}\xspace}

\newcommand{\Vset}[1]{\ensuremath{\mathrm{Vert}(#1)}\xspace}
\newcommand{\Collapse}{\ensuremath{\mathrm{Cls}}\xspace}
\newcommand{\supp}{\ensuremath{\mathrm{supp}}\xspace}

\newcommand{\BOIA}{{\tt BDD-OIA}\xspace}
\newcommand{\Clevr}{{\tt Clevr}\xspace}
\newcommand{\MNIST}{{\tt MNIST}\xspace}
\newcommand{\MNISTAdd}{{\tt MNIST-Add}\xspace}
\newcommand{\MNISTMax}{{\tt MNIST-Max}\xspace}

\newcommand{\MNISTSumParity}{{\tt MNIST-SumParity}\xspace}

\newcommand{\MNISTXOR}{{\tt MNIST-XOR}\xspace}
\newcommand{\Kandinsky}{{\tt Kandinsky}\xspace}

\newcommand{\PNSP}{\texttt{PNSPs}\xspace}
\newcommand{\DPL}{\texttt{DPL}\xspace}
\newcommand{\SL}{\texttt{SL}\xspace}
\newcommand{\LTN}{\texttt{LTN}\xspace}

\newcommand{\ABL}{\texttt{ABL}\xspace}
\newcommand{\bears}{\texttt{bears}\xspace}
\newcommand{\nesydm}{\texttt{NeSyDM}\xspace}
\newcommand{\rsbench}{\texttt{rsbench}\xspace}
\newcommand{\countrss}{\texttt{countrss}\xspace}

\newcommand{\MZero}{\raisebox{-1pt}{\includegraphics[width=1.85ex]{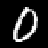}}\xspace}
\newcommand{\MOne}{\raisebox{-1pt}{\includegraphics[width=1.85ex]{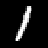}}\xspace}
\newcommand{\MTwo}{\raisebox{-1pt}{\includegraphics[width=1.85ex]{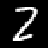}}\xspace}
\newcommand{\MThree}{\raisebox{-1pt}{\includegraphics[width=1.85ex]{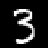}}\xspace}
\newcommand{\MFour}{\raisebox{-1pt}{\includegraphics[width=1.85ex]{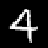}}\xspace}
\newcommand{\MFive}{\raisebox{-1pt}{\includegraphics[width=1.85ex]{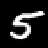}}\xspace}
\newcommand{\MSix}{\raisebox{-1pt}{\includegraphics[width=1.85ex]{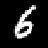}}\xspace}
\newcommand{\MSeven}{\raisebox{-1pt}{\includegraphics[width=1.85ex]{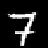}}\xspace}
\newcommand{\MEight}{\raisebox{-1pt}{\includegraphics[width=1.85ex]{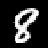}}\xspace}
\newcommand{\MNine}{\raisebox{-1pt}{\includegraphics[width=1.85ex]{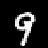}}\xspace}

\newcommand{\changed}[1]{#1}
\newcommand{\rechanged}[1]{#1}

\newcommand{\cmark}{\ding{51}}
\newcommand{\xmark}{\ding{55}}
\newcommand{\CMARK}{\textcolor{teal}{\cmark}\xspace}
\newcommand{\XMARK}{\textcolor{WildStrawberry}{\xmark}\xspace}
\newcommand{\QMARK}{\textbf{?}\xspace}

\hypersetup{
    colorlinks=true,
    citecolor=teal,
    linkcolor=violet,
    urlcolor=teal,
    pdftitle={Symbol Grounding in Neuro-Symbolic AI: A Gentle Introduction to Reasoning Shortcuts},
    pdfauthor={Emanuele Marconato, Samuele Bortolotti, Emile van Krieken, Paolo Morettin, Elena Umili, Antonio Vergari, Efthymia Tsamoura, Andrea Passerini, Stefano Teso},
    pdfkeywords={Neuro-Symbolic AI, Shortcuts, Reasoning Shortcuts, Symbol Grounding},
}

\newcommand*{\colorboxed}{}
\def\colorboxed#1#{%
  \colorboxedAux{#1}%
}
\newcommand*{\colorboxedAux}[3]{%
  \begingroup
    \colorlet{cb@saved}{.}%
    \color#1{#2}%
    \boxed{%
      \color{cb@saved}%
      #3%
    }%
  \endgroup
}

\begin{document}

\title{Symbol Grounding in Neuro-Symbolic AI:\\ A Gentle Introduction to Reasoning Shortcuts}

\author{
    Emanuele Marconato\thanks{
    $^*$ Corresponding author. $\dagger$ Equal contribution.}$^\dagger$\\
    University of Trento\\Italy\\
    \texttt{name.surname@unitn.it}\\
    \And
    Samuele Bortolotti$^\dagger$\\
    University of Trento\\Italy\\
    \texttt{name.surname@unitn.it}\\
    \And
    Emile van Krieken$^\dagger$\\
    Vrije Universiteit Amsterdam\\The Netherlands\\
    \texttt{e.van.krieken@vu.nl}\\
    \AND
    Paolo Morettin\\
    University of Trento\\Italy\\
    \texttt{name.surname@unitn.it}\\
    \And
    Elena Umili\\
    Sapienza University of Rome\\Italy\\
    \texttt{umili@diag.uniroma1.it}\\
    \And
    Antonio Vergari\\
    University of Edinburgh\\United Kingdom\\
    \texttt{avergari@ed.ac.uk}\\
    \AND
    Efthymia Tsamoura\\
    Huawei Labs\\Cambridge United Kingdom\\
    \texttt{efthymia.tsamoura@gmail.com}\\
    \And
    Andrea Passerini\\
    University of Trento\\Italy\\
    \texttt{name.surname@unitn.it}\\
    \And
    Stefano Teso\\
    University of Trento\\Italy\\
    \texttt{name.surname@unitn.it}\\
}

\setshorttitle{Symbol Grounding in Neuro-Symbolic AI}
\setshortauthors{Marconato, Bortolotti, van Krieken et al.}

\maketitle

\begin{abstract}
    Neuro-symbolic (NeSy) AI aims to develop deep neural networks whose predictions comply with \textit{prior knowledge} encoding, \eg safety or structural constraints.  As such, it represents one of the most promising avenues for reliable and trustworthy AI.
    The core idea behind NeSy AI is to combine neural and symbolic steps: neural networks are typically responsible for mapping low-level inputs into high-level symbolic concepts, while symbolic reasoning infers predictions compatible with the extracted concepts and the prior knowledge.
    Despite their promise, it was recently shown that -- whenever the concepts are not supervised directly -- NeSy models can be affected by \textit{\textbf{Reasoning Shortcuts (RSs)}}. That is, they can achieve high label accuracy by grounding the concepts incorrectly.
    RSs can compromise the interpretability of the model's explanations, performance in out-of-distribution scenarios, and therefore reliability. 
    At the same time, RSs are difficult to detect and prevent unless concept supervision is available, which is typically not the case.
    However, the literature on RSs is scattered, making it difficult for researchers and practitioners to understand and tackle this challenging problem.
    This overview addresses this issue by providing a gentle introduction to RSs, discussing their causes and consequences in intuitive terms. 
    It also reviews and elucidates existing theoretical characterizations of this phenomenon.  Finally, it details methods for dealing with RSs, including mitigation and awareness strategies, and maps their benefits and limitations.
    By reformulating advanced material in a digestible form, this overview aims to provide a unifying perspective on RSs to lower the bar to entry for tackling them.
    Ultimately, we hope this overview contributes to the development of reliable NeSy and trustworthy AI models.
\end{abstract}

\maketitle

\tableofcontents
\newpage

\section{Introduction}

In cognitive science and psychology, \textit{symbols}—or more generally, \textit{concepts}—are the compositional building blocks of human thought \citep{mandler2004foundations, spelke2007core}. 
They enable the abstraction and structuring of perception, supporting high-level cognitive functions \citep{whitehead1927symbolism, johnson1994mental} such as language, reasoning, and planning.
In recent years, there has been growing interest in using concepts in neural networks \citep{smolensky1987analysis, sun1992variable,  greff2020binding}, which bring not only improved generalization but also greater interpretability to their decision-making processes.
While the emergence of abstract concepts in the human brain is well-documented \citep{mandler2004foundations}, how---and even whether---such entities arise in neural networks remains fundamentally unclear \citep{jobengio2017measuring, lake2018generalization, park2024linear}.

Neuro-symbolic (NeSy) models aim to bridge this gap by integrating neural learning with symbolic reasoning \citep{harnard1990grounding, mcmillan1991rule, sun1992variable}, thereby enabling explicit manipulation of concepts within the inference process. %
Some approaches embed prior symbolic structure---such as logical rules \citep{de2021statistical}---while others attempt to learn such structure through specialized, trainable components \citep{lake2018generalization, ellis2021dreamcoder}.
A key promise of this integration is \textbf{\textit{reliability}} and \textbf{\textit{trustworthiness}}: by making all decisions traceable to interpretable, well-defined concepts, decisions become more transparent \citep{kambhampati2022symbols}. Moreover, thanks to the modularity of symbolic components, learned concepts can be reused in novel scenarios without significant performance degradation.
Yet, the effectiveness of this integration hinges on solving a core challenge in NeSy AI: the problem of \textit{\textbf{symbol grounding}} \citep{harnard1990grounding}---that is, how to connect low-level perceptual data with high-level, abstract concepts.

\begin{figure}[!t]
    \centering
    \includegraphics[width=0.975\linewidth]{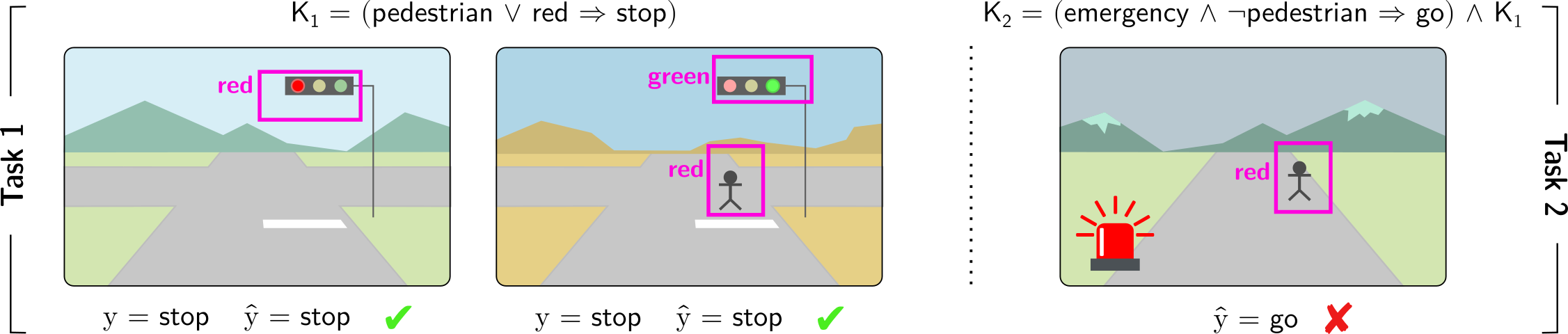}
    \caption{\textbf{Reasoning shortcuts are failures of correct symbol grounding}. Consider an autonomous driving car that must drive in compliance with traffic rules. 
    (Left) The car learns to make \textit{\textbf{correct predictions}} by leveraging background knowledge $\BK_1$ in Task $1$, \ie \textit{``if there is a pedestrian or a red light the car has to stop''}. However, in the process, it may incorrectly associate the concept of \texttt{red\_light} with the presence of either \textit{pedestrians} or \textit{red lights} in the dashcam, resulting in what is known as a \textbf{\textit{reasoning shortcut}}.
    (Right)
    Consequently, the learned concepts can lead to incorrect (and potentially catastrophic) predictions when the background knowledge changes. For example, $\BK_2$ in Task $2$ introduces an exception in the presence of an emergency situation, \ie ``if there is an emergency and no pedestrians are on the street, the car may proceed''.\protect\footnotemark}
    \label{fig:rss-main-fig}
\end{figure}

\begin{tcolorbox}[
    colback=gray!2!white,
    colframe=gray,
    arc=0pt,
    outer arc=0pt
]
\begin{example}
    Consider a model trained on autonomous driving scenarios, where the car must decide whether to \texttt{stop} or \texttt{go} based on the visual input, as per \cref{fig:rss-main-fig}. The model leverages prior knowledge $\BK_1$ which imposes that whenever \texttt{pedestrian} or \texttt{red\_light} are detected, the car must \texttt{stop}.
    Attaining accurate action predictions requires assigning valid binary values to the concepts \texttt{pedestrian} and \texttt{red\_light} (among others) from the visual input. 
    Therefore, concepts are grounded by the model through learning to give correct driving predictions. 
    Because \texttt{pedestrian} and \texttt{red\_light} are treated symmetrically by the knowledge (due to the disjunction)\footnote{We will precisely define the symmetries of the knowledge in later sections.} the self-driving car may confuse the two while still achieving correct predictions (\eg in \cref{fig:rss-main-fig} (Left), \texttt{red\_light} fires when either ``pedestrians'' or ``red lights'' are detected).
    \label{ex:sym-grounding}
\end{example}
\end{tcolorbox}

This example highlights a fundamental and underappreciated incompatibility between how NeSy models are trained and what is expected of them: on the one hand, the symbolic knowledge explicitly assigns \textit{names} (like {\tt pedestrian} and {\tt red\_light}) to the symbols with the expectation that these reflect their meaning, on the other, the model might assign them completely different semantics. \rechanged{In our example,} training the autonomous car to give correct predictions does not guarantee that concepts activate when they should\rechanged{: provided other objects apart from red lights can be perceived by the sensors, the autonomous car's \texttt{red\_light} may not solely contain information about the presence of ``red lights'' in the raw sensory data but also information about the other objects, if the knowledge is ambiguous enough to allow so.}

\cref{ex:sym-grounding} illustrates the central challenge of symbol grounding {in Neuro-Symbolic AI}: ensuring that abstract {predicted} concepts inside the model maintain a consistent and semantically correct link to the real-world entities they are meant to represent.
Unfortunately, in many learning contexts NeSy models---even when they perform nearly optimally on a task---may fail to assign the right meaning to learned concepts, leading to concepts with \textit{unintended semantics} \citep{marconato2023neuro, marconato2023not}. 
This issue originates from \textbf{\textit{reasoning shortcuts}} (RSs): unwanted concept assignments that allow models to reach correct label predictions. These have been shown to affect many NeSy models \citep{marconato2023not, yang2024analysis}. 
While the decision-making appears correct on the surface, the underlying concepts are flawed, undermining the reliability and trustworthiness of the NeSy model.
For example, in the context of autonomous driving of \cref{ex:sym-grounding} and of \cref{fig:rss-main-fig}, the correct ``{\tt stop}'' prediction can be achieved by mistaking ``{\tt pedestrian}s'' with ``{\tt red\_light}s'', as both choices entail the same decision.

RSs essentially imply that concept learning is ill-posed in the context of NeSy AI: in many tasks, there is simply not enough information for a NeSy model to learn the correct concepts. Still, RSs are non-trivial, for several reasons. Again, they cannot be detected based on training (and possibly test) accuracy alone. Moreover, researchers, practitioners and stakeholders are likely to assume that the names and semantics of the symbols in the knowledge have to be aligned \textit{by construction} -- precisely because this is what happens in logic programming.
For instance, imagine having to write a logic program for the addition of two digits. The first step would be to choose how many predicates to use and how to name them. The most intuitive choice is to introduce ten binary predicates to model the (one-hot encoding of the) left digit and ten for the right one.  These symbols will then be used in the constraints assuming their name matches their meaning.  However, as we will see, there is no guarantee that a NeSy model trained to solve said task will use them as expected: it might well use the first ten predicates for the right digit and the others for the left one.
This is at best surprising, and at worst deleterious.

While this misalignment does not induce a drop in prediction accuracy on data \textit{in-distribution}, it becomes critical in situations where such distinctions matter.
In fact, \textbf{\textit{poorly grounded concepts may not transfer to out-of-distribution scenarios}}. This has consequences in applications where the data differs from the training distribution, \eg in continual learning \citep{marconato2023neuro}: despite being able to ensure predictions comply with the prior knowledge, NeSy models may behave erratically due to incorrectly ground concepts, similarly to \cref{fig:rss-main-fig} (Right), where the wrong activation of model concepts leads to a faulty prediction.
This also \textit{\textbf{impacts the model's interpretability}}.  When asked why it decided to ``{\tt stop}'' in \cref{fig:rss-main-fig} (Left), the model would explain that it stopped because there is a {\tt red\_light} on the read, while clearly this is not the case, and similarly for \cref{fig:rss-main-fig} (Right).
Finally, incorrect grounding also \textit{\textbf{affects down-stream applications}} like Neuro-Symbolic verification \citep{xie2019embedding}, which enables designers to formally verify whether learned models satisfy predetermined constraints defined in terms of high-level concepts, like ``does this autonomous vehicle always stop when there is a {\tt pedestrain} on the road?''. Clearly, incorrect grounding of {\tt pedestrian} would impair the trustworthiness of these tools.
In a nutshell, RSs can compromise some of those desirable properties -- among which reliability and interpretability -- that NeSy architectures are designed to possess, making them a primary target for mitigation.

RSs were observed in early NeSy work \citep{manhaeve2018deepproblog, chang2020assessing, topan2021techniques} and were shown to affect a variety of NeSy tasks \citep{bortolotti2024benchmark}, including high-stakes ones like autonomous driving.
Yet, they only recently received a formal treatment, revealing that RSs are hard to tackle \citep{marconato2023not, UmiliKR23, yang2024analysis, delong2024mars}, despite arising naturally from certain symmetries in the symbolic component of NeSy systems.
In turn, theoretical studies \citep{marconato2023not, wang2023learning, yang2024analysis, bortolotti2025shortcuts} have begun to identify under which conditions RSs can be provably avoided and when they cannot.
Intuitively, these conditions depend on four factors:  the structure of the prior knowledge, the coverage of the training set, the hypothesis class of learnable neural networks, and the choice of loss function.  Robustness to RSs often requires pairing standard NeSy training methods with \textbf{\textit{mitigation strategies}} that encourage better concept grounding by targeting the four aforementioned factors.
Building on these theoretical insights, we summarize and categorize existing mitigation strategies -- including knowledge editing, multi-task learning, prototypes, entropy maximization, and contrastive learning -- and analyse their effectiveness, scope and overall cost.  We conclude by noting how designing effective mitigation strategies that do not require a significant annotation cost remains an open problem.

Overall, we aim to consolidate the scattered literature on RSs in NeSy AI and offer a comprehensive overview of different approaches to mitigate them.

\paragraph{Contributions} This article provides a gentle introduction to reasoning shortcuts, unifies the existing literature on the topic, and connects it to the well-known symbol grounding problem. %
We provide a general perspective on RSs, highlighting that they cannot be avoided simply by designing {different neuro-symbolic architectures}. 
To this end, we compile all relevant theory on RSs through the perspectives of identifiability and statistical learning. 
We then organize known mitigation strategies to prevent RSs into a taxonomy, and explain which strategies can effectively reduce the likelihood of RSs.
Finally, we identify open problems and future directions, showing that RSs --- and thus correct symbol grounding --- extend beyond the NeSy models studied so far. %

\footnotetext{%
    To be precise, the prior knowledge for Task 2 would be $\BK = (\texttt{alarm} \implies \BK_1) \land (\lnot \texttt{alarm} \implies \BK_2)$ where \(\BK_1\) is the knowledge used for Task 1, and $\BK_2 = (\texttt{emergency} \land \lnot \texttt{pedestrian} \implies \texttt{go})$.%
}

\paragraph{Outline} The remainder of the article is organized as follows. \cref{sec:prelims} introduces the preliminaries necessary for understanding the theoretical material. 
\cref{sec:reasoning-shortcuts} presents the problem of reasoning shortcuts at a high level, providing examples and discussing their causes and impacts. 
\cref{sec:rss-theory} delves into the details by introducing the mathematical tools required to analyze reasoning shortcuts, exploring the issue from two perspectives: identification and statistical learning. 
\cref{sec:handling-rss} adopts a more applied perspective by examining the root causes of RSs, together with practical strategies for diagnosing their presence and for mitigating or addressing their effects, and discusses the benefits of each strategy.
\cref{sec:extensions} explores various extensions of the reasoning shortcut problem across different domains and discusses open problems and future directions. 
Finally, \cref{sec:related-work} reviews relevant related research, and \cref{sec:discussion} provides concluding remarks. %

% \newpage %
\section{Preliminaries}
\label{sec:prelims}

\paragraph{Notation.}  Throughout, we denote scalar constants by lowercase letters $x$, scalar (random) variables by uppercase letters $X$, vectors of constants $\vx$ and (random) variables $\vX$ in bold typeface, and sets (\eg $\calX$) by calligraphic letters.
Throughout, we use $p(X)$ to indicate the distribution of $X$ and $p(x)$ to indicate the probability of the event $X=x$.
If $\varphi$ is a logical formula over variables $\vX$, we say that the constants $\vx$ entail the formula $\varphi$ ($\vx \models \varphi$) if and only if replacing the variables $\vX$ in the formula $\varphi$ with $\vx$ makes the formula true. 
In this case, we say that $\vx$ satisfies or is consistent with the formula; otherwise, it violates or is inconsistent with the formula.  
See \cref{tab:glossary} for a glossary.

\begin{table}[!h]

    \centering
    \caption{Glossary of used symbols.} %
    \label{tab:glossary}
    \scalebox{0.85}{
    \begin{tabular}{llcll}
        \toprule
        \textbf{Symbol}
            & \textbf{Meaning}
        & & \textbf{Symbol}
            & \textbf{Meaning}
        \\
        \midrule
        $x$, $y$, $z$,
            & Scalar constants
        & &
        $X$, $Y$, $Z$
            & Scalar variables
        \\
        $\vx$, $\vX$
            & Vectors
        & &
        $\calX, \calY,\calC$
            & Sets
        \\
        $f(\vx)$, $p(\vc \mid \vx)$
            & Concept extractor
        & &
        $\beta(\vc)$, $p(\vy \mid \vc; \BK)$
            & Inference layer
        \\
        $\models$
            & Logical entailment
        & &
        $\BK$
            & Prior knowledge
        \\
        $\theta$
            & Network parameters
        & &
        $\calD$
            & Dataset
        \\
        \bottomrule
    \end{tabular}
    }
\end{table}

\subsection{From Neuro-Symbolic AI to Neuro-Symbolic Predictors}
\label{sec:nesy-ai}

Neuro-symbolic models aim to solve the long-standing problem of integrating learning and reasoning.
Over time, many radically different NeSy architectures have emerged \citep{garcez2022neural, de2021statistical,feldstein2024mappingneurosymbolicailandscape} that differ not only in what kind of logic they implement reasoning with (\eg propositional \citep{hoernle2022multiplexnet, ahmed2022semantic,DBLP:conf/aaai/BuffelliT23} vs. first-order \citep{lippi2009prediction, diligenti2012bridging, manhaeve2018deepproblog} and classical \citep{zhou2019abductive} vs. fuzzy \citep{donadello2017logic, van2020analyzing} vs. probabilistic \citep{manhaeve2018deepproblog, ahmed2022semantic,DBLP:conf/icml/FeldsteinJT23}), but also in how they interpret logic reasoning itself (\eg model vs. proof-based inference \citep{de2021statistical}) and in how they implement the computations (\eg fully neural \citep{rocktaschel2016learning} vs. hybrid neural-symbolic \citep{manhaeve2018deepproblog}).
This variety is reflected by the diversity of tasks that NeSy architectures can tackle, which range from hierarchical classification \citep{giunchiglia2020coherent, hoernle2022multiplexnet, ahmed2022semantic} and knowledge base completion \citep{rocktaschel2016learning}, to open-ended reasoning \citep{manhaeve2018deepproblog, badreddine2022logic}
and learning knowledge graph embeddings \citep{maene2025embeddings}.

Reasoning shortcuts have been studied mainly in the context of \textit{\textbf{NeSy predictors}}, a class of NeSy architectures specialized for constrained prediction tasks \citep{giunchiglia2022deep, dash2022review, marconato2023neuro}, denoted here as \textit{\textbf{NeSy tasks}};  a discussion of RS in other NeSy architectures is left to \cref{sec:extension-nesy-ai}.
A NeSy task requires predicting labels that are consistent with known constraints.  Each such task is defined by four elements:
a description of the \textit{\textbf{input}} $\vX$ with domain $\calX$,
a description of the $k$ \textit{\textbf{concepts}} $\vC = (C_1, \ldots, C_k)$ with domain $\calC = \calC_1 \times \ldots \times \calC_k$,
a description of $m$ discrete \textit{\textbf{labels}} $\vY = (Y_1, \ldots, Y_m)$ with domain $\calY = \calY_1 \times \ldots \times \calY_m$, %
and \textit{\textbf{prior knowledge}} $\BK$.
{Here, $\calX$ is the input domain (e.g., a vector space), $\calC$ is a finite set of \textit{\textbf{concept vectors}} $\vc$, and so are the label sets $\calY_i$.}

The input $\vX$ is usually high-dimensional and sub-symbolic (\eg an image), while the concepts $\vC$ are high-level and possibly human readable properties of the input (\eg the objects appearing in the image).
The prior knowledge $\BK$ encodes the constraints that we wish the model to satisfy---\eg safety constraints or relevant regulations---as a single propositional logical formula.\footnote{Although some NeSy predictor architectures support first-order logical formulas, we restrict our description to propositional formulas, for ease of exposition.  We postpone a discussion of how RSs transfer to the first-order case to \cref{sec:extension-nesy-ai}.}
Both the concepts $\vC$ and the labels $\vY$ appear as logic variables in $\BK$.
We ground these notions using a running example:

\begin{tcolorbox}[
    colback=gray!2!white,
    colframe=gray,
    arc=0pt,
    outer arc=0pt
]
\begin{example}
    \label{ex:boia}
    Consider the simplified \BOIA autonomous driving task~\citep{xu2020boia}. %
    Here, the inputs $\vx$ are dashcam images
    and the (binary) label $y$ indicates whether the vehicle is allowed to proceed.
    The task involves three binary concepts $C_\texttt{grn}$, $C_\texttt{red}$, $C_\texttt{ped}$ encoding the presence of green lights, red lights, and pedestrians in the input image, respectively,
    and the prior knowledge $\BK$ encodes the constraint that if either a pedestrian or a red light is visible in the image, the vehicle must stop: $\BK = (C_\texttt{ped} \lor C_\texttt{red} \Leftrightarrow \lnot Y)$.
    In this scenario, a NeSy predictor has to predict vehicle actions that are both accurate and comply with traffic regulations.
\end{example}
\end{tcolorbox}

Given a (noiseless) training set $\calD = \{ (\vx, \vy) \}$, a NeSy predictor is tasked with learning a mapping from inputs $\vx$ to labels $\vy$ that comply with the knowledge, that is, $(\vc, \vy) \models \BK$.\footnote{Here, $(\vc, \vy)$ is to be read as the concatenation of $\vc$ and $\vy$.}
The key challenge is that, just like in \cref{ex:boia}, the prior knowledge $\BK$ is not specified at the input level, but rather over the concepts.  These, in turn, are complex functions of the input that, in practice, cannot be manually specified.
For this reason, NeSy predictors adopt modular architectures that comprise a learnable \textit{\textbf{concept extractor}} responsible for predicting the concepts $\vC$ from the inputs $\vX$, and an \textit{\textbf{inference layer}} responsible for inferring the labels $\vY$ from $\vC$ compatibly with the prior knowledge $\BK$.
The former is typically implemented as a feed-forward neural network,\footnote{Although other architectures can be used \citep{diligenti2012bridging}.} and the latter using some form of (differentiable) symbolic reasoning.
Given an input $\vx$, a NeSy predictor first applies the concept extractor to obtain a distribution $p(\vC \mid \vx)$ over concepts, and then applies the inference layer to obtain a distribution $p(\vY \mid \vC; \BK)$ over outputs $\vy$, given the concepts.  Overall, the predictor defines a predictive distribution $p(\vY \mid \vX; \BK)$ (see \cref{sec:nesy-predictors} for details).

\begin{tcolorbox}[
    colback=gray!2!white,
    colframe=gray,
    arc=0pt,
    outer arc=0pt
]
\begin{example}
    Consider the \BOIA~\citep{xu2020boia} dataset illustrated in~\cref{ex:boia}. Here, the input $\vx \in \calX$ is a dashcam image. The concept extractor is a neural network that takes $\vx$ and predicts the probabilities of three independent binary concepts $(C_{\texttt{grn}}, C_{\texttt{red}}, C_{\texttt{ped}})$. For example, given an image $\vx$, the extractor may output $p(C_{\texttt{red}}=1 \mid \vx) = 0.9, \quad p(C_{\texttt{ped}}=1 \mid \vx) = 0.2, \quad p(C_{\texttt{grn}}=1 \mid \vx) = 0.1$. %
    These concept predictions are then passed to an inference layer, which combines them with prior knowledge $\BK = (C_{\texttt{ped}} \lor C_{\texttt{red}}) \Leftrightarrow \texttt{stop}$ to obtain the final decision $y$. Depending on the chosen implementation, this reasoning step may enforce the rule strictly, approximate it through fuzzy logic, or learn it implicitly during training. In this example, since the concept extractor assigns high probability to $C_{\texttt{red}}=1$, the system will likely infer $Y=\texttt{stop}$, indicating that the vehicle should stop.
\end{example}
\end{tcolorbox}

In most cases, NeSy predictors can be, and in fact are, trained using label supervision only.  This is not surprising if we consider that per-concept annotations can be expensive to collect in practice.
However---and this is critical for our discussion of RSs---this also means that \textit{the concepts themselves are often not supervised, and therefore act as latent variables}.

\subsection{The Variety of NeSy Predictor Architectures}
\label{sec:nesy-predictors}

\begin{figure}[!t]
    \centering
    \includegraphics[width=0.9\linewidth]{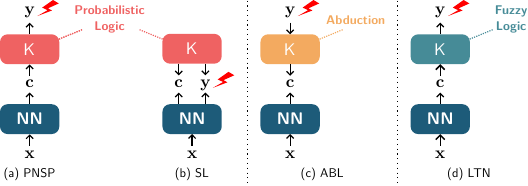}
    \caption{\textbf{Schematic illustration of NeSy predictors}.  All of them employ a neural network for extracting concepts and prior knowledge $\BK$.
    \PNSP (a) map inputs $\vx$ to (a distribution over) concepts $\vc$ and then uses probabilistic logic to infer labels $\vy$ consistent with $\BK$.
    The \SL (b) maps inputs to both concepts $\vc$ and labels $\vy$ and uses $\BK$ to define a loss term penalizing inconsistent predictions.
    During training, \ABL (c) abduces concepts from the ground-truth label $\vy$ using $\BK$ and uses them as pseudo-labels to train the concept extractor.
    \LTN (d) works similarly to \PNSP but uses \textit{fuzzy logic} to make $\BK$ differentiable.
    Lightning strikes indicate supervision.
    }
    \label{fig:nesy-architectures}
\end{figure}

Different NeSy predictor architectures differ in how they implement and use the concept extractor and the inference layer.
Next, we introduce the four architectures that have been most extensively studied in the RS literature, namely probabilistic NeSy predictors, the Semantic Loss, Logic Tensor Networks, and Abductive Learning. 
These are illustrated in \cref{fig:nesy-architectures}.

\paragraph{Probabilistic NeSy Predictors} A common method for designing NeSy architectures is the family of \textit{\textbf{Probabilistic NeSy Predictors}} (\PNSP) \citep{manhaeve2018deepproblog, manhaeve2021neural, yang2020neurasp, huang2021scallop}.
\PNSP can be viewed as combining a neural concept extractor with a probabilistic-logic reasoning layer \citep{de2015probabilistic}, as shown in \cref{fig:nesy-architectures} (a). 
Perhaps the best-known member of this family is DeepProbLog (\DPL)~\citep{manhaeve2018deepproblog, manhaeve2021neural}, a fully-fledged neuro-symbolic programming language based on Prolog and its probabilistic extension, ProbLog \citep{de2007problog, kimmig2011implementation}. %
From a simplified point of view, \PNSP define the following predictive distribution for any $\vx$ and $\vy$:
\[
    p(\vy \mid \vx; \BK)
    = \sum_{\vc} p(\vy, \vc \mid \vx; \BK)
    = \sum_{\vc} p(\vy\mid \vc; \BK) \, p(\vc \mid \vx)
    = \frac{1}{Z_{\vx}} \sum_{\vc} \Ind{(\vc, \vy) \models \BK} \, p(\vc \mid \vx)
    \label{eq:dpl-likelihood}
\]
Here, $p(\vC \mid \vx)$ is the predictive distribution over concepts computed by the concept extractor, %
$\Ind{\cdot}$ is the indicator function, and $Z_{\vx}$ is a normalization constant ensuring that the result is indeed a conditional distribution.  In words, the probability of a label $\vy$ is the sum of the probabilities of all concept vectors $\vc$ that are consistent with that label $\vy$ according to the knowledge $\BK$. %
Inference in \PNSP amounts to computing a most likely label, which usually requires solving the Maximum A Posteriori (MAP) problem $\argmax_{\vy} \ p(\vy \mid \vx ; \BK)$ \citep{koller2009probabilistic}.
Importantly, if a label $\vy$, together with the predicted concepts, would violate the knowledge $\BK$, then \textit{its probability is exactly zero and it will not be predicted}.
Learning in \PNSP is implemented via maximum likelihood estimation.  
That is, given a training set $\calD = \{ (\vx, \vy) \}$, learning amounts to maximizing the log-likelihood of the data, namely
\[
    \calL(p, \calD, \BK) = \frac{1}{|\calD|}
            \sum_{ (\vx, \vy) \in \calD } \ \log {p (\vy \mid \vx; \BK)}.
    \label{eq:dpl-learning-objective}
\]
by tuning the parameters of the concept extractor.  In practice, this is done via gradient descent, as the predictive distribution $p(\vy \mid \vx; \BK)$ is differentiable \citep{manhaeve2018deepproblog}.

\PNSP require evaluating the predictive probability $p(\vy \mid \vx; \BK)$, which involves summing over a potentially exponential number (in $k$, the cardinality of the concept vector) of vectors $\vc$, as per \cref{eq:dpl-likelihood}.  Hence, both inference and learning are worst-case intractable.\footnote{Specifically, \cref{eq:dpl-likelihood} can be understood as an instance of (weighted) model counting, the problem of computing (the total weight of all) models, or valid assignments, of a propositional logic formula. This problem is well known to be \#P-complete in general, see \eg \citep{roth1996hardness}.}
\DPL works around this issue by exploiting \textit{\textbf{knowledge compilation}} techniques \citep{darwiche2002knowledge} that leverage symmetries in the summation to rewrite it into a data structure -- a \textit{probabilistic circuit} \citep{choi2020probabilistic,vergari2021compositional,maene2025klay,derkinderen2025deeplogneurosymbolicmachine} -- that is potentially much more compact and supports efficient evaluation.
This allows \DPL to scale to practical NeSy tasks.
Other \PNSP refine and approximate these ideas (specifically, \cref{eq:dpl-likelihood}) to further improve scalability \citep{manhaeve2021approximate, huang2021scallop, winters2022deepstochlog, de2023differentiable, van2022anesi, choi2025ctsketch, chen2025neural}.

\paragraph{Semantic Loss}  %
The \textit{\textbf{Semantic Loss}} (\SL) \citep{xu2018semantic} also relies on probabilistic logic and knowledge compilation, but
uses them to convert the prior knowledge $\BK$ into a differentiable penalty term.  This can be used to steer any neural network classifier toward allocating low or even zero probability to inconsistent outputs.
{While the SL is a general purpose strategy, here we discuss how it can be exploited for defining a NeSy predictor \citep{marconato2023neuro}. The idea is to apply the SL to a neural network that acts \textit{both} as concept extractor and as classifier, \ie that outputs both $p(\vC \mid \vX)$ and $p(\vY \mid \vX)$.  This encourages the concepts and labels predicted by the network to be logically consistent with each other according to $\BK$}.\footnote{Whether this is a single neural network or two specialized networks makes no difference from a modeling perspective.}
The SL grows proportionally to how much probability mass the concept extractor associates to invalid configurations, or more precisely:
\[
    \SL(p_\theta, (\vx, \vy), \BK) =
        - \log \sum_\vc
        \Ind{ (\vc, \vy) \models \BK } p_\theta(\vc \mid \vx)
    \label{eq:semantic-loss}
\]
As in \DPL, efficient computation of this sum relies on knowledge compilation. 
During training, the \SL is paired with a standard supervised loss $\ell$ (\eg the cross-entropy loss), resulting in a joint objective of the form:
\[
    \calL(p_\theta, \calD, \BK) = \frac{1}{|\calD|} \sum_{(\vx, \vy) \in \calD}
        \ell(p_\theta, (\vx, \vy)) + \mu \, \SL(p_\theta, (\vx, \vy), \BK)
    \label{eq:empirical-semantic-loss}
\]
with $\mu > 0$ a hyperparameter.  Learning amounts to minimizing this joint objective on training data. At test time, predictions are computed directly by the neural network classifier $p_\theta(\vy \mid \vx)$ rather than by a symbolic layer as in \PNSP. %

\paragraph{Logic Tensor Networks}  Another well-known approach is \textit{\textbf{Logic Tensor Networks}} (\LTN{s})~\citep{donadello2017logic, badreddine2022logic}, which share elements with both \PNSP and \SL.
As with the \SL, they transform the symbolic knowledge $\BK$ into a differentiable real-valued function $\calT_{\BK}$ that evaluates how well predictions conform to the logical constraints using \textit{fuzzy logic}~\citep{van2020analyzing}.
This transformation is a proper \textit{relaxation}.  Specifically, it converts all the original Boolean variables in \BK (namely, the concepts and the labels) into real-valued variables in the range $[0, 1]$ encoding \textit{degrees of truth}.  At the same time, it converts all logic connectives (conjunctions, disjunctions, and negations) into real-valued operators capable of handling soft degrees of truth.  The translation yields a function $\calT_{\BK}$ that takes as input a distribution over the concepts $\vC$ and a distribution over the labels $\vY$, and outputs the degree (in $[0, 1]$) to which these satisfy the knowledge $\BK$, the higher the better.
By default, \LTN{s} employ a transformation based on the \textit{product real logic}~\citep{donadello2017logic} T-norm to ensure that $\calT_{\BK}$ is fully differentiable, although other options are available~\citep{van2020analyzing}.
During training, \LTN{s} penalize the concept extractor $p(\vC \mid \vX)$ for violating the prior knowledge by minimizing the following: %
\[
    \calL(p, \calD, \BK) = 1- \frac{1}{|\calD|} \sum_{(\vx, \vy) \in \calD} \calT_{\BK} \big( p(\vC \mid \vx), \Ind{\vY = \vy} \big)
\]
Here, $\Ind{\vY = \vy}$ is a (deterministic) distribution that assigns all probability mass to the ground-truth label $\vy$.
The purpose of this objective is to encourage $p(\vC \mid \vX)$ to concentrate its mass on concepts that satisfy $\BK$.
Similarly to \PNSP, \LTN{s} employ the prior knowledge at inference time.  
They proceed in two steps: they first compute the most probable concept vector $\hat\vc = \argmax_{\vc} p_\theta(\vc \mid \vx)$ in a forward pass, then select the label $\hat\vy$ that maximizes the satisfaction of the knowledge $\calT_{\BK}(\hat\vc, \Ind{\vY = \hat\vy})$.

\paragraph{Abductive Learning}  
Another well-studied approach is \textit{\textbf{Abductive Learning}} (\ABL)~\citep{zhou2019abductive}.  Compared to the above NeSy predictors, it works backwards: rather than inferring a prediction from the concepts, it constrains the predicted concepts using the ground-truth label.
To this end, during training, \ABL finds the concept vectors $\hat\vc$ that are both close to those predicted by the concept extractor (according to some pre-defined distance metric) and that according to $\BK$ entail the ground-truth label $\vy$.  It then uses them as pseudo-labels to supervise the concept extractor.
The pseudo-labels are obtained by solving the following optimization problem: %
\[
    \hat \vc = \argmin_{\vc' \in \calC} d(\bar \vc, \vc') \qquad \mathrm{s.t.} \qquad  (\vc', \vy) \models \BK
    \label{eq:abl-prediction-concept}
\]
where $\bar \vc = \argmax_{\vc \in \calC} p(\vc \mid \vx)$,
and $d$ is a suitable distance metric.  The constraint ensures that $\hat\vc$ entails the ground-truth label $\vy$.
The choice of the distance metric influences the type of weak supervision obtained, thereby intuitively biasing learning toward certain solutions.
The training objective of the concept extractor is to maximize the (log)-likelihood of the pseudo-labels using $\hat \vc$ from \cref{eq:abl-prediction-concept}:
\[
    \calL(p, \calD, \BK) = \cfrac{1}{| \calD |} \sum_{(\vx, \hat\vc)} \log p(\hat \vc \mid \vx)
\]
At inference time, \ABL first predicts the concepts and then uses the symbolic knowledge to obtain the final prediction. %

\subsection{NeSy Predictor Architectures: Differences and Similarities}
\label{sec:nesy-predictors-diff-sim}

These different NeSy predictor architectures all share a key property:  \textit{\textbf{if the concept extractor allocates all probability mass to a single concept vector, then all architectures will output a label compatible with the rules of propositional logic}}.  Consider \MNISTAdd~\citep{manhaeve2018deepproblog}, where the task is to predict the sum (\eg $y = 9$) of the digits appearing in two \MNIST \citep{lecun1998mnist} images (\eg $\vx = \MFour\MFive$).  The concepts $\vC = (C_1, C_2)$ encode the individual digits (\eg $\vc = (4, 5)$).\footnote{Here and elsewhere, we simplify the presentation by using numerical variables. It is always possible to encode these and the corresponding constraints into propositional logic.} Prior knowledge enforces the prediction to be their arithmetic sum: $\BK = (Y = C_1 + C_2)$. In this example, if the concepts $C_1 = 2$ and $C_2 = 3$ are predicted with certainty, then all architectures will predict the label $Y = 5$ with certainty---thus matching what any any logical encoding of the arithmetic sum would do.  However, for concept predictions that are not certain, different architectures can output different label distributions.  
This distinction is particularly important during optimization, as gradients are shaped by how probability mass over concepts is aggregated and filtered by the inference mechanism \citep{van2020analyzing}.

At the same time, these four NeSy predictor architectures differ in several subtle but significant ways.
The first one concerns \textit{\textbf{efficiency}}.  \LTN and \ABL can have an advantage over probabilistic logic approaches (like \PNSP and the \SL) in that inference---and therefore training---does not require summing over all possible concept configurations, making it potentially more efficient, although knowledge compilation and approximation strategies help bridge the gap \citep{huang2021scallop}.
This, however, often comes at a cost, in that \LTN and \ABL tend to be more susceptible to local minima~\citep{badreddine2022logic}.
On the other hand, fuzzy relaxations (as used by \LTN) may not be entirely accurate to the original prior knowledge, in the sense that, for certain problems, the optima of the satisfaction function $\calT_{\BK}$ may not correspond to models (that is, $0$--$1$ solutions) of $\BK$ \citep{giannini2018convex, van2022analyzing}.  
Moreover, the fuzzy transformation may lead to mathematically different relaxations for logically equivalent constraints \citep{di2020efficient, van2022analyzing}.  
Probabilistic logic is not affected by these issues \citep{xu2018semantic}.
\ABL is special in that it does not require the reasoning step to be differentiable, and as such it does not need to relax or extend the semantics of the prior knowledge at all.

A second difference concerns \textit{\textbf{validity guarantees}}, which set apart layer-based approaches from penalty-based ones.
The \SL ``bakes'' the prior knowledge directly into the neural network, meaning that finding a high-quality output amounts to a simple forward pass over the network: $\BK$ plays no role during inference.  This also results in a lower memory footprint compared to \PNSP, as the probabilistic circuit can be dropped after training \citep{di2020efficient}.
The downside is that, while \PNSP and \ABL ensure invalid outputs will not be predicted, this is not the case for the \SL, unless the neural network attains exactly zero Semantic Loss during training, and even then this property might not carry over to test and out-of-distribution samples.
Regardless, it has been shown that \PNSP and the \SL have the same effect on the underlying neural network, that is, when learned to optimality, both models yield concept extractors that comply with the prior knowledge $\BK$ \citep{marconato2023neuro}.
\LTN sits in-between these alternatives, as it uses the prior knowledge at inference time, but may output inconsistent predictions if the most likely inputs violate the knowledge.
To resolve this issue, fuzzy logic layers such as CCN+~\citep{giunchiglia2024ccn} ensure all outputs comply with the knowledge.

\subsection{NeSy Predictors: Benefits}
\label{sec:nesy-predictors-benefits}

Despite these differences, all NeSy predictors offer a number of key benefits:
\begin{itemize}[leftmargin=2em]

    \item[\textbf{B1}] \textit{\textbf{Performance}}:  Just like regular neural networks, they are fully-fledged deep learning architectures that can handle complex low-level inputs via end-to-end training and latent representations.

    \item[\textbf{B2}] \textit{\textbf{Validity}}: Unlike regular neural networks, their predictions \textit{comply with the prior knowledge $\BK$}, possibly also with guarantees and out-of-distribution, cf. \cref{sec:nesy-predictors-diff-sim}. This is essential in high-stakes applications where predictions have to comply with safety or structural constraints.

    \item[\textbf{B3}] \textit{\textbf{Reusability}}: It is straightforward to \textit{reuse the learned concept extractors in downstream tasks}, \eg out-of-distribution tasks~\citep{marconato2023not}, continual learning scenarios \citep{marconato2023neuro}, model verification \citep{xie2022neuro, zaid2023distribution, morettin2024unified}, and shielding~\citep{yang2023safe}.

    \item[\textbf{B4}] \textit{\textbf{Interpretability}}: Users can not only inspect the concept-level predictions and the symbolic inference steps to make sense of the model's predictions, they can also \textit{trace back the model's prediction to the underlying concepts}  using gradient-based \citep{sundararajan2017axiomatic} or formal \citep{huang2021efficient} explainability techniques.  This supports stakeholders in assessing the reliability of the predictor and potentially enables them to supply corrective feedback \citep{teso2023leveraging}.

\end{itemize}
Compared to regular neural networks, which primarily target \textbf{B1}, benefits \textbf{B2}--\textbf{B4} make NeSy predictors an ideal choice for high-stakes applications \citep{di2020efficient, hoernle2022multiplexnet, marconato2023not, yang2023safe}.
However, reusability (\textbf{B3}) and interpretability (\textbf{B4}) \textit{\textbf{hinge on the concepts being grounded appropriately}}.  In fact, unless the learned concepts possess reasonable semantics, their meaning might be opaque to stakeholders, making it difficult to properly explain the model's predictions with them.  Furthermore, as exemplified by \cref{fig:rss-main-fig}, reusing poorly grounded concepts in downstream applications may lead to unintended consequences.
This is precisely where reasoning shortcuts enter the picture, as discussed next.

% \newpage %
\section{A Gentle Introduction to Reasoning Shortcuts}
\label{sec:reasoning-shortcuts}
\label{sec:a-gentle-intro-to-rss}

When does a NeSy predictor ground the concepts incorrectly?
It may be tempting to assume that---as long as the training data $\calD = \{(\vx, \vy)\}$ is abundant and noiseless, and the prior knowledge is complete and correct\footnote{That is, for all inputs, it correctly and unambiguously allows to infer the ground-truth label from the ground-truth concepts, as in \MNISTAdd.}---NeSy predictors that minimize the training loss will predict the concepts correctly, just like in standard supervised learning. 
However, this is not the case.  
This is due to \textit{\textbf{reasoning shortcuts}} (RSs), which we informally define as follows:

\begin{tcolorbox}[
    colback=orange!2!white,
    colframe=orange,
    arc=0pt,
    outer arc=0pt
]
    \begin{definition}[Reasoning Shortcut, Informal]
        \label{def:rs-informal}
        A reasoning shortcut is a situation in which a NeSy predictor attains accurate label predictions that comply with the prior knowledge by {grounding concepts incorrectly}.
    \end{definition}
\end{tcolorbox}

To ground intuition, consider the following example:

\begin{tcolorbox}[
    colback=gray!2!white,
    colframe=gray,
    arc=0pt,
    outer arc=0pt
]
\begin{example}[RSs in \MNISTAdd~\citep{manhaeve2018deepproblog}]
    \label{ex:mnist-rss-full} 
    Consider a toy instance of \MNISTAdd where the training set consists of just two examples:
    \[
        (\MFour\MFive) \mapsto 9 \quad \text{and} \quad (\MThree\MTwo) \mapsto 5.
    \]
    Assume that the concept extractor processes the \MNIST images separately.  Ideally, it would ground the concept correctly, that is, learn the intended image-concept mapping -- that is, $\{ \MTwo \mapsto 2, \MThree \mapsto 3, \MFour \mapsto 4, \MFive \mapsto 5 \}$ -- as it adheres to the constraints and attains high label accuracy.
    But it can alternatively ground them incorrectly, that is, learn the less intuitive input-concept mapping $\{ \MTwo \mapsto 4, \MThree \mapsto 1, \MFour \mapsto 3, \MFive \mapsto 6\}$, which also complies with the constraints and achieves high label accuracy.
\end{example}
\end{tcolorbox}

While this is not a realistic application, the same issue affects also high-stakes tasks:

\begin{tcolorbox}[
    colback=gray!2!white,
    colframe=gray,
    arc=0pt,
    outer arc=0pt
]
\begin{example}[RSs in \BOIA]
    \label{ex:boia-rs}
    Consider \cref{ex:boia}.
    A NeSy predictor that grounds the concepts of green light, red light, and pedestrian correctly achieves high accuracy, as it will also correctly infer that it has to stop whenever a pedestrian or a red light appear in the input image.
    However, a different NeSy predictor that systematically confuses pedestrians with red lights achieves the same accuracy, since, according to \BK, both lead to the correct
    ${\tt stop}$ action \citep{marconato2023not}, as shown in \cref{fig:boia-rs} below.
\end{example}

\begin{minipage}{\linewidth}
    \centering
    \begin{minipage}{0.3\linewidth}
        \centering
        \scalebox{0.9}{
        \begin{tabular}{cc|c}
            \toprule
            $C_{\tt red}$
                & $C_{\tt ped}$
                & $Y$
            \\
            \midrule
            0
                & 0
                & 1 (go)
            \\
            0
                & 1
                & 0 (stop)
            \\
            1
                & 0
                & 0 (stop)
            \\
            1
                & 1
                & 0 (stop)
            \\
            \bottomrule
        \end{tabular}
        }
    \end{minipage}%
    \begin{minipage}{0.7\linewidth}
        \centering
        \includegraphics[height=10em]{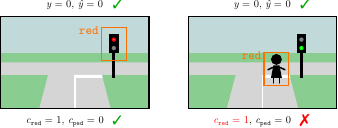}
    \end{minipage}
    \captionof{figure}{Left: truth table of the simplified prior knowledge $\BK$ used in \cref{ex:boia}: the two concepts are interchangeable, in that as soon as one of them fires, the predictor infers the same (correct) action.
    Right: illustration of this RS: the model can confuse pedestrians and red lights with no drop in training loss.}
    \label{fig:boia-rs}
\end{minipage}

\end{tcolorbox}

These examples show that correct and faulty NeSy predictors cannot be distinguished by label accuracy alone, and therefore \textit{\textbf{there is no reason why, during training, NeSy predictors should favor one concept mapping over the other}}.
They also indicate that RSs can occur even when the prior knowledge is accurate and complete, and the training data is noiseless.
Moreover, RSs can persist even if the training data is exhaustive (\eg if the \BOIA training set encompasses examples containing all possible combinations of green lights, red lights, and pedestrians), as the correct and faulty models yield the same predictions on all examples.

\subsection{Causes, Frequency and Impact}
\label{sec:causes-frequency-impact}

\paragraph{Why do RSs arise?}  Roughly speaking, RSs stem from two related issues: on the one hand, the prior knowledge $\BK$ may allow inferring the correct labels $\vy$ from improperly grounded concept vectors $\vc$; on the other, the concept extractor is expressive enough (\eg has enough layers) to acquire any such incorrect input-concept mapping. %
Our two examples satisfy both conditions.
Their combination \textit{\textbf{introduces ambiguity in the learning problem}}, meaning that NeSy predictors are free to learn incorrect concept mappings and still achieve high accuracy.
Properly understanding the root causes of RSs, however, requires more technical background, which we provide in \cref{sec:rss-theory}, hence we postpone a more comprehensive discussion to \cref{sec:root-causes}.
Importantly, clarifying the causes of RSs allows us to discriminate between risky and safe NeSy tasks, and to design RS mitigation strategies.  
We explore these topics in \cref{sec:handling-rss}.

\paragraph{Do RSs occur in practice?}  
The above discussion entails that RSs can occur whenever the prior knowledge is not ``strict'' enough, but it does not prove that they \textit{must} occur.  
Arguably, NeSy predictors \textit{can} learn the intended concepts even in this case.  
The question is then whether RSs occur in practice.  
While it is difficult to gauge their frequency in real-world applications, the literature abounds with situations in which NeSy predictors trained in standard conditions fall prey to RSs \citep{manhaeve2018deepproblog, wang2019satnet, chang2020assessing, manhaeve2021neural, marconato2023not, marconato2024bears, bortolotti2024benchmark, yang2024analysis, delong2024mars}, suggesting that \textit{\textbf{RSs occur naturally in NeSy benchmarks}}.

\paragraph{Do RSs make NeSy predictors unusable?}  The short answer is \textit{no}.  More precisely, in applications where the only goal is to obtain accurate (\textbf{B1}) and valid (\textbf{B2}) predictions, RSs are not an issue.
In fact, \cref{def:rs-informal} makes it clear that RSs are a symbol grounding issue that leaves label accuracy unchanged.
Hence, \textit{\textbf{the label predictions of affected and unaffected NeSy predictors can be just as accurate}}.
We also remark that RSs \textit{\textbf{have no impact on validity}} (\textbf{B2}).  As mentioned in \cref{sec:nesy-predictors-diff-sim}, predictors like \PNSP, \ABL, and \LTN ensure their output is \textit{always} consistent with the prior knowledge by explicitly searching for a label $\vy$ that, paired with the predicted concepts vectors $\hat\vc$ (or their predictive distribution $p(\vC \mid \vx)$), satisfies the knowledge $\BK$.  Validity is built-in, \ie it does not depend on the concepts predicted by the model.

{A tricky aspect of RSs is that we cannot detect them by monitoring label predictions alone}.
We will discuss more reliable diagnostic techniques in \cref{sec:diagnosis} and strategies for rendering predictors aware of their own RSs in \cref{sec:awareness}. 
{Furthermore, \textit{\textbf{RS mitigation strategies can greatly reduce the presence of RSs or even entirely remove them}} to ensure NeSy predictors are also reusable (\textbf{B3}) and interpretable (\textbf{B4}). We overview these strategies in \cref{sec:mitigations}.}

\paragraph{What consequences do RSs have?}  RSs can seriously compromise reusability (\textbf{B3}) and interpretability (\textbf{B4}).
Let us begin from reusability.
RSs may exploit concept groundings that do not work in other application settings, meaning that affected concepts suffer from \textit{\textbf{poor generalization beyond the specific data and prior knowledge used during training}}~\citep{marconato2023not, li2023learning, bortolotti2024benchmark}.
To see this, consider the following example, taken from \citep{marconato2023not}:

\begin{tcolorbox}[
    colback=gray!2!white,
    colframe=gray,
    arc=0pt,
    outer arc=0pt
]
\begin{example}
    \label{ex:boia-ambulance}
    Consider the NeSy predictor confusing pedestrians with red lights in~\cref{ex:boia-rs}.
    Imagine pairing it with prior knowledge intended for an autonomous ambulance use-case, encoding that, in case of an emergency, it is allowed to cross red lights.  
    When emergencies arise, the resulting NeSy predictor would decide to $y = {\tt go}$ when there are pedestrians on the road.
\end{example}
\end{tcolorbox}

RSs also affect concept reuse in downstream applications where the semantics of the concepts matters.
For instance, in \textit{\textbf{continual learning}} one is concerned with learning models that generalize across a sequence of learning tasks.  
NeSy predictors were shown to struggle with this precisely because of RSs \citep{marconato2023neuro}:

\begin{tcolorbox}[
    colback=gray!2!white,
    colframe=gray,
    arc=0pt,
    outer arc=0pt
]
\begin{example}
    Consider the shortcut predictor from \cref{ex:boia-rs}.
    Imagine fine-tuning it on a new NeSy task that also includes the concept of ``stop sign'' and a new rule based on it, and specifically that whenever a stop sign is detected, the vehicle must stop.
    When fine-tuned on this new task, the concept extractor may mistakenly learn to equate stop signs and red lights, whose semantics are corrupted.
\end{example}
\end{tcolorbox}

Another affected application is \textit{\textbf{neuro-symbolic verification}} \citep{liu2021algorithms, xie2022neuro, zaid2023distribution, morettin2024unified,manginas2025a}.
In model verification, the goal is to formally verify whether a neural network satisfies some property of interest, such as robustness to adversarial attacks or fairness.  
This involves writing a formal specification of the property, translating the model into a logical formula (or a similar representation), and then using tools from automated reasoning to check whether the model's formula is compatible with the specification.
NeSy verification is similar, except that the property of interest is specified in terms of high-level concepts.  Following the neuro-symbolic setup, their definitions are provided by a trained concept extractor -- provided by a third party -- that also gets translated into a logical formula for the purpose of verification.  Clearly, if the concepts are incorrect, this will affect the meaning of the property to be verified.

Finally, \textit{\textbf{interpretability}} (\textbf{B4}) also relies on the concepts being grounded appropriately \citep{marconato2023not, koh2020concept, poeta2023concept, tapli2026rethinking}.
Unless these possess human-aligned semantics, their meaning might be opaque to stakeholders, impairing understanding:

\begin{tcolorbox}[
    colback=gray!2!white,
    colframe=gray,
    arc=0pt,
    outer arc=0pt
]
\begin{example}
    In~\cref{ex:boia-rs}, imagine the input scene portrays a pedestrian.
    An affected NeSy predictor would (correctly) predict {\tt stop}, but its explanation would indicate that the decision depends on the presence of a red light.
    This explanation is misleading, and in fact it does not reflect the input at all.
\end{example}
\end{tcolorbox}

% \newpage %
\section{Theory of Reasoning Shortcuts}
\label{sec:rss-theory}

In this section, we outline the two mainstream formalizations of RSs.
\cref{sec:rss-as-identifiability} discusses RSs from an \textit{identifiability} perspective, investigating under what conditions models achieving high likelihood {can} in fact learn correct concepts in the infinite sample case. 
\cref{sec:rss-as-statistical-learning} instead is concerned with guarantees -- in terms of empirical risk minimization -- for learning correct concepts from finite data.
The so-far vague notion of ``correct'' concept will be clarified in the following sections.

\paragraph{Additional notation}  
{Going forward, we will often treat distributions over a finite set of values as vectors in a simplex.
We will write $\Delta_\calC$ to indicate the simplex defined by the values contained in $\calC$:
$$
    \textstyle
    \Delta_\calC = \{\vp \in [0, 1]^{|\calC|}\mid \sum_{i=1}^{|\calC|} p_i = 1\}
$$
Furthermore, we will write $\Vset{\Delta_\calC}=\{\vp \in \{0, 1\}^{|\calC|}\mid \sum_{i=1}^{|\calC|} p_i = 1\}$ to indicate the \textit{vertices} of the simplex $\Delta_\calC$.  This is simply a collection of one-hot vectors, one for each value in $\calC$.}

\subsection{Setting and Assumptions}
\label{sec:assumptions}

In order to formally describe RSs, we have to clarify the link between the observed data, the ground-truth concepts, %
and the concepts learned by the predictor.  This link is provided by the \textit{\textbf{data generation process}}, described next, which forms a solid basis for both theoretical approaches.

\begin{figure}[!h]
    \centering
    \usetikzlibrary{positioning,fit,calc}

\begin{tikzpicture}[
    scale=1,
    transform shape,
    node distance=1cm and .75cm,
    minimum width={width("G")+15pt},
    minimum height={width("G")+15pt},
    mynode/.style={draw,ellipse,align=center},
    simplenode/.style={align=center}
]
    \node[simplenode] (X) {$ $};
    \node[mynode, right=of X] (C1) {$C_1$};
    \node[simplenode, right=of C1] (Ch) {$\cdots$};
    \node[mynode, right=of Ch] (Ck) {$C_k$};
    \node[mynode,  fill=black!13!white, above=of Ch] (Xc) {$\vX$};
    \node[mynode, left=of X] (Gk) {$G_k$};
    \node[simplenode, left=of Gk] (Gh) {$\cdots$};
    \node[mynode, left=of Gh] (G1) {$G_1$};
    \node[mynode,  fill=black!13!white, above=of Gh] (Xg) {$\vX$};

    \node[mynode,  below=of Ch, fill=black!13!white] (Yhat) {$\vY$};
    \node[mynode,  below=of Gh, fill=black!13!white] (Ytrue) {$\vY$};

    \node[draw=orange, thick, fit=(C1)(Ck), inner sep=5pt%
    ] (Cbox) {};
    \node[draw=orange, thick, fit=(G1)(Gk), inner sep=5pt%
    ] (Gbox) {};

    \draw[->, thick, dashed, orange] ([yshift=0pt]Gbox.north) 
    to[out=45, in=135] node[midway, above] {\color{orange}$\alpha_f$}
    ([yshift=0pt]Cbox.north);
    
    \path
        (Xc) edge[-latex](C1)
        (Xc) edge[-latex]  node[near start, right] {$f$}  (Ck) 
        (Xg) edge[-latex]  node[near start, left] {$f^*$}  (G1)
        (Xg) edge[-latex] (Gk)
        (C1) edge[-latex] (Yhat)
        (Ck) edge[-latex]  node[near end, right] {$\beta$}  (Yhat)
        (G1) edge[-latex] node[near end, left] {$\beta^*$} (Ytrue)
        (Gk) edge[-latex] (Ytrue);

\end{tikzpicture}
    \caption{
    \textbf{Left}. The data generation process: $f^*$ maps inputs $\vX$ onto (distributions over) ground-truth concepts $\vG$, and $\beta^*$ maps these to (distributions over) labels $\vY$.  In a standard learning setting, only $\vX$ and $\vY$ are observed (in light gray), whereas $\vG$ are latent.
    \textbf{Right}. A NeSy predictor maps inputs $\vX$ to concepts $\vC$ via a learned $f$ and infers labels $\vY$ via an inference layer $\beta$.  By \cref{eq:definition-of-alpha}, doing so entails a map $\alpha_f$ (in \textcolor{orange}{\textbf{orange}}) from ground-truth concepts $\vG$ to learned concepts $\vC$.
    }
    \label{fig:generative-process}
\end{figure}

\paragraph{The data generation process}  First we distinguish between \textit{\textbf{ground-truth concepts}} $\vg =(g_1, \ldots, g_k) \in \calC$ underlying the data and \textit{\textbf{learned concepts}} $\vc =(c_1, \ldots, c_k) \in \calC$.
For instance, in~\cref{ex:boia-rs} there are two binary concepts encoding the presence of pedestrians and red traffic lights: $g_{\tt ped}$ and $g_{\tt red}$ are the true values, while $c_{\tt ped}$ and $c_{\tt red}$ are the predicted values.

The \textit{training {examples}} $(\vx, \vy)$ are sampled from a ground-truth joint distribution $p^*(\vX, \vY) = p^*(\vY \mid \vX) p^*(\vX)$, which is determined by the ground-truth concepts $\vg \in \calC$ associated with the input $\vx$.
Specifically, the values of the ground-truth concepts are sampled from a ground-truth conditional distribution $p^*(\vG \mid \vX)$, while the ground-truth labels $\vy$ are sampled from $p^*(\vY \mid \vG; \BK)$.  It is assumed that this distribution implements the background knowledge $\BK$, that is, it assigns zero or low probability to labels incompatible with the constraints.\footnote{\rechanged{More specifically, for technical reasons, we assume the inference layer of the predictor behaves as in \cref{eq:condition-on-beta}.}}
It is useful to think of these two {distributions} as \textit{functions}: 
the former as a function $f^*: \calX \to \Delta_{\calC}$ mapping {each input to a \textit{distribution}} over ground-truth concepts, and the latter as a function  $\beta^*: \Delta_\calC \to \Delta_\calY$ mapping {each distribution} over concepts to {a distribution} over labels.
Elements $f(\vx) \in \Delta_\calC$ are probability distributions $p(\vC \mid \vx)$ over the possible concept vectors $\vc \in \calC$. Whenever $f(\vx)$ allocates all probability mass to one specific $\vc \in \calC$, \ie it is one-hot, it is a vertex of the simplex and as such it belongs to $\Vset{\Delta_\calC} \subset \Delta_\calC$. Moreover, abusing notation, we write
\[
    \beta^*(\vc) := \beta^*( \Ind{\vC=\vc} ) \; \forall \vc \in \calC
\]
to indicate the map $\beta^*:\calC \to \Delta_\calY$ from the set of \textit{concepts} to distribution over labels. 

Overall, $p^*(\vY \mid \vX; \BK)$ results from composing $f^*$ and $\beta^*$. A visualization of the data generation process is reported in \cref{fig:generative-process} (Left).
With this in mind, we present the first assumption capturing how data is generated, which will be used throughout the theoretical material.

\begin{tcolorbox}[
    colback=violet!2!white,
    colframe=violet,
    arc=0pt,
    outer arc=0pt
]
\begin{assumption}[Completeness] 
    \label{assu:completeness-dgp}
    Let $f^*: \calX \to \Delta_\calC$ be the ground-truth concept extractor, and $\beta^*: \Delta_\calC \to \Delta_\calY$ the inference function induced by the prior knowledge $\BK$.
    Let $p^*(\vX)$ be the marginal distribution over input variables. \rechanged{We assume that the joint data distribution is generated as:}
    \[
        p^*(\vX, \vG, \vY) =
        \changed{p^*(\vY \mid \vG; \BK) p^*(\vG \mid \vX) p^*(\vX) }
        \label{eq:joint-input-label-concept-distr}
    \]
    where $p^*(\vG \mid \vX) := f^*(\vX)$ and
    $p^*(\vY \mid \vG; \BK) = { \beta^*(\vG) } $. In particular, we have  
    $p^*(\vY \mid \vX; \BK) := (\beta^* \circ f^*)(\vX)$.
\end{assumption}
\end{tcolorbox}

Under \cref{assu:completeness-dgp},\footnote{
\rechanged{\cref{assu:completeness-dgp} can be viewed as the distributional analogue of the invertibility assumption adopted by \citep{marconato2023not}. 
In that case, inputs are generated through an invertible map $f^\star:(\vG, \vS) \mapsto \vX$, 
where $\vG$ denotes the ground-truth concepts and $\vS$ the style variables. The invertibility of $f^\star$ guarantees that the concepts can be recovered from the input (as in \cref{assu:concept-extractor}) and therefore constitute a sufficient statistic of $\vX$ for predicting $\vY$. 
Our formulation encodes the same requirement directly at the level of probability distributions, without requiring concepts to be deterministically associated to the input $\vX$.}
} 
the ground-truth concept extractor becomes a \textit{sufficient statistic} of the input $\vX$ for inferring the ground-truth labels $\vY$, that is, the ground-truth concept distribution output by $f^*$ retains all information necessary to find the correct labels with $\beta^*$.
\rechanged{This is possible in all tasks from \rsbench \citep{bortolotti2024benchmark}, where ground-truth concepts are fully predictive of final labels. Notice that, however,  
}
this is not necessarily true in all practical cases, \eg when the concept space is ``too restrictive'' compared to the label space.  For example, in \MNISTAdd, if there are only two ground-truth digits, then sums above $18$ cannot be generated.
{The above assumption excludes such cases from the training data.}
\changed{
Furthermore, \cref{assu:completeness-dgp} ensures the labels are independent from the input given the ground-truth concepts, that is, $\vX \indep \vY \mid \vG$, and that all pairs $(\vG, \vY) \sim p^*(\vG, \vY)$ are consistent with the knowledge, \ie $(\vG, \vY) \models \BK$.
}

\paragraph{Simplifying assumptions on the data generation process} 
While \cref{assu:completeness-dgp} %
underlies all theoretical works on RSs \citep{yang2024analysis, wang2023learning, marconato2023not, he2025learnability}, 
we introduce two additional assumptions that make the theoretical analysis more manageable. 
The first assumption restricts the ground-truth concept extractor $f^*$: It should assign all probability mass to a single vector of concepts $\vg$ for each input $\vx$.

\begin{tcolorbox}[
    colback=violet!2!white,
    colframe=violet,
    arc=0pt,
    outer arc=0pt
]
\begin{assumption}[Extrapolability]
    \label{assu:concept-extractor}
    The ground-truth concept extractor is a function $f^*: \calX \to \Vset{\Delta_\calC}$, \ie for all inputs $\vx \in \calX$, the ground-truth concept distribution $f^*(\vx)$ is a one-hot (deterministic) distribution. 
\end{assumption}
\end{tcolorbox}

\cref{assu:concept-extractor} allows us to retrieve ground-truth concepts $\vG$ from inputs $\vX$ via $f^*$. The next assumption ensures that, with the prior knowledge $\BK$, each vector of concepts $\vg$ is associated with a single label $\vy$.

\begin{tcolorbox}[
    colback=violet!2!white,
    colframe=violet,
    arc=0pt,
    outer arc=0pt
]
\begin{assumption}[Determinism]
    \label{assu:inference-layer}
    {The ground-truth inference layer defines a function $\beta^*: \calC \to \Vset{\Delta_\calY}$ that maps each concept vector $\vc\in\calC$ to a one-hot (deterministic) distribution over labels $\beta^*(\vc)$.}
\end{assumption}
\end{tcolorbox}

In words, \cref{assu:inference-layer} ensures that $\beta^*$ assigns a single label $\vy$ to each concept vector $\vc$. %
\cref{assu:concept-extractor,assu:inference-layer} often apply in experimental regimes. 
For \MNISTAdd and \BOIA, the concepts in the input image can be unambiguously determined (\cref{assu:concept-extractor} is satisfied), and all concept vectors $\vc$ determine a single output label (\cref{assu:inference-layer} is satisfied), \eg in \MNISTAdd we use $Y = C_1 +C_2$. 
Together, \cref{assu:concept-extractor,assu:inference-layer}
ensure that each input $\vx \in \calX$ is mapped to a single label $\vy \in \calY$. 
The same holds for many common NeSy datasets \citep{bortolotti2024benchmark}. %

Before proceeding, we have to define the support of the input and ground-truth concept distributions. We consider the marginal distributions $p^*(\vX)$ and 
$p^*(\vG) := \bbE_{\vx \sim p^*(\vX)} [f^*(\vx)]$. 
We indicate the support of input data with $\mathrm{supp}(\vX) \subseteq \calX$, which contains the inputs with non-zero probability according to $p^*(\vX)$. 
Similarly, we denote the support of ground-truth concepts by $\mathrm{supp}(\vG) \subseteq \calC$.

\subsection{Perspective: Identification}
\label{sec:rss-as-identifiability}

{In this section, we ask whether a NeSy predictor trained in a standard supervised fashion is guaranteed to learn the ground-truth concept extractor.  Specifically, we want to determine whether---depending on the choice of knowledge, training data, and architectural bias---the ground-truth concept extractor is the \textit{unique maximum} of the likelihood of the training data. If so, then any NeSy predictor that attains maximum likelihood will necessarily learn the ground-truth concept extractor, and thus ground all concepts correctly.  This question puts us firmly in the context of \textit{identifiability theory}; for \changed{additional background}, please see \cref{sec:related-work}.}

Below we provide a revised perspective on the identifiability of RSs~\citep{marconato2023neuro, marconato2023not, bortolotti2025shortcuts} by clarifying the setting, presenting a first non-identifiability result from \citep{marconato2023neuro} in \cref{thm:likelihood-nec-suff}, and reviewing the characterization of RSs from \citep{marconato2023not,bortolotti2025shortcuts} in \cref{thm:count-rss,thm:identifiability-rss}.  {All results assume the data follows the generating process described in \cref{fig:generative-process} and that we have access to possibly infinite amounts of data}.

\subsubsection{Model class of NeSy predictors}  \label{sec:non-identifiability-of-concepts}

Like in the data generation process, NeSy predictors can be understood as a pair of functions: a concept extractor $f: \calX \to \Delta_\calC$ and an inference layer $\beta: \Delta_\calC \to \Delta_\calY$.
We indicate with $\calF$ the space of learnable concept extractors $f$.
Most theoretical accounts on RSs work in a \textit{non-parametric} setting, whereby $\calF$ contains all possible functions from $\calX$ to $\Delta_\calC$. In words, they assume the neural network implementing the concept extractor can express \textit{any} function $f: \calX \to \Delta_\calC$ \citep{marconato2023not, yang2024analysis, bortolotti2025shortcuts}.  We do the same, but we will relax this assumption when discussing mitigation and awareness strategies in \cref{sec:mitigations} and \cref{sec:awareness}.
Furthermore, we also assume that,  when provided with the prior knowledge $\BK$, the inference layer $\beta$ of a NeSy predictor architecture behaves exactly like the ground-truth inference layer $\beta^*$, \ie
\[
    \beta(\vc) = \beta^*(\vc),%
    \; \forall \vc \in \calC .\footnote{Recall that, for all $\beta \in \calB$, we use the notation $\beta(\vc) := \beta( \Ind{\vC =\vc})$.} 
    \label{eq:condition-on-beta}
\]
We denote with $\calB$ the space of inference layers that satisfy \cref{eq:condition-on-beta}.
The choice of the NeSy predictor architecture (\eg \PNSP or \LTN) determines how the inference layer $\beta$ behaves, as discussed in \cref{sec:nesy-predictors}.
\changed{Notice that, for \LTN, \cref{eq:condition-on-beta} is respected if all fuzzy logic operators are created from a t-norm \citep{van2022analyzing}.}

A specific NeSy predictor thus amounts to a pair $(f, \beta) \in \calF \times \calB$,\footnote{Unless mentioned otherwise, we present the results in the non-parametric setting where $\calF$ contains all possible concept extractors that map inputs to concept conditional probabilities \citep{bortolotti2025shortcuts}. %
} its label {distribution} is given by:
\[
    p_f(\vY \mid \vX; \BK) := (\beta \circ f)(\vX)\, ,
    \label{eq:whatever123}
\]
For this class of models,\footnote{\changed{\cref{eq:whatever123} captures \PNSP, (some versions of) \LTN and \ABL but not the \SL, as the latter does not have separate concept extraction and reasoning stages.
Nevertheless, (deterministic) RSs behave essentially the same in all these models, also in practice, so we employ this equation as a useful abstraction throughout.}} we consider the maximum log-likelihood objective over infinite data:
\[
    {\arg\max}_{f \in \calF}\ \bbE_{(\vx, \vy) \sim p^*(\vX, \vY)} [ \log p_f ( \vy \mid \vx; \BK) ] \, .
\]
{It turns out that this problem may not admit a unique solution}, as explained next.

\subsubsection{Non-identifiability of the concept extractor}
We start by presenting the necessary and sufficient conditions for a NeSy predictor to achieve maximum likelihood on data and then show that this can lead to non-identifiability. Before proceeding, for a probability measure $p^*(\vX)$  and any two measurable functions $r(\vx)$ and $s(\vx)$ with respect to $p^*$, we will use the shorthand
\[
    r(\vX) = s(\vX)
\]
to mean that $r$ takes values equal to  $s$ for $p^*$-almost every $\vx \in \mathrm{supp}(\vX)$.

\begin{tcolorbox}[colback=teal!2!white, colframe=teal!75!black,arc=0pt,outer arc=0pt]
\begin{theorem}[Revisited from \citep{marconato2023neuro}]
\label{thm:likelihood-nec-suff}
    Under \cref{assu:completeness-dgp},
    a NeSy predictor $(f, \beta) \in \calF \times \calB$ attains maximum likelihood with respect to the distribution $p^*(\vX, \vY)$ if and only if%
    \[
        (\beta \circ f) (\vX) = (\beta^* \circ f^*) (\vX)%
        \, .
    \]
\end{theorem}
\end{tcolorbox}

\changed{Unless otherwise stated, proofs for all results can be found in the original papers.}
This immediately yields the following important corollary: when the same label can be predicted using different concept vectors, \textbf{we cannot identify the concept extractor}, \ie $f^* \in \calF$ is not the only choice that leads to obtaining maximum likelihood {for the ground-truth inference layer $\beta^*$}, but there are alternatives $f \neq f^*$.

\begin{tcolorbox}[colback=teal!2!white, colframe=teal!75!black,arc=0pt,outer arc=0pt]

\begin{corollary}[Non-identifiability]
    \label{cor:non-idf-nesy-rss}
    Let $\Delta_\calY^* \subseteq \Delta_\calY$ be the subspace obtained from mapping inputs to labels, which is given by $\Delta^*_\calY := (\beta^* \circ f^*) (\mathrm{supp}(\vX))$, and let $\Delta^*_\calC \subseteq \Delta_\calC$ be the preimage of $\Delta^*_\calY$ over the ground-truth inference layer $\beta^*$, defined as $ \Delta^*_\calC := \{\vp \in \Delta_\calC \mid \,  \beta^*(\vp) \in \Delta_\calY^*   \}$.
    Under \cref{assu:completeness-dgp}, if $\beta^*$ is not injective from $\Delta_\calC^*$ to $\Delta_\calY^*$, then %
    \[
        (\beta^* \circ f) (\vX) = (\beta^* \circ f^*) (\vX) \centernot \implies f(\vX) = f^*(\vX) \, .
    \]
\end{corollary}
\end{tcolorbox}

\begin{proof}
    By construction. If $\beta^*: \Delta^*_\calC \to \Delta^*_\calY$ is not injective, we can find $\vp, \vp' \in \Delta_\calC^*$ such that $\vp \neq \vp'$ and
    \[
        \beta^*(\vp) = \beta^*(\vp') \, . 
        \label{eq:condition-equality}
    \]
    Hence, for all $\vx \in \mathrm{supp}(\vX)$ such that $f^*(\vx) = \vp$ we can construct another function $f \in \calF$ such that $f(\vx) = \vp'$. By \cref{eq:condition-equality}, the NeSy predictor $(f, \beta^*)$ will obtain maximum likelihood (\cref{thm:likelihood-nec-suff}), but $f(\vX) \neq f^*(\vX)$. 
\end{proof}

\cref{cor:non-idf-nesy-rss} formalizes the intuition in \citep{marconato2023neuro} that, when {ground-truth inference layer $\beta^*$ maps multiple concept vectors to the same label, an optimal NeSy predictor can learn an RS.} 
\textbf{\textit{The important consequence is that the concept extractor $f$ attaining maximum likelihood is not unique}}, even when using the ground-truth inference layer $\beta^*$ and an infinite amount of data. 
In other words, \textit{\textbf{{we do not always correctly ground concepts}}}. 
Motivated by these failures modes, we provide a formal definition of \textbf{\textit{shortcut NeSy predictors}} that can be encountered in training:\footnote{This was the original definition of RSs in~\citep{marconato2023not} with unlimited support $\mathrm{supp}(\vX)$. 
Here, we change the original name to specialize it to NeSy predictors and distinguish it from the abstract notion of RSs, as explained in \cref{def:reasoning-shortcut}.} 

\begin{tcolorbox}[
    colback=orange!2!white,
    colframe=orange,
    arc=0pt,
    outer arc=0pt
]
    \begin{definition}[\rechanged{Shortcut NeSy predictor} \citep{marconato2023not}]
    \label{def:faulty-nesy-predictor}
    Consider a data generation process that satisfies \cref{assu:completeness-dgp} and let $(f^*, \beta^*) \in \calF \times \calB$ be the ground-truth concept extractor and inference layer.
    We say that $(f, \beta) \in \calF \times \calB
    $ is a \textbf{\textit{shortcut NeSy predictor}} if $(f, \beta)$ attains maximum likelihood on data (\cref{thm:likelihood-nec-suff}) and 
    \changed{we have
    \[
    f(\vX) \neq f^*(\vX) \, .
    \]}
    \vspace{-1.5em}
    \end{definition}
\end{tcolorbox}

From the definition, the set of shortcut NeSy predictors both depends on the choice of admitted concept extractors $\calF$ and the possible inference layers $\calB$. 
We will show in \cref{sec:mitigations} how, by constraining the space $\calF$ through mitigation strategies, certain shortcut NeSy predictors can vanish. 
From \cref{sec:nesy-predictors}, each NeSy architecture specifies a particular $\beta \in \calB$, which limits shortcut NeSy predictors to concept extractors $f\in \calF$ for which $(f, \beta)$ is faulty. 
This also implies that some concept extractors $f \neq f^*$ may  constitute a shortcut NeSy predictor when paired with the inference layer  of one model class, \eg \PNSP, but not of another, \eg \LTN.

While \cref{cor:non-idf-nesy-rss} establishes {non-identifiability in general}, it does not {provide insight} on {what characterizes \changed{RSs, and in particular} shortcut NeSy predictors}. 
{Without this characterization, it is unclear when the ground-truth concept extractor can be identified and RSs consistently avoided.}
We next describe results that characterize \changed{RSs in the problem, allowing us to identify the ground-truth concept extractor $f^*$.}

\subsubsection{Concept remapping distributions.}
Following \citep{marconato2023not, bortolotti2025shortcuts}, we shift the perspective from describing which concept extractors $f \in \calF$ are valid optima of the likelihood, to studying how ground-truth concept vectors can be remapped by NeSy models while retaining optimal likelihood.
Formally, under \cref{assu:completeness-dgp}, we let $p^*(\vX \mid \vG)$ be the posterior distribution induced by the ground-truth concept extractor $f^*$. 
Then, to describe how concepts learned by a NeSy predictor $(f, \beta) \in \calF \times \calB$ relate to ground-truth concepts, we define 
\[  
    \textstyle
    \alpha_f (\vg) := \bbE_{\vx \sim p^*(\vX \mid \vg)} \left[ 
        f(\vx)
    \right] \,,
    \label{eq:definition-of-alpha}
\]
the \textit{concept remapping distribution} $\alpha_f: \calC \to \Delta_\calC$ induced by $f$~\citep{van2025independenceRS} . 
These distributions $\alpha_f$'s describe how ground-truth concept vectors are mapped to learned concept vectors by $f$. 
Hereafter, we use $\calA$ to denote the space of these $\alpha_f$'s,
which is a simplex \citep{bortolotti2025shortcuts}.
Furthermore, we call $\Vset{\calA}$ the set of \textit{(deterministic) concept remappings}: for each ground-truth vector $\vg \in \calC$, the output of a concept remapping $\alpha_f(\vg)$ is one-hot. 
That is, for such (deterministic) concept remappings $\alpha_f$, there is a single concept vector $\vc \in \calC$ with probability one:    
\[
    \max_{\vc\in \calC} \alpha_f(\vg)_\vc = 1 \, . 
\]
Any element $\alpha_f \in \calA$ can be expressed using convex combinations of concept remappings, $\va \in \Vset{\calA}$, that is:
\[  \textstyle
    \alpha_f (\vg) = \sum_{\va \in \Vset{\calA}} \lambda^f_\va \, \va(\vg),
    \label{eq:convex-combination-rs}
\]
where we have $\lambda^f_\va \geq 0$ for all $\va \in \Vset{\calA}$ and $\sum_{\va \in \Vset{\calA}} \lambda^f_\va = 1$. 
Hence, a NeSy predictor can be seen both as a pair $(f, \beta) \in \calF \times \calB$ and \changed{is associated to} a pair $(\alpha_f,  \beta) \in \calA \times \calB$.
Notice that, because the learnable $\alpha$'s depend on the space of learnable concept extractors $\calF$, constraining $\calF$ also reduces the size of $\calA$. This fact is also explicit in some mitigation strategies (see \cref{sec:mitigation-disentanglement}). %

\changed{
We depict the precise relation between the learnable concept extractors $\calF$ and concept remapping distributions $\calA$ in \cref{fig:relation-extractors-alphas}, where a one-to-one correspondence exists only between concept remappings $\alpha \in \Vset{\calA}$ and functions $f \in \calF$ restricted to the support of $\vX$.\footnote{For the precise relation, refer to {\citep[Lemma 1]{marconato2023not}}.}
Studying concept remapping distributions is especially helpful under \cref{assu:concept-extractor}. 
Then, the concept remapping for the ground-truth %
$f^*$  is the identity function $ \mathrm{id}(\cdot)$ \citep{marconato2023not}, \ie 
\[
    \alpha_{f^*} (\vG) = \mathrm{id}(\vG) .
\]
}

% \newpage 
\subsubsection{Reasoning shortcuts as unintended, optimal concept remapping distributions}

Based on this construction, we can give a formal definition of what a reasoning shortcut is and illustrate it with an example:

\begin{tcolorbox}[
    colback=orange!2!white,
    colframe=orange,
    arc=0pt,
    outer arc=0pt
]
    \begin{definition}[Reasoning Shortcut for $\beta$ (Formal)]
    \label{def:reasoning-shortcut}
    Let $\beta \in \calB$ be the inference layer of a NeSy predictor (\eg \PNSP or \LTN) and $\calF$ the space of learnable concept extractors.
    We say that a concept remapping distribution $\alpha \in \calA$ is a \textbf{\textit{reasoning shortcut for the inference layer}}  $\beta$ %
    if 
    \rechanged{(i) $\alpha \neq \mathrm{id}$, (ii)
    there exists a concept extractor $f \in \calF$ such that
    $\bbE_{\vx \sim p^*(\vX \mid \vG)}[f(\vx)] = \alpha(\vG)$, and (iii) we have
    \[
        (\beta \circ \alpha) (\vg) = \beta^* (\vg), \forall \vg \in \mathrm{supp}(\vG) \, .
    \]
    }
    \vspace{-1.5em}
    \end{definition}
\end{tcolorbox}

With this definition, we formalize reasoning shortcuts as concept remapping distributions that \changed{differ from the identity on at least one ground truth concept $\vg \in \calC$}.\footnote{In contrast to {\citep[Definition 1]{marconato2023not}}, we refer to RSs as the maps $\alpha \in \calA$ for a given $\beta \in \calB$, not as the concept extractors $f \in \calF$ that induce shortcut NeSy predictors. 
}
\changed{There are two reasons why $\alpha$ might not be the identity.  
The first and most obvious is that we are dealing with a shortcut NeSy predictor $(f, \beta)$ whose concept extractor $f$ induces a RS $\alpha_f \ne \mathrm{id}$ in distribution.
The second and more subtle reason is that, even if the NeSy predictor is not faulty, in the sense that it grounds all concepts correctly \textit{in distribution}, the resulting map $\alpha$ may still behave differently from the identity outside of the support $\mathrm{supp}(\vG)$, that is, \textit{out-of-distribution}.}
\changed{
This may happen when the model fails to generalize combinatorially \citep{montero2021role, hwang2023maganet, duan2025state}. 
When the training data includes all combinations of ground-truth concepts, \ie $\mathrm{supp}(\vG) = \calC$, RSs originate exclusively from shortcut NeSy predictors. 
}
The definition of RSs allows us to focus on concept remapping distributions $\alpha \in \calA$ instead of concept extractors, that is, on the symbolic level rather than the sub-symbolic level. 

\begin{figure}[!t]
    \centering
    \includegraphics[width=0.35\textwidth]{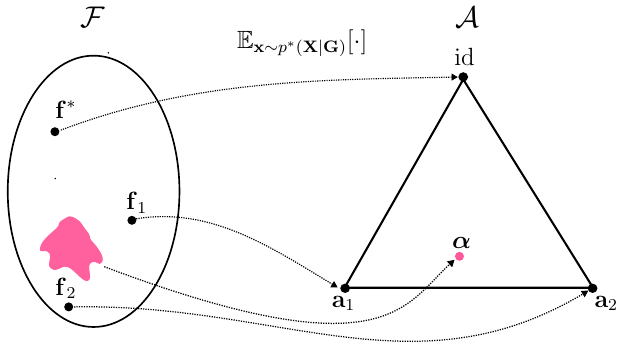}

   \caption{
   We connect concept extractors $f \in \calF$ and maps $\alpha \in \calA$ through \cref{eq:definition-of-alpha}. 
   Here, we represent the space of learnable concept extractors $\calF$ and the space $\calA$ as a two-dimensional simplex. 
   The three vertices correspond to the concept remappings $\mathrm{id}, \va_1, \va_2 \in \Vset{\calA}$. 
   $\mathrm{id}$ is the intended solution, and the remaining mappings are deterministic RSs.
   From \textcolor{teal}{\citep[Lemma 1]{marconato2023not}}, any concept remapping $\va \in \Vset{\calA}$ is in one-to-one correspondence with a concept extractor $f \in \calF$ with domain  restricted to $\mathrm{supp}(\vX)$ and, under \cref{assu:concept-extractor}, $f^*$ is in one-to-one correspondence with the identity function. %
    However, non-deterministic concept remapping distributions $\alpha \in \calA \setminus \Vset{\calA}$ can arise from several concept extractors in $\calF$, here represented by the \textcolor{magenta}{magenta} colored subset mapping to one point in the simplex $\calA$. 
   }
   \label{fig:relation-extractors-alphas}
\end{figure}

Notice that, likewise for shortcut NeSy predictors,  \cref{def:reasoning-shortcut} specializes RSs to the specific inference layer $\beta$ at hand. 
This, in turn, implies that some concept remapping distributions $\alpha$'s can be an RS for one NeSy model class, say \PNSP, but not for another, say \LTN. 
This may happen when two inference layers $\beta, \beta' \in \calB$ differ in behavior when mapping non-deterministic distributions $p(\vC \mid \vx) \in \Delta_\calC \setminus \Vset{\Delta_\calC}$. 
This distinction is particularly evident between probabilistic logic approaches (which essentially implement integration through marginalization, like \PNSP and \SL) and fuzzy logic methods (where the inference layer depends on the fuzzy logic chosen). 
However, all \textit{deterministic} concept remappings $\alpha \in \Vset{\calA}$ that are RSs, are RSs for \textit{all} inference layers. 
This is because, by \cref{eq:condition-on-beta}, they output the same values on one-hot concept distributions $\Vset{\Delta_\calC}$ (see also \citep[Appendix A]{marconato2023not}).
We will refer to these RSs as \textbf{\textit{deterministic RSs}}.

% \vspace{-2em}

\begin{tcolorbox}[
    colback=orange!2!white,
    colframe=orange,
    arc=0pt,
    outer arc=0pt
]
\begin{definition}[Deterministic Reasoning Shortcut]
    \label{def:deterministc-rss}
    We say that $\alpha \in \Vset{\calA}$ is a \textbf{deterministic reasoning shortcut} if $\alpha$ is a reasoning shortcut for $\beta^* \in \calB$ (\cref{def:reasoning-shortcut}).  
\end{definition}
\end{tcolorbox}

\newpage

\begin{tcolorbox}[
    colback=gray!2!white,
    colframe=gray,
    arc=0pt,
    outer arc=0pt
]
\begin{example}
\label{ex:rss-bdd-oia-extended}
    Consider a simplified version of
    \;\BOIA \citep{xu2020boia} {with label} $y \in \{\tt go, stop \} = \calY$, three binary concepts $(C_{\tt green}, C_{\tt red}, C_{\tt ped})$ with domain $\calC$, and prior knowledge $\BK = (C_{\tt ped} \lor C_{\tt red} \Leftrightarrow (Y = \tt stop)) \land (C_{\tt green} \land \lnot(Y = {\tt stop}) \Leftrightarrow (Y = \tt go))$.
    In \cref{fig:example-rss}, we visualize the four concept remapping distributions appearing in \cref{fig:relation-extractors-alphas}. 
    The concept remapping distributions $\alpha \in \calA$ determine how ground-truth concepts are mapped to conditional probability distributions on model concepts. 
    First, we can find an RS ($\va_1$ in \cref{fig:example-rss} left-center) that predicts $C_{\tt ped} = 1$ with probability one whenever $G_{\tt red}=1$ or $G_{\tt ped}=1$. 
    Similarly, we have an RS ($\va_2$ in \cref{fig:example-rss} right-center) that predicts $C_{\tt red} = 1$ with probability one whenever $G_{\tt red}=1$ or $G_{\tt ped}=1$. 
    A nondeterministic RS ($\alpha$ in the rightmost part of \cref{fig:example-rss}) assigns non-zero probability to both $C_{\tt red} = 1$ and $C_{\tt ped} = 1$. 
\end{example}

\begin{minipage}{\linewidth}

\centering

\includegraphics[width=0.8\linewidth]{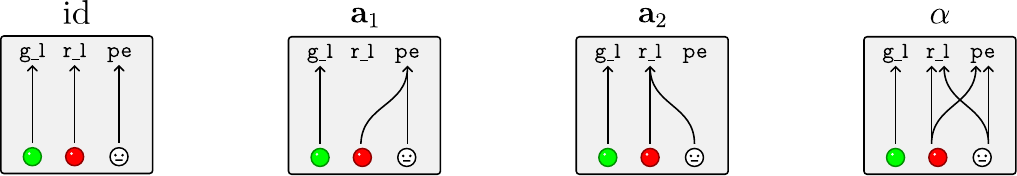}

\captionof{figure}{
The four concept remapping distributions appearing in \cref{fig:relation-extractors-alphas} according to \cref{ex:rss-bdd-oia-extended}. 
    To aid visualization, we display the concept remapping distributions as directed graphs where arrows denote how ground-truth concepts are mapped to the predicted concepts.
    On the left, $\mathrm{id}$ corresponds to the ground-truth map $\alpha_{f^*}$ from the data generation process. 
    At the center, $\va_1, \va_2 \in \Vset{\calA}$ represent two deterministic RSs. 
    On the right, $\alpha \in \calA \setminus \Vset{\calA}$ is a non-deterministic RS. 
    Among these RSs, $C_{\tt green}$ is always predicted correctly, whereas $C_{\tt red}$ and $C_{\tt ped}$ are incorrectly predicted.
}
\label{fig:example-rss}
\end{minipage}
\end{tcolorbox}

\subsubsection{Characterization of deterministic reasoning shortcuts.}
The description using maps $\alpha \in \calA$ is useful because we can count the number of deterministic RSs explicitly. 
As noted above, under \cref{assu:concept-extractor}, the ground-truth concept remapping $\alpha_{f^*}$ is the identity $\mathrm{id}(\cdot)$. Intuitively, any learnable $\alpha \in \Vset{\calA}$ that differs from the identity function is a deterministic RS. 
The following result makes the number of deterministic RSs explicit:

\begin{tcolorbox}[
    colback=teal!2!white,
    colframe=teal,
    arc=0pt,
    outer arc=0pt
]
    \begin{theorem}[Number of deterministic reasoning shortcuts \citep{marconato2023not}]
    \label{thm:count-rss}
        Let $\calF$ be a space of concept extractors. Under \cref{assu:completeness-dgp,assu:concept-extractor,assu:inference-layer}, the number of {deterministic} RSs (\cref{def:reasoning-shortcut}) is
        \[
            \textstyle
            \sum_{\alpha \in \Vset{\calA_\calF}} \Ind{
                \bigwedge_{\vg \in \mathrm{supp}(\vG)}
                    (\beta^* \circ \alpha)(\vg) = \beta^*(\vg)
            } - 1 \, ,
            \label{eq:count-rss}
        \]
        where $\calA_\calF := \{ \alpha \in \calA : \exists\ f \in \calF \text{ s.t. } \alpha = \alpha_f
        \}$.
    \end{theorem} 
\end{tcolorbox}

{When this count is greater than $1$, the NeSy task admits (deterministic) RSs. A NeSy predictor can attain high label accuracy by learning any one of them or any (non-deterministic) mixture thereof.}
{Because the count does not depend on the particular choice of $\beta \in \calB$, \textbf{\textit{all NeSy architectures have the same deterministic reasoning shortcuts}} \citep{marconato2023not}.} %
This happens because all NeSy inference layers $\beta \in \calB$ integrating the prior knowledge $\BK$ satisfy \cref{eq:condition-on-beta}. 

When $\Vset{\calA_\calF}$ contains all maps, \ie $\Vset{\calA_\calF} = \Vset{\calA}$, and $\mathrm{supp} (\vG)$ is complete, \ie $\mathrm{supp} (\vG) = \calC$, it is possible to explicitly count the number of deterministic reasoning shortcuts in the task, see \citep[Appendix C]{marconato2023not}.%
The number of deterministic RSs rapidly explodes based on how many concept vectors correspond to a single label through $\beta$. 
For example, if we can predict a label $\vy_0$ from $M$ different concept vectors, the number multiplies by a factor of $M^M$. 
Accounting for practical restrictions in the set $\Vset{\calA_\calF}\subseteq \Vset{\calA}$ due to, \eg the choice of specific neural backbones, complicates the explicit counting of RSs, but this can be done using model counting, see \cref{sec:rscount}. \changed{We illustrate next how \cref{eq:count-rss} depends on the quantities inside the sum, but defer the in-depth treatment of causes and mitigations of RSs to \cref{sec:mitigations}.} 

\begin{tcolorbox}[
    colback=gray!2!white,
    colframe=gray,
    arc=0pt,
    outer arc=0pt
]
\begin{example}[%
\changed{Deterministic Reasoning Shortcuts in XOR~\citep{marconato2023not, van2025independenceRS}}%
]

    Consider two binary concepts $G_1, G_2 \in \{0,1\}$ that both generate an input $\vx \sim p(\vX \mid G_1, G_2)$, which can be though as the image visualization of the values (\eg $\vg = (1,0)$ corresponds to $\vx=\MOne \MZero$). 
    The label $Y$ results from the XOR operation between the binary concepts: $p^*(Y \mid \vG; \BK) = \Ind{Y = G_1 \oplus G_2}$. 
    The deterministic maps in $ \Vset{\calA}$ can be seen as all maps $\alpha: \{ 0, 1\}^2 \to \{0,1\}^2 $, and so $|\Vset{\calA}| = 4^4$. 
    We analyze how the RS count in \cref{eq:count-rss} varies when: (\textbf{1}) the support of ground-truth concepts is complete and all deterministic $\alpha$'s are allowed;  (\textbf{2}) we impose a constraint on deterministic $\alpha$'s.

    \vspace{0.5em}
    
    \textbf{Case 1}.
    When $\calA_\calF = \calA$ and the support of ground-truth concepts in the training data is exhaustive, \ie $\supp(\vG) = \{0,1\}^2$, we obtain a total of $2^2 \cdot 2^2 - 1$ deterministic RSs. To show this, we construct one where
    \[
        \alpha(0,0) = (0,0), \; \; 
        \alpha(0,1) = (0,1), \; \;
        \alpha(1,0) = (0,1), \; \;
        \alpha(1,1) = (0,0) \, .
    \]
    This shows that, for $b_0, b_1 \in \{0,1\}$, we can choose between mapping both $\{(0,0), (1,1)\} \mapsto (b_0,b_0)$ and $\{(0,1), (1,0)\} \mapsto (1-b_1,b_1)$ so that we get valid label predictions $(\beta^* \circ \alpha)(G_1, G_2) = \beta(G_1, G_2) $. Another deterministic RS flips the values of the bits, resulting in
    \[
        \alpha(0,0) = (1,1), \; \; 
        \alpha(0,1) = (1,0), \; \;
        \alpha(1,0) = (0,1), \; \;
        \alpha(1,1) = (0,0) \, .
    \]
    Denote with $\calC_0 = \{(0,0), (1,1)\}$ and  $\calC_1 = \{(0,1), (1,0)\}$.
    We consider maps $\alpha$ where for all $\vc\in \calC_0$, $\alpha(\vc) \in \calC_0$, so there are two valid options for each element of $\calC_0$. 
    The same holds for $\calC_1$. 
    Now each map $\alpha$ that satisfies these constraints is a valid map. 
    There are then $|\calC_0|^{|\calC_0|} \cdot |\calC_1|^{|\calC_1|}=2^2\cdot 2^2=16$ valid $\alpha$'s, and all of them are deterministic RSs except the identity.

    \vspace{0.5em}

    \textbf{Case 2}.
    We consider full support of ground-truth concepts in the training data, \ie $\supp(\vG) = \{0,1\}^2$, but limit the maps $\alpha$'s to be factorized, that is $\alpha(G_1,G_2) = (\hat \alpha (G_1), \hat \alpha (G_2))$. This means that $\Vset{\calA_\calF} \subset \Vset{\calA}$, and in total there are $2^2 \cdot 2^2$ deterministic $\alpha$'s in $\Vset{\calA_\calF}$.\footnote{%
    \changed{We later discuss in \cref{sec:rss-as-statistical-learning} and in \cref{sec:mitigation-disentanglement} that this can result from concept extractors being \textit{architecturally disentangled}.}%
    }
    Because each $\hat \alpha$ depends only on the binary value of the $i$-th concept,
    \ie $\hat \alpha:\{0,1\} \to \{0,1\}$,
    we only have one deterministic RS resulting from $\hat \alpha(b) = 1 -b$ for $b \in \{0,1\}$. This gives the map 
    \[
    \begin{aligned}
        &\alpha(0,0) = (\hat \alpha(0),\hat \alpha(0)) = (1,1), \; \; 
         \alpha(0,1) = (\hat \alpha(0),\hat \alpha(1)) = (1,0), \; \; \\
        &\alpha(1,0) = (\hat \alpha(1),\hat \alpha(0)) = (0,1), \; \; 
         \alpha(1,1) = (\hat \alpha(1),\hat \alpha(1)) = (0,0) \, . 
    \end{aligned}
    \]
    In fact,
    one can check that other combinations with $\hat \alpha(G_i) = b$ with $b$ a binary constant, are not valid $\alpha$'s.
    
\end{example}
\end{tcolorbox}

\cref{thm:count-rss} only counts \textit{deterministic RSs}. 
If the count is reduced to zero, then NeSy predictors may still suffer from \textit{non-deterministic} RSs (those $\alpha$'s  in $\calA \setminus \Vset{\calA_\calF}$). 
To ensure that we can deal with \textit{all} RSs by avoiding deterministic RSs, we need to introduce an additional assumption on the inference layer $\beta$:

\begin{tcolorbox}[
    colback=violet!2!white,
    colframe=violet,
    arc=0pt,
    outer arc=0pt
]
\begin{assumption}[Extremality \citep{bortolotti2025shortcuts}]
    \label{assu:monotonic}
    The inference layer $\beta \in \calB$ satisfies extremality if, for all $\lambda \in (0,1)$ and for all $\vc \neq \vc' \in \calC$ such that %
    $\argmax_{\vy \in \calY} \beta (\vc)_\vy \neq \argmax_{\vy \in \calY} \beta(\vc')_\vy$,
    where $\beta(\cdot)_\vy$ is the $\vy$-th component of $\beta$,
    we have
    \[
        \max_{\vy \in \calY} \beta(  \lambda \Ind{\vC = \vc} + (1- \lambda) \Ind{\vC = \vc'}  )_\vy 
        < 
        \max \left( 
        \max_{\vy \in \calY} \beta (\vc)_\vy , \max_{\vy \in \calY} \beta (\vc')_\vy
        \right) \, .
        \label{eq:max-condition}
        \nonumber
    \]
\end{assumption}
\end{tcolorbox}

\begin{figure}[!t]
    \centering
    \includegraphics[width=0.5\linewidth]{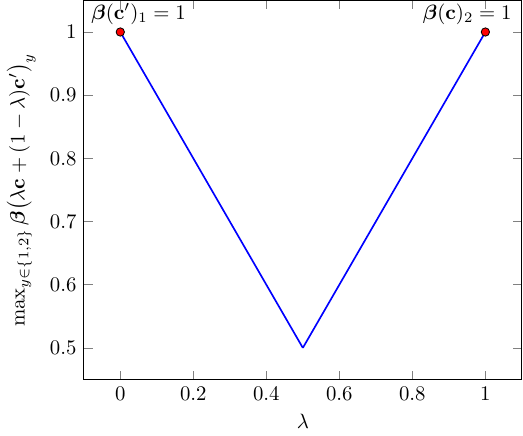}
    \caption{The extremality condition holds for \PNSP and \SL. %
    }
    \label{fig:extremality-DPL-SL}
\end{figure}

A NeSy predictor satisfies \cref{assu:monotonic} if its inference layer $\beta$ is ``peaked'':  for any two concept vectors $\vc$ and $\vc'$ with different predictions, the label probability given by $\beta$ for a \textit{mixture} of these predictions is no larger than the label probability it associates to $\vc$ and $\vc'$.
This happens in \PNSP on \MNISTAdd, where the knowledge specifies that $\beta: (\vC =(0,1)^\top) \mapsto (Y=1)$ and $\beta: (\vC=(0,2)^\top) \mapsto (Y=2)$, both with probability one. Any convex combination crossing \rechanged{$\vc'=(0,1)^\top$} to \rechanged{$\vc = (0,2)^\top$} would not result in any label with probability one, see \cref{fig:extremality-DPL-SL}. 
E.g., \PNSP will assign a higher probability to the sum $Y=1$ when $\lambda < 0.5$. 
In essence, distributing probability mass across different sets of concepts does not increase label likelihood. %
Extremality holds for \PNSP and \SL, \changed{and holds for \LTN when using fuzzy logic operators based on t-norms and t-conorms~\citep{van2022analyzing}}. %
\changed{An investigation for which models \cref{assu:monotonic} holds is reported in \citep[Appendix E]{bortolotti2025shortcuts}. }

Under \cref{assu:monotonic}, the next result holds:

\begin{tcolorbox}[
    colback=teal!2!white,
    colframe=teal,
    arc=0pt,
    outer arc=0pt
]
    \begin{theorem}[Identifiability \citep{bortolotti2025shortcuts}]
    \label{thm:identifiability-rss}

    Under \cref{assu:completeness-dgp,assu:concept-extractor,assu:inference-layer} and for all $\beta \in \calB$ satisfying \cref{assu:monotonic},
    \changed{if the count of {deterministic} RSs is zero (\cref{eq:count-rss}), then}
    any NeSy predictor $(f, \beta) \in \calF \times \calB$ attaining maximum likelihood (\cref{thm:likelihood-nec-suff}) 
    \changed{has $\alpha_f(\vG) = \mathrm{id}(\vG)$} and so
    finds the ground-truth concept extractor, \ie $f(\vX) = f^*(\vX)$.
    \end{theorem}
\end{tcolorbox}

\changed{
This result highlights that \textit{extremality reduces the space of equivalent optima} by preventing different concept assignments from being indistinguishable at the label level.
It 
indicates we can mitigate RSs for all NeSy predictors that obey \cref{assu:monotonic}.} 
In fact, zeroing the count on deterministic RSs (\cref{thm:count-rss}) implies avoiding all RSs in the learning task \changed{and, in particular, avoiding shortcut NeSy predictors}.
For NeSy predictors model classes obeying \cref{assu:monotonic}, the identifiability result gives that, in the absence of deterministic RSs, \textbf{\textit{maximizing the likelihood on labels permits the correct grounding of the concepts}}.
This, in turn, presupposes finding additional mitigations that, paired with maximum-likelihood \cref{eq:dpl-learning-objective}, guarantee the  \textit{updated count} of deterministic reasoning shortcuts equals zero. We will discuss next (\cref{sec:mitigations}) how different mitigations act on the count reduction.

\begin{remark}
    As a corollary of \cref{thm:identifiability-rss},  
    for probabilistic logic methods (including \PNSP like \DPL \citep{manhaeve2018deepproblog}, \SL \citep{xu2018semantic}, and \texttt{SPL} \citep{ahmed2022semantic}), all non-deterministic RSs can be constructed from convex combinations of deterministic RSs {\citep[Proposition 3]{marconato2023not}}. This guarantees that if there are no deterministic RSs, then there are also no other RSs. 
    Moreover, as will be discussed in \cref{sec:awareness}, it is possible to leverage this geometrical aspect to construct non-deterministic RSs from convex combinations of deterministic ones, whose uncertainty on (some of) the learned concepts reveals which of them are affected by RSs.
\end{remark}

\subsection{Perspective: Statistical Learning}
\label{sec:rss-as-statistical-learning}

Learning a NeSy model amounts to learning (the parameters of) a concept extractor that---once combined with the inference layer---attains high label likelihood.  From a statistical learning perspective, this is equivalent to finding a model with minimal empirical \textit{label} risk. Statistical learning theory allows one to bound the true risk of a model---that is, to assess how well it generalizes---based on its observed empirical risk and other factors, such as training set size and model complexity \changed{\citep{valiant1984theory, vapnik2013nature, understanding-ml}.}
Since the inference layer of a NeSy model is fixed and supplied upfront via the background knowledge $\BK$, it is natural to ask when and how minimizing the \textit{label} risk causes a reduction of the \textit{concept} risk. Or similarly, when does low risk on label predictions bound the mismatch between the predicted and the ground-truth concepts?

We expect that if RSs are present, such a bound could be vacuous for some concepts.  In fact, from the results in the previous section, we know that the concepts predicted by a shortcut NeSy predictor can drastically differ from ground-truth ones, even when the model is trained with an infinite amount of data.
To make this explicit, we follow the analysis of \citep{yang2024analysis}, which indicates from a statistical learning perspective the relation between label and concept risks.
\changed{The subsequent study of \citet{he2025learnability} follows similar steps. %
}

\subsubsection{Knowledge complexity}

The statistical learning perspective on RSs follows steps similar to identifiability analysis (\cref{sec:rss-as-identifiability}).
First, data are assumed to be generated as per \cref{assu:completeness-dgp}, whereby the ground-truth concepts and labels are {sampled from} the ground-truth concept extractor $f^*: \calX \to \Delta_\calC$ and inference function $\beta^*:\Delta_\calC \to \Delta_\calY$. From the joint distribution $p^*(\vX, \vG, \vY)$ of \cref{eq:joint-input-label-concept-distr}, it is possible to determine the marginal distribution over labels $p^*(\vY)$. This marginal distribution is used together with the knowledge $\BK$ to determine the complexity of the knowledge:

\begin{tcolorbox}[colback=orange!2!white, colframe=orange,arc=0pt,outer arc=0pt]
\begin{definition}[Knowledge complexity \citep{yang2024analysis}]
    \label{def:knowledge-complexity}
    The knowledge complexity of $\BK$ for labels distributed according to the label distribution $p^*(\vY)$ (\cref{assu:completeness-dgp}) is defined as
    \[
        \mathrm{KC}(\BK; p^*) := \bbE_{\vy \sim p^*(\vY)} [ \sum_{\vc \in \calC} \Ind{(\vc, \vy) \not \models \BK  } ] \, .
    \]
\end{definition}
\end{tcolorbox}
This quantity gives a global description of how many concepts are not compatible with the knowledge, based on the labels $\vy \in \mathrm{supp}(\vY)$ observed during training. Intuitively, \changed{when only few concepts allow inferring any given label vector},
the knowledge complexity increases, indicating that the knowledge is more explicit about what concept vectors lead to that label vector. On the other hand, a low knowledge complexity indicates that many concept vectors are mapped to same label vector. 

\changed{For instance, in the \BOIA example (\cref{ex:boia}) with three binary concepts ($|\calC| = 8$), \texttt{go} can only be inferred from $C_{green} = 1$, $C_{ped}= 0$, and $C_{red}=0$, meaning that $1$ out of $8$ concept vectors $\vc$ entail {\tt go} and $7$ out of $8$ entail {\tt stop}.}
Hence, whenever both \texttt{go} and \texttt{stop} instances are observed, we get $\mathrm{KC}(\BK; p^*) < |\calC|- 1 =7$. 
Notice that, by construction, the complexity of any knowledge $\BK$ on any distribution $p^*(\vY)$ over labels is upper-bounded by
\[
    \mathrm{KC}(\BK; p^*) \leq |\calC|- 1 \, .\footnote{
    \changed{
    Notice that, for corner cases in which observed labels in $\mathrm{supp(\vY)}$ cannot be predicted from the concepts, \ie $(\vc, \vy) \not \models \BK$ for all $\vc \in \calC$, we get KC$(\BK; p^*) = |\calC|$. We systematically avoid it through \cref{assu:completeness-dgp}, which ensures that concepts are predictive of the observed labels. 
    }
    }
\]
Next, we show how the complexity of the knowledge is intimately connected to the emergence of RSs.

\subsubsection{Impossibility result in bounding the reasoning shortcut risk}
We start by recalling the relevant notation introduced in \cref{sec:rss-as-identifiability}.
For a NeSy predictor $(f, \beta) \in \calF \times \calB$, the corresponding concept and label distributions are given by
\[
    p_f (\vC \mid \vX) := f(\vX), \quad  
    p_f(\vY \mid \vX; \BK) := (\beta \circ f)(\vX) \, . 
\]
We assume, as before, that the space of learnable concept extractors $\calF$ is unrestricted and that the inference layer $\beta$ matches the ground-truth inference layer at the vertices (\cref{eq:condition-on-beta}).
The expected label risk and expected concept risk are given by
\[
   \calR_\vY(f) :=  - \bbE_{(\vx, \vy) \sim p^*(\vX, \vY)} [ \log p_f(\vy \mid \vx; \BK)], 
   \quad 
   \calR_\vC(f) := -\bbE_{(\vx, \vg) \sim p^*(\vX, \vG)} [ \log p_f( \vC=\vg \mid \vx) ] \, ,
\]
respectively.
Their difference gives the expected \textit{reasoning shortcut risk}~\citep{yang2024analysis}:
\[
    \calR_{RS}(f) := \calR_\vC (f) - \calR_\vY (f) \, .
    \label{eq:rss-risk-expected}
\]
Notice that when the RS risk is positive, concept risk exceeds label risk; hence, despite more accurate label predictions, concepts are less accurately predicted. Conversely, a concept risk less than or equal to zero indicates that concept predictions are at least as accurate as label predictions.
The next result connects RSs to the \textit{knowledge complexity} showing that, when it is not high enough, an analogous result to \cref{cor:non-idf-nesy-rss} holds:

\begin{tcolorbox}[
    colback=teal!2!white,
    colframe=teal,
    arc=0pt,
    outer arc=0pt
]
\begin{theorem}[Unbounded RSs risk \citep{yang2024analysis}]
    \label{thm:non-idf-risk-rss}
    Under \cref{assu:completeness-dgp},
    if {the knowledge complexity is bounded by the size of the concept space as} $\mathrm{KC}(\BK; p^*) < |\calC| - 1$, then there is a concept extractor $f$ such that $\calR_\vY = 0$ and $\calR_\vC \to \infty$, so that the RS risk approaches infinity, \ie
    \[
        \calR_{RS} \to \infty \, .
    \]
\end{theorem}
\end{tcolorbox}

\changed{\cref{thm:non-idf-risk-rss} shows that minimizing the label loss alone does not  appropriately guide concept learning. 
If the inference layer implementing knowledge $\BK$ maps many concept vectors to the same label, a model can achieve perfect label accuracy while predicting completely wrong concepts. In this case the concept risk can grow arbitrarily large even though the label risk is zero.}
Notice that the main causes for this behavior are that: 
(1) the knowledge complexity
\changed{is not maximal},
and therefore each label $\vy \in \mathrm{supp}(\vY)$ is associated to multiple concept vectors $\vc \in \calC$, 
(2) the fact that the space of learnable concept extractors is unrestricted and can express all possible conditional concept distributions. 
\changed{Both are natural targets for mitigation.  For instance, one can increase knowledge complexity by introducing more specialized constraints into the knowledge, whenever feasible.  Alternatively, one can shrink the space of learnable concept extractors $\calF$ by constraining the extractor's architecture.} %
\changed{We will discuss these and strategies further in \cref{sec:mitigations}.}

\paragraph{Relation to \citep{he2025learnability}} 
\rechanged{\citet{he2025learnability} provide a complementary learnability analysis of neuro-symbolic learning through the notion of a derived constraint satisfaction problem (DCSP). Their identification result (ref. to \citep[Lemma 4.7]{he2025learnability} shows that a NeSy task is learnable if and only if the associated DCSP admits a unique solution; 
when multiple solutions satisfy the constraints, the ground-truth concepts cannot be identified from label supervision alone, even with infinite data. This procedure resembles the RSs count in \cref{eq:count-rss}, although their enumeration focuses only on those RSs that affect in-distribution concept risk. 
Moreover, for general NeSy tasks they show that the asymptotic concept error is controlled by the disagreement among DCSP solutions. 
From the perspective of reasoning shortcuts, while \citet{yang2024analysis} emphasize the mismatch between label risk and concept risk, \citet{he2025learnability} characterize this phenomenon in terms of the solutions to the DCSP brought by the knowledge and derive corresponding learnability guarantees.}
\changed{Notice that the statistical view allows to derive bounds on the reasoning shortcut risk, which has been done for the case of concept smoothing and pretraining of the concept extractor. Overall, the goal is to provide guarantees about avoiding  learning shortcut NeSy predictors.  We refer the reader to \citep{yang2024analysis,he2025learnability} for more details on these bounds.}
\rechanged{A closely related constraint-based perspective is developed by~\citet{takemura2026constraint}, who  formalize reasoning shortcuts as a constraint satisfaction problem and ask when the constraints uniquely determine the intended concept mapping. They identify a \emph{discrimination property} --- no valid mapping can be turned into another by swapping two concept values --- which is necessary but not sufficient for shortcut-freeness.}

\subsubsection{PAC learnability with factorized concept extractors}

\changed{
Note that \cref{thm:non-idf-risk-rss} implicitly assumes the hypothesis space $\calF$ is very expressive.  A natural question is whether this negative result can be worked around if we restrict $\calF$.  In many NeSy tasks, inputs naturally decompose into independent units (e.g., digits in \MNISTAdd \citep{manhaeve2018deepproblog}). 
This motivates studying factorized concept extractors, where each input component is processed independently. %
}

The work by \citet{wang2023learning} studies one special case which is often available in NeSy tasks: when the input can be naturally decomposed into a tuple of independent units, \eg the digits in \MNISTAdd, and the same concept extractor can be employed to predict the concepts of each input instance, \eg the value of each digit is obtained by mapping the input alone, what conditions are required to avoid RSs and guarantee correct learning of the concept extractor? 
This setting, which builds on factorized concept extractors of the form $f = \bar f_1 \times \cdots \times \bar f_k$, has been studied via the \textit{probably approximate correct} (PAC) framework \changed{\citep{valiant1984theory, understanding-ml}}, and connects the \textit{learnability} of NeSy predictors to that of weakly-supervised models \citep{relaxed-supervision,pmlr-v48-raghunathan16}.

\subsubsection{Setting}
\changed{Following \citep{wang2023learning}, we consider training samples $(\vx_i, y_i) \in \calX \times \calY$ with a single categorical label $y$ (instead of a vector label $\vy$ used in the previous sections) and where each input $\vx$ is a sequence of $k$ vectors ${\vx = (\bar \vx_1, \ldots, \bar \vx_k)} \in \calX=\bar \calX^k$, with shared domain $\bar \calX \subseteq \bbR^n$.  Similarly, the concept vectors $\vc =(c_1, \ldots, c_k) \in \calC$ concatenates $k$ copies of the same concept $\bar \calC$, one for each element of the input, with overall domain $\calC = \bar \calC^k$.  The simplest way to visualize this setup is to think of \MNISTAdd (\cref{ex:mnist-rss-full}): the elements $\bar \vx_i$ would be the individual \MNIST digits, $\bar C$ any digit, and $y$ their sum.}

\changed{Next, we have to fix and construct the family of learnable concept extractors $f \in \calF$.  We do so by imposing that $f$ extracts the $k$ concepts by applying the same function $\bar f : \bar\calX \to \Delta_{\bar\calC} \in \bar\calF$ to each of them in turn, that is,
\[
    f(\vx) = (\bar f(\bar \vx_1), \ldots, \bar f(\bar\vx_k)) ,
    \label{eq:disentangled-f-definition}
\]
In \MNISTAdd, this would equate to processing each digit \textit{independently from the others} with the same function $\bar f$.
Overall, this construction implies that $\calF$ \textit{factorizes} as $\calF = \bar\calF^k$ or, equivalently, that $f$ is \textit{architecturally disentangled}, as we will discuss in \cref{sec:mitigation-disentanglement}.  This is in contrast with \cref{cor:non-idf-nesy-rss,thm:non-idf-risk-rss}, which neither assume the concept extractor is architecturally disentangled nor that the predicted concepts are independent.}

\changed{Recall that $f$ and $\bar f$ output \textit{distributions} over concept values, \eg distributions over digits.  We need an easy way to refer to their \textit{most likely values}, \eg the predicted digits themselves.  For this purpose, we introduce the shorthand $[\bar f] (\bar \vx) := \argmax_{c \in \bar \calC} \bar f(\bar \vx)_c $ to denote their predictions.  By construction, we have that $[\bar f]: \bar\calX \to \bar\calC$, and we denote with $[\bar \calF]$ the space of these functions.}

\changed{As customary in statistical learning, we are interested in bounding the \textit{risk} of the model.  \citet{wang2023learning} do so for the risk induced by the zero-one losses over labels and concepts, which we introduce next.
For the concepts, the \textit{zero-one loss} is $\ell^{01}_\vC(c,c') := \Ind{c \neq c'}$ for any two values $c, c'\in\bar \calC$ and therefore the expected concept risk is:
\[ \textstyle
    \calR^{01}_\vC(f) := \bbE_{(\vx, \vg) \sim p^*(\vX, \vG)} \left[ 
        \frac{1}{k} \sum_{j=1}^k \ell^{01}_\vC ([\bar f](\bar \vx_j), g_j)
    \right] .
\]
Here, $p^*(\vX, \vG)$ is the ground-truth distribution of inputs and concept vectors.
For the labels, the zero-one loss (also known as \textit{zero-one partial loss} \citep{wang2023learning}) is simply $\ell^{01}_{\beta^*}(\vc, y) := \Ind{\beta^*(\vc) \neq y }$, where$\beta^* \in \calB$ is the ground-truth inference layer.  It follows that the expected label risk is:
\[
    \calR^{01}_\vY(f, \beta^*) :=  
     \bbE_{(\vx, y) \sim p^*(\vX, Y)} \left[ 
     \ell_{\beta^*}^{01}\big(
     ([\bar f](\bar \vx_1), \ldots, [\bar f](\bar \vx_k)), y 
     \big)
     \right] .
\]
Here, $p^*(\vX, Y)$ is the ground-truth distribution that the training data $(\vx_i, y_i) \in \calX \times \calY$ is sampled from.  (This setup is reminiscent of \cref{sec:rss-as-statistical-learning}.)}

\changed{
We say that $\mathcal{F}$ is \emph{realizable} when there exists a concept extractor ${f^* \in \mathcal{F}}$ that achieves zero risk, that is, ${\calR}^{01}_\vY(f^*;\beta^*)=0$.
As customary in \textit{empirical risk minimization}, we train the concept extractor using a training set $\calD_{p^*}$ with $m_{p^*}$ many examples $(\vx, y)$ sampled from $p^*(\vX, Y)$.
Hence, the associated \textit{empirical risk} is:
\[
    \hat{\calR}^{01}_\vY(f, \beta^*,\calD_{p^*})  := \frac{1}{m_{p^*}} \sum_{(\vx, y) \in \calD_{p^*}} \ell_{\beta^*}^{01}(([\bar f](\bar x_1), \dots, [\bar f](\bar x_k)),y) .
\]
PAC learnability of this NeSy task \citep{understanding-ml, wang2023learning} asks whether there exists a large enough training set that allows one to learn a concept extractor with low enough concept risk with high enough probability.  Formally, we ask if, for any distribution $p^*(\vX, Y)$ and any $\epsilon$ and $\delta \in (0,1)$, the \textit{zero-one concept risk} of the empirical risk minimizer $f$ is less than or equal to $\epsilon$ (\ie $\calR_\vC^{01}(f) \leq \epsilon$) with probability at least $1 - \delta$ when the number of training examples $m_{p^*}$ is at least $m_{\epsilon, \delta}$.
Next, we present how \citet{wang2023learning} derive such a $m_{p^*}$ for this learning setting.}

\subsubsection{PAC learnability with $k$-unambiguous inference layers} \label{sec:slr-pac-learning:learnability-known-classifier}

\changed{
To prove the PAC learnability of NeSy tasks, we must bound the concept risk ${ \calR_\vC^{01}(f) }$ by the label risk ${ \calR^{01}_\vY (f;\beta^*) }$ for any distribution of the training data. Below, we report the conditions that $\beta^*$ has to satisfy to ensure that this is the case.
We already know that if the $\beta^*$ allows to infer the same label from multiple concept vectors, then label supervision is not enough to recover the ground-truth concepts because the model cannot distinguish between correct and incorrect concepts based on labels alone.  This holds even if the concept extractor is factorized (\cref{eq:disentangled-f-definition}). %
The next condition formalizes when the inference layer avoids this ambiguity
}\rechanged{ by focusing on diagonal vectors $\vc = (c, \ldots, c)$:}

\begin{tcolorbox}[colback=orange!2!white, colframe=orange,arc=0pt,outer arc=0pt]
\begin{definition}[$k$-unambiguity \citep{wang2023learning}] \label{definition:k-unambiguous-mapping}
    \changed{For $k$-dimensional concept vectors $\vc \in \calC$, }
    an inference layer $\beta^*$ is \textbf{$k$-unambiguous} if  
    for any two \rechanged{diagonal} concept vectors $\vc = (c,\dots,c)$ and $\vc' = (c',\dots,c')$ $\in \calC$, such that $c \ne c'$, 
    we have that ${\beta^*(\vc') \neq \beta^*(\vc)}$. Otherwise, the inference layer $\beta^*$ is \textbf{$k$-ambiguous}.
\end{definition}
\end{tcolorbox}

{
This definition essentially guarantees that 
passing \rechanged{diagonal} vectors of the form $\vc = c (1, \ldots, 1)^\top$ and $\vc' = c' (1, \ldots, 1)^\top$ to $\beta^*$ never returns the same label predictions for $c \neq c'$, guaranteeing the injectivity of $\beta^*$ for these pairs.   
Although \cref{definition:k-unambiguous-mapping} might seem strict, NeSy tasks like \MNISTAdd satisfy this condition, see \cref{ex:mnist-rss-full}. 
Moreover, it is possible to extend $k$-unambiguity to the case where $\beta^*$ is non-deterministic and to establish the relationship with small ambiguity degree from partial label learning \citep{learnability-pll}.  %
With this condition, the following result holds:
}

\begin{tcolorbox}[
    colback=teal!2!white,
    colframe=teal,
    arc=0pt,
    outer arc=0pt
]
\begin{lemma} [Risk bounds under $k$-unambiguity \citep{wang2023learning}] \label{lemma:ERM-learnability-M-unambiguity}
    If the space of learnable concept extractors $\calF$ is constructed like in \cref{eq:disentangled-f-definition} and $\beta^*$ is $k$-unambigous, 
    then we have:
    \[
    \calR^{01}_\vC(f)
        \le \calO(\calR^{01}_\vY (f;\beta^*)^{1/k})
        \quad \text{as} \;\;\; \calR^{01}_\vY (f;\beta^*) \to 0
    \]
    Moreover, if $\beta^*$ is $k$-ambiguous, then a concept extractor $f$ attaining label risk ${\calR^{01}_\vY(f;\beta^*)=0}$ can have a concept risk of ${\calR^{01}_\vC(f)=1}$. %
\end{lemma}
\end{tcolorbox}

\changed{\cref{lemma:ERM-learnability-M-unambiguity} shows that under $k$-unambiguity, minimizing label risk also drives the concept extractor toward correct concept predictions. Hence,
}
{
$k$-unambiguity is central to avoiding RSs for the specific concept extractors modeled in $\calF$. 
In fact, whenever $\beta^*$ is $k$-ambiguous, concept risk can be non-zero, resulting in RSs.
This happens naturally for the XOR of two bits, but not for the sum of the two, as detailed next.
}

\begin{tcolorbox}[
    colback=gray!2!white,
    colframe=gray,
    arc=0pt,
    outer arc=0pt
]
\begin{example}
    Consider a setting where the input $\vx =(\bar \vx_1, \bar \vx_2)$ comprises two bits, \eg $\vx=(\MZero, \MOne)$. Consider the knowledge for the \texttt{XOR} between the two, \ie $\BK = (C_1 \oplus C_2 \Leftrightarrow Y)$. 
    The function $\beta^*$ constructed from $\BK$ is $k$-ambiguous ($k=2$), as there are only two vectors with the same concept value that return the label. 
    That is, $\beta^*((0,0)^\top)=\beta^*((1,1)^\top)$, since $0 \oplus 0  = 1 \oplus 1=0$.
    We can see that, in this setting, there is a deterministic RS that confuses the zero for one and vice versa, that is, $\MZero\mapsto 1$ and $\MOne \mapsto 0$. 
    Consider now, instead, the sum of the two digits, with the knowledge given by $\BK =(Y = C_1 + C_2)$. 
    Here, the inference layer $\beta^*$ is $k$-unambiguous ($k=2$), and there are no concept remapping distributions other than the identity that give the correct label prediction. $k$-unambiguity is also satisfied for the complete \MNISTAdd, which is a NeSy task that does not have any RS for the disentangled concept extractors in $\calF$.
\end{example}
\end{tcolorbox}

Based on this result, \citet{wang2023learning} show that it is possible to  guarantee PAC learnability of the NeSy task. The result incorporates the Natarajan dimension of the space of classifiers $[\bar f]$, denoted as $d_{[\bar \calF]}$. 
We refer readers to \citep{understanding-ml} for the explicit definition of the Natarajan dimension.

\begin{tcolorbox}[
    colback=teal!2!white,
    colframe=teal,
    arc=0pt,
    outer arc=0pt
]
\begin{theorem}[ERM learnability under $k$-unambiguity \citep{wang2023learning}]
    \label{thm:ERM-learnability-v2} 
    Let $\calF$ be the space of learnable concept extractors $\calF$ constructed like in \cref{eq:disentangled-f-definition}. Suppose that $\mathcal{F}$ is realizable under $\ell^{01}_\vY$ and $[\bar \calF]$ has a finite Natarajan dimension $d_{[\bar \calF]}$.
    Then, for any $\epsilon, \delta \in (0,1)$, there exists a universal constant $C_0>0$, such that with probability at least $1-\delta$, the empirical label risk minimizer $f$ on the dataset $\calT_{p^*}$, such that 
    $\hat{\calR}^{01}_\vY(f;\beta^*;\calT_{p^*})=0$, has a concept risk $\calR^{01}_\vC(f) < \epsilon$, if 
    \begin{equation}
    \begin{aligned}
        m_{p^*}
        \ge 
        C_0\frac{\eta_1}{\epsilon^k} 
        \left( 
            \eta_2 \log \frac{1}{\epsilon^{k} }
            +
            \log \frac{1}{\delta} + \eta_3
        \right),
    \end{aligned}
    \label{eq:bound-pac-disent}
    \end{equation}
    where $\eta_1, \eta_2, \eta_3 \in \bbR^+$ are constants that depend on $k$, the %
    $d_{[\bar \calF]}$, and the number of labels $|\calY|$.
\end{theorem}
\end{tcolorbox}

\changed{ 
This shows that if the inference layer is $k$-unambiguous and the concept extractor class has finite complexity (as determined by the Natarajan dimension), then minimizing empirical label risk is sufficient to reduce the concept risk with high probability.
}
We refer the reader to
\citep{wang2023learning} for the explicit values of the constants. Furthermore, the work presents other results regarding the PAC learnability of the NeSy task under unknown $\beta^*$'s and error bounds when considering the semantic loss \citep{xu2018semantic}. 
\changed{As a final note, we remark that the bound in \cref{eq:bound-pac-disent} holds only in the setting of factorized concept extractor and generalizing it to wider hypothesis spaces $\calF$ is open.}

\subsection{Relationship between Theories}
\label{sec:slr-vs-identifiability} 

A common trait between the two perspectives can be found in \cref{cor:non-idf-nesy-rss} and \cref{thm:non-idf-risk-rss}: both results indicate that, even in the limit of infinite data, when conditions on the knowledge are not met (either because of non-injectivity or low knowledge complexity), RSs can arise during training.
Both results highlight a general problem in NeSy predictors: when we have a large hypothesis space of learnable concept extractors $\calF$ and
if the knowledge admits multiple solutions for label predictions, \textit{there is no unique and correct grounding of the concepts}. 
Conversely, if a one-to-one mapping between concepts and labels is implemented by the knowledge (either because the ground-truth inference layer $\beta^*$ is injective in \cref{cor:non-idf-nesy-rss} and $\mathrm{KC}(\BK, p^*) = |\calC| -1$ in \cref{thm:non-idf-risk-rss}), achieving optimal likelihood on data implies learning well-grounded concepts.
Achieving this condition, however, may be difficult in practice whenever the number of possible concept vectors exceeds that of labels, see e.g. \BOIA in \cref{ex:rss-bdd-oia-extended}.  Constrained hypothesis spaces for the concept extractor are more likely to meet the conditions for avoiding RSs with a less restricted $\BK$, as shown in \cref{lemma:ERM-learnability-M-unambiguity}, and lead to statistical learning bounds for recovering ground-truth concepts with finite-sized datasets (\cref{thm:ERM-learnability-v2}).

\changed{
We notice that the two perspectives differ in their main object of study. While the identifiability picture treats concept remapping distributions and studies identification of $f^*$ (\eg \cref{thm:identifiability-rss}), the statistical learning picture focuses on shortcut NeSy predictors and on bounding the concept risk (\eg \cref{thm:ERM-learnability-v2}). 
Because the statistical learning theory perspective ensures the avoidance of shortcut NeSy predictors, avoiding RSs is guaranteed only when $\calC = \mathrm{supp}(\vG)$, \ie when RSs in the problem are caused by shortcut NeSy predictors. 
In other cases where models are tested on unobserved concept combinations as in combinatorial generalization \citep{duan2025state}, \ie when training and test distributions differ, ensuring concept risk minimization might not ensure avoiding all RSs. 
}

\changed{Moreover, the} two perspectives differ in how they specialize in treating mitigations. 
The identifiability perspective \citep{marconato2023not,marconato2024bears, bortolotti2025shortcuts, van2025independenceRS} has so far investigated the impact of strategies on count reduction for deterministic $\alpha$'s, studying how different mitigations change the quantities in the count (\cref{eq:count-rss}). 
This also includes evaluating how the count reduces when different strategies are used together, but it is limited to studying optima of the learning problem. 
The statistical learning perspective \citep{wang2023learning, yang2024analysis, he2025learnability} has mainly explored how different mitigations that constrain the space of learnable concept extractors $\calF$ can be used to bound the empirical RSs risk. 
These results do not explicitly require studying the optima of the learning problem, in the sense that, constraints on $\calF$ can be directly translated to bounds on the RS risk (\cref{eq:rss-risk-expected}). In fact, strategies like concept smoothing change the optima of the learning problem, and may not allow reaching maximum likelihood.
Despite the different treatments, we note that the effect of some mitigation strategies can be described from both perspectives: a mitigation strategy can both reduce the count in \cref{eq:count-rss} and reduce the risk in \cref{eq:rss-risk-expected}. 
In this sense, we view both perspectives as complementary: through identifiability theory, one can analyze how mitigations act to reduce deterministic RSs at optimal likelihood, whereas through statistical learning theory, bounds on the empirical concept risk can be found even when likelihood is not optimal.
Formulating a theory that takes both perspectives into account remains open.

\newpage %
\section{Handling Reasoning Shortcuts}
\label{sec:handling-rss}

We proceed by detailing the root causes of RSs (in \cref{sec:root-causes}), and outlining existing diagnostic tools (\cref{sec:diagnosis}), mitigation strategies (\cref{sec:mitigations}), and awareness strategies (\cref{sec:awareness}).

\subsection{Root Causes}
\label{sec:root-causes}

As anticipated in \cref{sec:causes-frequency-impact}, RSs arise whenever the NeSy learning process does not penalize models that learn incorrect concepts, \ie concepts that do not match the ground-truth ones.
Moreover, the theory in \cref{sec:rss-theory} provides additional details.
Specifically, \cref{thm:non-idf-risk-rss} suggests that RSs can occur when the knowledge complexity $\mathrm{KC}$ (\cref{def:knowledge-complexity}) is less than $|\calC|-1$ and the space of learnable concept extractors $\calF$ is broad enough. \cref{cor:non-idf-nesy-rss} supports the same conclusion.
Following \citep{marconato2023not}, additional insights can be gleaned by studying the structure of \cref{eq:count-rss}, reported below for reference, which \textit{counts} the number of deterministic RSs (\cref{def:deterministc-rss}) affecting a NeSy task.  Since this is useful for understanding the mitigation strategies in \cref{sec:mitigations}, we analyze it here:
\vspace{1em}%
\[
    \eqnmark[WildStrawberry]{EMBVertA}{
        \color{black} \sum_{
            \alpha \in \textcolor{WildStrawberry}{\Vset{\calA_\calF}}
        }
    }
    \colorboxed{orange}{
    \mathbbm{1}\!\big\{
        \eqnmark[orange]{EMBObjective}{
        \color{black}
            \eqnmark[violet]{EMBSuppG}{\color{black}\bigwedge_{
                \vg \in \textcolor{violet}{\mathrm{supp}(\vG)} 
            }}
                (\eqnmark[teal]{EMBBeta1}{\beta^*} \circ \alpha)(\vg)
                =
                \eqnmark[teal]{EMBBeta2}{\beta^*}(\vg)
            }    
    \big\}
    } -1.
    \label{eq:causes}
\]
\annotate[yshift=0em]{below,left}{EMBVertA}{\color{WildStrawberry}learnable maps $\alpha$}
\annotate[yshift=-0.25em]{below,right}{EMBSuppG}{support}
\annotatetwo[yshift=1em]{above}{EMBBeta1}{EMBBeta2}{prior knowledge $\BK$}
\annotate[yshift=-0.75em]{left}{EMBObjective}{optimality condition $\calL$} %
\vspace{1em}%

\noindent The count is controlled by the four colored elements, which we discuss below.

\subsubsection{The knowledge}  \label{sec:cause-prior-knowledge}  The \textbf{\textit{\textcolor{teal}{prior knowledge $\BK$}}} determines the inference layer $\beta^*$. To understand its role,
\begin{wrapfigure}[10]{r}{0.3\textwidth}
    \centering
    \small
    \begin{tabular}{cc|c|c}
        \toprule
        $C_1$ & $C_2$ & $Y_1$ & $Y_2$ \\
        \midrule
        0 & 0 & 0 & 0 \\
        0 & 1 & 1 & 1 \\
        1 & 0 & 2 & 1 \\
        1 & 1 & 3 & 1 \\
        \bottomrule
    \end{tabular}
    \caption{$\BK_1$ is immune to RSs, $\BK_2$ is not.}
    \label{tab:k-increase-count}
\end{wrapfigure}
imagine two NeSy tasks with two binary concepts that are identical except for the prior knowledge $\BK$: one uses $\BK_1 = (Y_1 = 2C_1 + C_2)$ and the other $\BK_2 = (Y_2 = C_1 \lor C_2)$, see \cref{tab:k-increase-count}.
Assuming the training set covers all classes (see \cref{sec:cause-training-support}), the task using $\BK_1$ does not admit any RSs:  each label $y$ can only be inferred by a unique choice of $\vc$, so high label accuracy entails high concept accuracy.  The task using $\BK_2$, however, \textit{is} affected by RSs:  the positive label can be inferred from $\vc = (0, 1)$, $\vc = (1, 0)$ and $\vc = (1, 1)$, hence NeSy predictors can confuse them with no impact on accuracy.
From the perspective of \cref{eq:causes}, this happens because, all else being equal, swapping $\BK_1$ with $\BK_2$ softens the equality condition, increasing the count.

\subsubsection{The training distribution}  \label{sec:cause-training-support}  Another important factor is the set of configurations of ground-truth concepts $\vg$ that underlie the training data, that is, the \textit{\textbf{\textcolor{violet}{support}}} $\mathrm{supp}(\vG)$ of the training distribution.
To see this, consider the \MNISTAdd example in \cref{ex:mnist-rss-full}.  Suppose we add two new examples, $(\MFive\MFive, 10)$ and $(\MFive\MThree, 8)$, to the training set.  Intuitively, the concept extractor \textit{has} to map $\MFive$ to the digit $5$, otherwise it would mispredict the $(\MFive\MFive, 10)$ example; likewise, now that $\MFive$ has only one correct grounding, it has to map $\MFour$ to $4$ otherwise it would mispredict the $(\MFive\MFour, 9)$ example. The same reasoning applies for the remaining examples, namely ($\MFive\MThree$, 8) and ($\MThree\MTwo$, 5).  In summary, this small change completely removes all RSs, as the only way to match all training examples is to assign each MNIST digit to its correct numeric value.
Intuitively, the larger the support -- \ie the more combinations of ground-truth concepts $\vg$ the training set represents -- the more restrictive the conjunction in \cref{eq:causes}, lowering the count.

\subsubsection{The optimality condition}  \label{sec:cause-optimality-condition} The \textit{\textbf{\textcolor{orange}{optimality condition}}} $\calL$ is responsible for filtering out those maps $\alpha$ that do not fit the ground-truth labels.
Consider \cref{ex:mnist-rss-full} again.  One possible RS is $\{ \MTwo \mapsto 4, \MThree \mapsto 1, \MFour \mapsto 5, \MFive \mapsto 4\}$, which collapses \MFive and \MTwo into the same concept value, $4$.  Now, imagine augmenting the training objective, by default the cross-entropy, with a reconstruction loss.  This mapping no longer achieves low loss, since the value $4$ decodes to both \MTwo and \MFive, which clearly hinders reconstruction.
In terms of \cref{eq:causes}, modifying the training objective changes the optimality condition, potentially decreasing the number of RSs.

\subsubsection{The family of learnable maps}  \label{sec:cause-learnable-maps}  
Finally, the sum in \cref{eq:count-rss} runs over (the vertices of) \textit{\textbf{\textcolor{WildStrawberry}{the space of learnable concept remappings}}} $\alpha$, denoted by $\Vset{\calA_\calF}$.  
This, in turn, depends on the space of learnable concept extractors $\calF$ (\cref{thm:count-rss}).
By introducing architectural bias in the architecture of the concept extractor %
(\eg smoothing concept extractors by tuning a temperature parameter), we can %
constrain the space of learnable concept remappings $\alpha$'s and therefore decrease the number of learnable RSs $\Vset{\calA_\calF}$.  
We will see in \cref{sec:mitigations} how to effectively control this aspect of the learning problem.

\subsection{How to Diagnose Reasoning Shortcuts}
\label{sec:diagnosis}

As mentioned in \cref{sec:a-gentle-intro-to-rss}, in-distribution label accuracy is not affected by RSs and is therefore insufficient for diagnosing them. 
In this section, we briefly outline more effective diagnostic techniques.
\subsubsection{Task-level diagnosis}  \label{sec:rscount}  Even before we commit to an architecture and train the model, we can \textit{\textbf{quantify how many deterministic RSs can affect a given NeSy task}}.
We accomplish this by evaluating \cref{eq:count-rss}, which is defined on known properties, such as the number of concepts, their cardinality and the given annotations.
However, the count is very hard to compute manually even for relatively small problems, as the number of maps $\alpha$ increases doubly exponentially with the number of concepts $k$.

The \countrss tool~\citep{bortolotti2024benchmark} works around this by building on the following observation:  all deterministic maps $\alpha$ in $\Vset{\calA}$ can be viewed as binary matrices that indicate which predicted concepts $C_j$ correspond to which ground-truth concepts $G_i$. 
The assumptions that constrain the number of mappings can be encoded into a propositional logical formula, whose solutions (that is, admissible binary matrices) are in one-to-one correspondence to the deterministic RSs $\alpha$.
Then, the count in \cref{eq:count-rss} can be reduced to \emph{model counting} (\#SAT)~\citep{modelcounting}, \ie the task of counting the solutions of a logical formula.

\begin{figure}[!h]
\newcolumntype{x}{>{\centering\let\newline\\\arraybackslash\hspace{0pt}}p{0.6cm}}

\begin{minipage}{0.4\linewidth}
    \centering
    \begin{tabular}{r|x x x x|}
        \multicolumn{1}{r}{}
        & \multicolumn{1}{c}{$C_{\tt red}$}
        & \multicolumn{1}{c}{$\neg C_{\tt red}$}
        & \multicolumn{1}{c}{$C_{\tt ped}$}
        & \multicolumn{1}{c}{$\neg C_{\tt ped}$}
        \\ \cline{2-5}
        $G_{\tt red}$ & 0 & 0 & 1 & 0 \\ %
        $\neg G_{\tt red}$ & 0 & 0 & 0 & 1 \\ %
        $G_{\tt ped}$ & 1 & 0 & 0 & 0 \\ %
        $\neg G_{\tt ped}$ & 0 & 1 & 0 & 0 \\ \cline{2-5}
    \end{tabular}

    {\small (a)}
\end{minipage}
\begin{minipage}{0.4\linewidth}
    \centering
    \begin{tabular}{r|x x x x|}
        \multicolumn{1}{r}{}
        & \multicolumn{1}{c}{$C_{\tt red}$}
        & \multicolumn{1}{c}{$\neg C_{\tt red}$}
        & \multicolumn{1}{c}{$C_{\tt ped}$}
        & \multicolumn{1}{c}{$\neg C_{\tt ped}$}
        \\ \cline{2-5}
        $G_{\tt red}$ & 1 & 0 & 0 & 0 \\ %
        $\neg G_{\tt red}$ & 0 & 1 & 0 & 0 \\ %
        $G_{\tt ped}$ & 0 & 0 & 0 & \textcolor{WildStrawberry}{1} \\ %
        $\neg G_{\tt ped}$ & 0 & 0 & 0 & 1 \\ \cline{2-5}
    \end{tabular}

    {\small (b)}

\end{minipage}
    
    \caption{Binary matrices encoding two possible mappings $\alpha$ from \cref{ex:boia}. (a) is a solution, resulting in a mapping that mistakes red lights with pedestrians and vice versa but is otherwise consistent with $\BK$ for any combination of ground truth concepts. (b) is not a solution, since the resulting mapping makes the predictor unable to detect pedestrians and give correct predictions on some, but not all, combinations of ground truth concepts.}
    \label{fig:countrss}
\end{figure}

Although one could use any \#SAT procedure, the complexity of the encoding for any task of practical interest warrants the use of approximate \#SAT solvers~\citep{approxmc}. Under the hood, \texttt{countrss} employs the state-of-the-art hashing-based counter \texttt{ApproxMC}\footnote{\url{https://github.com/meelgroup/approxmc}}, which outputs an $\epsilon$ approximation of the exact count with probability $\delta$. \texttt{countrss} inherits these statistical (PAC-style) guarantees, providing close approximations to \cref{eq:count-rss} with high probability~\footnote{Regardless of the solver used, the reduction assumes disentanglement. Otherwise, the resulting number of logical variables in the encoding would be orders of magnitude above the capabilities of current solvers.}.
The estimates returned by \texttt{countrss} can provide an initial indication of the hardness of the learning problem and help make informed architectural choices according to the task at hand. \texttt{countrss} is included in the \texttt{rsbench} library~\citep{bortolotti2024benchmark}.

\rechanged{An alternative to count-based diagnosis is to certify shortcut-freeness directly. \citet{takemura2026constraint} cast the problem as a constraint satisfaction problem and propose a sound Answer Set Programming~\citep{lifschitz2019answer} procedure that checks whether a given constraint set uniquely determines the intended concept mapping. Whereas \texttt{countrss} estimates \emph{how many} shortcuts a task admits, this procedure answers \emph{whether} any exist, making the two complementary.}

\subsubsection{Model metrics}  \label{sec:metrics} Whenever concept annotations are available, the most effective way to evaluate the quality of the learned concepts is to use standard metrics such as accuracy, $F_1$ score, and concept-level confusion matrices.
Another useful metric is \emph{concept collapse} \citep{bortolotti2024benchmark}.  Roughly speaking, it quantifies to what extent the learned concept extractor compresses distinct ground-truth concepts into the same learned concept.  More formally, let $C \in [0,1]^{m \times m}$ be a concept confusion matrix, where $m$ denotes its size (\eg, $m = 2^k$ when all ground-truth concepts are observed). We define $\Collapse = 1 - \tfrac{p}{m}$, where $p = \sum_{j=1}^m \Ind{\exists i \,:\, C_{ij} > 0}$. %
We remark that, however, collapse cannot detect all RSs.  For instance, imagine a NeSy predictor trained on some variant of \MNISTAdd affected by an RS mapping \MZero to $1$ and \MOne to $0$.  This RS is completely invisible to the collapse metric, as the model does not collapse ground-truth concepts together, cf. \cref{sec:mitigation-reconstruction}.  Still, it can be useful to gauge the presence of RSs that do.
Finally, a major downside of these metrics is that they all require concept-level annotations, which are not always available.

{With insufficient or missing concept annotations, another approach is to train multiple NeSy predictors to high label accuracy, and check their concept extractors align on concept predictions. 
When (deterministic) RSs are many in a NeSy task, it is likely that each predictor finds a different RS. 
Furthermore, if there are no RSs, concept extractors giving optimal likelihood must predict the same concepts, although the reverse is not necessarily true. 
The alignment between different concept extractors of the ensemble can be indirectly measured via the overall conditional entropy of the averaged concept predictions of the ensemble. Precisely, for $r \in \bbN$ members of the ensemble, we can account the mean conditional Shannon entropy $\mathsf{H}(\frac{1}{r}\sum_j p_j(\vC \mid \vx))$.   
This quantity is central for unsupervised RS-aware methods such as \bears~\citep{marconato2024bears} and \nesydm \citep{van2025neurosymbolic}, see \cref{sec:awareness}.  
}

\subsection{A Tour of Mitigation Strategies}
\label{sec:mitigations}

Several strategies \changed{relevant to} mitigating RSs have been proposed.
We list them in \cref{tab:mitigations} and discuss them next.

\subsubsection{Concept supervision}  \label{sec:mitigation-supervision}  The most direct approach for encouraging correct grounding is to exploit \textit{\textbf{concept annotations}} during training, thus altering the \textit{\textbf{\textcolor{orange}{optimality condition}}}.
This requires introducing an additional loss term penalizing mispredicted concepts.  Doing so effectively constrains the concept extractor $f$ to recover the annotations, dramatically reducing the number of learnable RSs, and potentially avoiding them altogether.

This solution is, however, neither cheap nor perfect.
It can be costly because concept-level annotations are frequently unavailable in the wild, and collecting them involves running potentially expensive annotation campaigns.
Following recent trends \citep{oikarinen2022label, yang2023language, rao2024discover, srivastava2024vlgcbm, yuksekgonul2023post}, one could obtain these annotations with foundation models.  This solution, however, risks injecting substantial amounts of noise into the learning loop~\citep{debole2025concept}: despite being trained on vast amounts of data, foundation models can mispredict even basic concepts~\citep{wust2025bongard}.

It is also imperfect because, in the worst case, ruling out all RSs requires exhaustively annotating \textit{all possible combinations of concepts}.  Unless this is the case, a sufficiently expressive concept extractor might fit all annotations on the supervised combinations and fluke the remaining ones.  Naturally, annotating an exponential (in the number of concepts) number of combinations can be infeasible.
A possible workaround is to restrict the expressivity of the concept extractor---\eg by processing inputs separately, as typically done in \MNISTAdd---so as to effectively transfer concept annotations across concept combinations.
Another option is to intelligently select instances or concepts to be annotated; see \cref{sec:mitigation-interaction}.

\begin{table}[!h]
    \caption{\textbf{Overview of mitigation strategies}.  The columns indicate, for each strategy,
    what component(s) of~\cref{eq:causes} it targets (\textbf{Target}),
    whether it requires extra annotations (\textbf{Annot}) and guarantees reducing the count in \cref{eq:count-rss} (\textbf{Count}),
    and how well it worked on different NeSy Tasks.  Specifically, ``\CMARK'' means that it prevents all RSs, ``\XMARK'' that it fails to do so, and ``\QMARK'' that it has not been evaluated. 
    \rechanged{Each \textbf{Target} value corresponds to a component of~\cref{eq:causes}: $\textcolor{teal}{\BK}$ reshapes the knowledge constraint $({\beta^*} \circ \alpha)(\vg) = \beta^*(\vg)$, while $\textcolor{WildStrawberry}{\calF}$ and $\textcolor{orange}{\calL}$ both act through $\textcolor{WildStrawberry}{\Vset{\calA_\calF}}$: $\textcolor{WildStrawberry}{\calF}$ as the family of representable maps $\alpha$, and $\textcolor{orange}{\calL}$ as the optimality condition filtering these to the optimal ones.}
    \textbf{Note:} With \MNISTAdd$^*$ we refer to the family of datasets based on the \MNISTAdd benchmark~\citep{manhaeve2018deepproblog}. %
    }
    \label{tab:mitigations}
    \centering
    \small
    \scalebox{0.85}{
        \begin{tabular}{clccccccc}
        \toprule
            & {\bf Strategy}
            & \multicolumn{3}{c}{\textbf{Properties}}
            & \multicolumn{1}{c}{\textbf{Arithmetic}} 
            & \multicolumn{2}{c}{\textbf{Logic}} 
            & \multicolumn{1}{c}{\textbf{High-Stakes}}
        \\
        \cmidrule(lr){3-5} \cmidrule(lr){6-6} \cmidrule(lr){7-8} \cmidrule(lr){9-9}
            &
            & {\bf Target}
            & {\bf Annot}
            & {\bf Count}
            & \MNISTAdd$^*$
            & \Kandinsky
            & \Clevr
            & \BOIA
        \\
        \midrule
            & Concept Supervision~\citep{koh2020concept}
            & $\textcolor{orange}{\calL}$ %
            & \CMARK
            & \CMARK
            & \CMARK & \CMARK & \CMARK & \XMARK
        \\
            & \changed{Knowledge editing}
            & $\textcolor{teal}{\BK}$
            & \XMARK
            & \XMARK
            & \QMARK & \QMARK & \QMARK & \QMARK
        \\
            & Multi-task Learning~\citep{caruana1997multitask}
            & $\textcolor{teal}{\BK}$
            & \CMARK
            & \CMARK
            & \CMARK & \QMARK & \QMARK & \QMARK
        \\
            & \changed{Prototype-based grounding}~\citep{snell2017prototypical}
                & $\textcolor{orange}{\calL}$
                & \CMARK
                & \CMARK
                & \CMARK & \CMARK & \CMARK & \XMARK
        \\
        \midrule
            & Weak Supervision \citep{yang2024analysis}
            & $\textcolor{orange}{\calL}$
            & \XMARK
            & \XMARK
            & \CMARK & \QMARK & \QMARK & \XMARK
        \\
            & Entropy Maximization~\citep{manhaeve2021neural}
            & $\textcolor{orange}{\calL}$
            & \XMARK
            & \XMARK
            & \QMARK & \QMARK & \QMARK & \XMARK
        \\
            & Smoothing \citep{szegedy2016rethinking}
            & $\textcolor{orange}{\calL}$
            & \XMARK
            & \XMARK
            & \QMARK & \QMARK & \QMARK & \QMARK
        \\
            & Reconstruction~\citep{kingma2013auto}
            & $\textcolor{orange}{\calL}$
            & \XMARK
            & \CMARK
            & \CMARK & \QMARK & \QMARK & \QMARK
        \\
            & Contrastive Learning~\citep{chen2020simple}
            & $\textcolor{orange}{\calL}$
            & \XMARK
            & \XMARK
            & \QMARK & \QMARK & \QMARK & \QMARK
        \\
            & Disentanglement~\citep{suter2019robustly}
            & $\textcolor{WildStrawberry}{\calF}$
            & \XMARK
            & \CMARK
            & \QMARK & \XMARK & \QMARK & \QMARK
        \\
        \bottomrule
    \end{tabular}

    }
\end{table}

\subsubsection{Knowledge editing}  \changed{\label{sec:knowledge-editing}  An alternative is to edit the \textbf{\textit{\textcolor{teal}{prior knowledge $\BK$}}} by introducing additional constraints, so as to remove ambiguities between concepts.
For instance, in \cref{ex:boia}, one can break the symmetry between {\tt pedestrian} and {\tt red\_light} by introducing a mutual exclusivity constraint between {\tt red\_light} and {\tt green\_light}.  This way, the model can no longer freely confuse them, provided they appear together in the data, as in \cref{fig:boia-rs} (Right).
While this strategy can be very effective, it requires figuring out what to change in the knowledge and sufficient expertise to implement the changes appropriately.
Moreover, it is not always applicable.  To see this, consider \MNISTAdd.  Here, the knowledge specifies the rule of addition:  there is no way to meaningfully edit it without altering the semantics of this operation.  This means that knowledge editing cannot be used to avoid RSs in any variants of this task.}

\subsubsection{Multi-task learning}  \label{sec:mitigation-multitask}  Another way to improve the \textbf{\textit{\textcolor{teal}{prior knowledge $\BK$}}} is to train a NeSy predictor to solve \textit{\textbf{multiple NeSy tasks}} (sharing some of the same concepts) jointly~\citep{caruana1997multitask}.
Doing so equates to training on a single NeSy task whose prior knowledge is the \textit{conjunction} of the prior knowledge of the different tasks \citep{marconato2023not}.  As per \cref{eq:causes}, doing so reduces the number of potential RSs. \changed{Intuitively, this constraints the map $\beta^*$ to distinguish among more different concept vectors than with a single task and excludes concept remapping distributions that are valid for tasks in isolation.  
} %

However, not all combinations of NeSy tasks are equally effective.
To see this, imagine training a NeSy predictor to output both the sum of two \MNIST digits and whether the first one is greater than the second one.  Observations take the form $((\MFour\MFive), (9, 0))$ and $((\MThree\MTwo), (5, 1))$, where the second label indicates whether the inequality holds. The RS in \cref{ex:mnist-rss-full}, \ie $\{ \MTwo \mapsto 4,\ \MThree \mapsto 1,\ \MFour \mapsto 3,\ \MFive \mapsto 6 \}$, is no longer valid, as it incorrectly predicts $(5, 0)$ for the second example.  However, an alternative shortcut exists, namely $\{ \MTwo \mapsto 1,\ \MThree \mapsto 4,\ \MFour \mapsto 3,\ \MFive \mapsto 6 \}$.
A better alternative is to pair \MNIST addition and multiplication. This is identical to \cref{ex:mnist-rss-full}, except now there are two labels, \eg $((\MFour\MFive), (9, 20))$ and $((\MThree\MTwo), (5, 6))$.  It can be verified that the only concept mapping that correctly predicts both the sum and the product is the intended one, \ie the identity, ruling out all RSs.

Naturally, multi-task learning requires gathering label annotations for all involved tasks, which can be non-trivial.  Another downside is that it may not be obvious how to construct a ``natural'' NeSy task that guarantees the removal of all RSs. 
We will discuss possible solutions in \changed{\cref{sec:open-problems}}. %

\subsubsection{Abductive weak supervision}  \label{sec:mitigation-abduction} 
This strategy requires training the concept extractor using \textit{\textbf{procedurally generated}} pseudo-labels obtained with logical abduction~\citep{yang2024analysis}. %
\ABL embodies this approach: in \ABL, the model uses logical abduction to infer the most plausible concept vector $\vc$ that logically explains the ground-truth label $\vy$ according to the prior knowledge. 
In practice, \ABL uses a pre-defined distance function $d(\vc, \vc')$ to select this concept vector from alternatives. 
Then, \ABL uses this new concept vector as the learning target for the loss function.
Choosing an appropriate distance function can provide \ABL with concept risk reduction; specifically, when the distance metric is always zero,
\citet{yang2024analysis} showed that the risk of learning a shortcut solution reduces compared to other approaches.  %
The downside is that there is no guarantee that the abduction process will recover the ground-truth concepts.

\subsubsection{\changed{Prototype-based grounding}} \label{sec:mitigation-prototype}
\changed{
Another way to mitigate RSs is to rely on \textbf{prototypes}, \ie anchoring each concept to a small \textit{support set} of representative examples~\citep{snell2017prototypical, bontempelli2023concept, andolfi2025right}. Following~\citet{andolfi2025right}, prototype-based grounding addresses RSs by constraining the concept extractor to assign concepts $\vc$ to an input $\vx$ based on their \textit{similarity} to the prototypes in the embedding space.
For each concept $\vc$, a prototype is computed as the centroid of the embeddings of a (possibly very small) set of labeled examples. Compared to a standard NeSy predictor, during training~\citet{andolfi2025right} include an extra objective in which the model has to correctly classify the elements in the \textit{support set} with respect to their corresponding prototypes based on distances (\eg L2 distance). At inference time, instead, the probability of a predicted concept depends on the distance between the input representation and the corresponding prototype, typically via a softmax over distances.
From the perspective of \cref{eq:causes}, prototype-based grounding targets the \textit{\textbf{\textcolor{orange}{optimality condition}}}: by enforcing concept assignments to respect distances to their prototypes, it imposes a specific geometry on the embedding space, thereby reducing the number of possible solutions the model can learn.
The main advantage of this approach is that it requires relatively few annotations, in contrast to dense concept supervision~\citep{marconato2023neuro}, as only a small set of samples per concept is needed to compute the prototypes. A limitation, however, is that this approach assumes concepts are reasonably clusterable in the learned embedding space and that representative prototypes can be found. In domains where concepts are highly entangled or lack clear visual or semantic representations, defining meaningful prototypes can be challenging.
\citet{andolfi2025right} provide guarantees on the number of RSs when prototype-based grounding is employed. The main intuition is that, under the assumption that concepts form separable clusters in the embedding space, anchoring each concept with at least one labeled prototype reduces the number of deterministic RSs to the same level as full concept supervision~\citep{marconato2023not}.
}

\subsubsection{Entropy maximization} \label{sec:mitigation-entropy} Perhaps the simplest \textit{unsupervised} mitigation strategy is \textit{\textbf{entropy maximization}}~\citep{manhaeve2021neural}.
{A simple idea here is to introduce an entropy-based loss term encouraging the model to evenly distribute probability mass across \textit{all} concept combinations. 
This can be made practical by averaging over all predictions in a batch, and maximizing the Shannon entropy $\mathsf{H}(p(\vC))$ ~\citep{manhaeve2021neural}.} 

Entropy maximization is easy to implement and computationally lightweight.
However, it only encourages the predictor to spread concept predictions, but does not guarantee that the learned concepts will match the ground-truth ones. %
Furthermore, there are no formal results that show this method reduces the number of RSs.%

\subsubsection{Smoothing}  \label{sec:mitigation-smoothing}  The idea behind \textit{\textbf{concept smoothing}} is to encourage the probabilities of the most and least probable concept vectors $\vc$ to be ``close''.
The main idea is to never allow NeSy predictors to learn "peaked" concept distributions, \ie $p(\vC \mid \vX)$ of the model is never one-hot.
Smoothing can be enforced either architecturally, \eg by introducing a temperature parameter that reduces the magnitude of the concept activations, or during training, by penalizing high activations on input samples.
This strategy permits reducing the RSs risk (\cref{eq:rss-risk-expected}), the full derivation can be found in \citep{yang2024analysis}.
This strategy, however, only circumvents \textit{some} RSs:  while smoothing prevents deterministic RSs, it does not affect non-deterministic ones.
\Eg imagine that $\BK$ allows inferring the same label vector $\vy$ from both $\vc'$ and $\vc''$ (that is, $\beta^*(\vc') = \beta^*(\vc'') = \vy$, and that a NeSy predictor predicts a \textit{mixture} of the two, \eg because it has learned the map $\alpha(\vg) = 0.5 \Ind{\vC=\vc'} + 0.5 \Ind{\vC=\vc''}$, where $\vg \in \{ \vc', \vc'' \}$.  The mapping $\alpha$ is smooth -- the highest probability it can output is $0.5$ -- but it is also a (non-deterministic) reasoning shortcut for \PNSP and \SL, as it produces a correct prediction: $0.5 \beta^*(\vc') + 0.5 \beta^*(\vc'') = \Ind{\vY=\vy}$.

\subsubsection{Reconstruction}  \label{sec:mitigation-reconstruction} Another option is to introduce a \textit{\textbf{reconstruction penalty}} into the learning process ~\citep{cottrell1987learning}.  Doing so encourages learning concepts that allow reconstructing the input $\vx$, or part of it.  This rules out RSs that conflate unrelated concepts together and thus prevent reconstruction.
To see how it works, consider \MNISTAdd (\cref{ex:mnist-rss-full}) again:  NeSy predictors achieving low reconstruction loss cannot map different \MNIST digits to the same concept, as doing so prevents them from faithfully reconstructing the original inputs (see \cref{sec:cause-optimality-condition}). For instance, the mapping ${ \MTwo \mapsto 4, \MThree \mapsto 1, \MFour \mapsto 5, \MFive \mapsto 4}$ implies that a single concept, $2$, has to decode into two different digits, \MTwo and \MFive, leading to subpar reconstruction loss.

While reconstruction can effectively rule out such RSs, it cannot prevent all of them.  In fact, a mapping like ${\MTwo \mapsto 4, \MThree \mapsto 1, \MFour \mapsto 3, \MFive \mapsto 6}$ satisfies the knowledge \textit{and} does not hinder reconstruction.
It is easy to see that this mitigation targets the \textit{\textbf{\textcolor{orange}{optimality condition}}} in \cref{eq:causes}, and that as such it can decrease the number of RSs.
Moreover, it comes at a cost: reconstruction requires adding a decoder module, increasing model size, computational overhead, and training complexity in practice.
Crucially, for reconstruction to be effective, the learned concepts must correspond to distinct and separable visual features, as in \MNISTAdd.  
This is not always the case. 
In real-world settings, \eg\ \BOIA~\citep{xu2020boia}, reconstructing the input image can be very difficult, requiring additional information such as object bounding boxes and segmentation masks. 
\Eg reconstructing a \texttt{pedestrian} requires isolating it from the background, which is not possible from raw pixel input alone. 
This makes reconstruction impractical in NeSy tasks involving complex, real-world inputs.

\subsubsection{Contrastive learning}  \label{sec:mitigation-contrastive} Another effective unsupervised strategy targeting the \textit{\textbf{\textcolor{orange}{optimality condition}}} is \textit{\textbf{contrastive learning}}~\citep{chen2020simple}.
The core idea is to encourage similar inputs to produce the same \textit{concepts}. 
This is typically achieved by forcing augmented views of the same input (positive pairs) to predict the same concepts, while ensuring that different inputs (negative pairs) yield distinct ones. For instance, given a \MNIST digit, slight rotations or shifts can be applied to create positive pairs, while the remaining images in the batch serve as negatives. This encourages the model to map together different visual appearances of the same concept (\eg \MThree) while keeping the others separate (\eg \MFive).

Although contrastive learning provides no formal guarantees for reducing RSs, in practice it has an effect similar to reconstruction, as it addresses RSs that collapse semantically distinct inputs into a single concept.
For instance, in~\cref{ex:mnist-rss-full} both strategies can prevent the model from mapping visually distinct digits such as \MThree and \MFive to the same concept, \eg $4$.
Contrastive learning is, however, easier to implement (\eg using stock implementations of the \texttt{infoNCE} loss~\citep{oord2018representation}) and significantly easier to optimize than reconstruction-based approaches.
Nevertheless, care must be taken when designing augmentations to ensure they preserve the semantic meaning of the underlying concepts. For example, rotating a digit such as \MSix in \MNIST can transform it into a \MNine, which completely changes the semantics of the image, introducing noise \citep{chang2020assessing}.

\subsubsection{Architectural disentanglement}  \label{sec:mitigation-disentanglement}  The last unsupervised strategy we consider is \textit{\textbf{architectural disentanglement}}.  While the notion of disentanglement has different meanings in the literature \citep{higgins2018towards, suter2019robustly}, here it simply means that the concept extractor processes independent objects in the input separately.

For instance, in visual recognition tasks in which multiple objects appear in the scene and each object is associated with a separate set of concepts (\eg shape, size and color), architectural disentanglement amounts to processing each object separately.
To see why this makes sense, consider \MNISTAdd again.  Here, the input comprises two digits.  When using
a single concept extractor that processes \textit{both} digits at once, there is a chance that it will swap the order of the two digits, as doing so leaves their sum unchanged.  Separate processing resolves this ambiguity.
More generally, this strategy prevents the model from confusing the concepts of one object with those of a different object, thus reducing the number of RSs.

A nice benefit of this architectural bias is that it dramatically reduces the space of learnable concept extractors $\calF$, which shrinks the space of learnable models (\ie $\Vset{\calA_\calF}$ in \cref{eq:causes}) and therefore the number of learnable RSs.  Moreover, it enables parameter sharing across objects, reducing sample complexity.
It is, however, not always applicable or easy to implement.  On the one hand, it only makes sense for NeSy tasks in which the inputs contain multiple independent objects, such as \MNISTAdd.  On the other hand, it requires being able to isolate what part of the input encodes what object, which can be non-trivial depending on the complexity of the input.  For instance, applying disentanglement to tasks like \BOIA requires using either ground-truth bounding boxes, which are seldom available, or predicted segmentations from tools such as Yolo~\citep{redmon2016you, shindo2023thinking} and FastRCNN~\citep{girshick2015fast, marconato2023not}.  These, however, can introduce noise in the learning process.

\subsubsection{Which mitigation strategy is best?}
\label{sec:mitigations-discussion} 

As shown in \cref{tab:mitigations}, different strategies offer different efficacy-cost trade-offs.
The main takeaway is twofold: on the one hand, \textit{\textbf{supervised strategies}} like concept supervision \cref{sec:mitigation-supervision} and multi-task learning \cref{sec:mitigation-multitask} can be very effective but also expensive; on the other, \textit{\textbf{unsupervised strategies}} are cheaper to implement but also less reliable.
Specifically, while some of them (like reconstruction \cref{sec:mitigation-reconstruction}, abduction-based weak supervision \cref{sec:mitigation-abduction} and smoothing \cref{sec:mitigation-smoothing}) provably avoid \textit{certain} kinds of RSs, they are ineffective for others.

There is no one-size-fits-all strategy and the best option is application-specific, in that it depends on factors like the cost of obtaining annotations and the kind of RSs one wishes to eliminate.
One option is to \textit{\textbf{mix multiple strategies}}, yet not all combinations are useful.
For instance, %
high smoothness
is at odds with both concept supervision and reconstruction: the former aims to spread probability mass across concepts, increasing their uncertainty and uninformativeness, whereas the latter attempt to fit the annotations with high certainty and to inject as much information as possible into the concepts, respectively.  No model can fully satisfy both objectives.
Finally, combining multiple strategies can yield overly complex loss functions that are challenging to optimize in practice, \eg because they require careful hyperparameter tuning.

\subsubsection{Can one simply recover learned concepts?}  Concepts learned without supervision risk encoding incorrect, misaligned semantics.  A natural workaround, that has been sometimes used in the literature \citep{daniele2023deep, tang2023perception}, is to figure out what concepts the model has learned in a \textit{post-hoc} fashion, by matching concept predictions with ground-truth concept annotations.

However, this strategy may not work: it only works insofar as the concept extractor \textit{identifies} (in a technical sense, see \citep{bortolotti2025shortcuts}) the ground-truth concepts modulo a permutation of their indices and element-wise invertible transformations \citep{bortolotti2025shortcuts}.  Given how difficult it is to ensure latent variables can be identified \citep{hyvarinen2024identifiability}, this is unlikely to happen by chance.  For instance, in \MNISTAdd, this strategy only works if the concept extractor has learned to predict the ground-truth digits, except shuffled---\eg it predicts $4$ with $8$, but predicts them perfectly when reordered.  On the flip side, this strategy cannot work whenever the concept extractor collapses together multiple ground-truth concepts, \eg it predicts $6$ whenever it sees a \MFive and an \MSix.

\subsection{Awareness Strategies}
\label{sec:awareness}

\begin{figure}[!t]
    \centering
    \includegraphics[width=\linewidth]{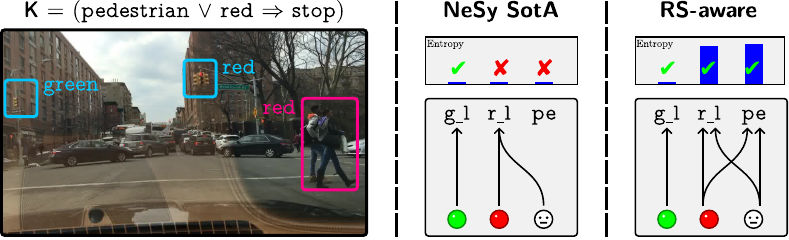}
    \caption{\textbf{NeSy predictors are not RS-aware} \citep{marconato2024bears}.
    Left: in the \BOIA autonomous driving task~\citep{xu2020boia}, NeSy predictors can achieve high accuracy and satisfy the prior knowledge even when they confuse pedestrians (\texttt{ped}) and red lights (\texttt{red})~\citep{marconato2023not}.
    Right: An RS-aware predictor assigns high confidence to shortcut concepts, while giving low confidence to concepts not affected by RS. %
    }
    \label{fig:awareness-intuition}
\end{figure}

Another strategy for dealing with RSs is to \textit{\textbf{make NeSy predictors aware of the RSs that affect them}}.
We say that a NeSy predictor is \textit{\textbf{RS-aware}} if it picks a non-deterministic RS that exhibits high confidence for those predicted concepts that are \textit{not} affected by RSs and low confidence for the others \citep{marconato2024bears}.
As customary, confidence can be assessed using the \textit{entropy} of the predictive concept distribution $p(\vC \mid \vx)$. %

While awareness does \textit{not} reduce RSs, RS-aware models supply stakeholders with valuable insights into the quality of the learned concepts, helping them to identify and distrust predictions obtained with poorly grounded concepts. 
To see this, consider the following example adapted from  \citep{marconato2024bears}:

\begin{tcolorbox}[
    colback=gray!2!white,
    colframe=gray,
    arc=0pt,
    outer arc=0pt
]
\begin{example}
    The left NeSy predictor in \cref{fig:awareness-intuition} is not RS-aware: despite confusing pedestrians with red lights, it is highly certain about the concepts it predicts.
    The one on the right is equally confused, but it is also RS-aware, in that it is more uncertain about low-quality concepts (pedestrian and red lights) than about high-quality ones (green lights). %
    This enables users to distinguish between these cases.
\end{example}
\end{tcolorbox}

Moreover, meaningful uncertainty estimates are also a powerful tool for steering interactive acquisition of annotations (\cref{sec:mitigation-interaction}), which can substantially reduce the cost of supervised mitigation (see \cref{sec:mitigation-supervision}).
RS-awareness is complementary to mitigation strategies: when all realistic mitigation strategies are exhausted, some RSs may still be present. 
In that case, RS-aware models will improve uncertainty quantification.
Unfortunately, in practice, NeSy predictors are \textit{not} RS-aware: they tend to be very confident even about poorly grounded concepts \citep{marconato2024bears,van2025neurosymbolic,bortolotti2026concise}.

What sets awareness apart from mere mitigation is that it is possible to make NeSy predictors RS-aware \textit{in a principled manner}, but \textit{without} concept supervision, as we will see.
As mentioned in \cref{sec:mitigation-entropy}, a simple heuristic is to \textit{\textbf{maximize the entropy}} of the predictive concept distribution. 
Is this always the best we can do? 
To answer this question, we will need to define RS-awareness. 

\subsubsection{RS-awareness as mixtures of deterministic RSs}

RS awareness can be understood as mixing over the set of deterministic RSs that the NeSy predictor is affected by \citep{van2025independenceRS}. 
We denote this set as $\Vset{\calA}_{\text{RS}} \subseteq \Vset{\calA}$, and it can be seen as the set of ``irreducible'' deterministic RSs that are still present after exhausting the feasible mitigation strategies. 
Formally, we consider RS-aware NeSy predictors as distributions that ``mix'' over the concept remapping distributions (\cref{eq:definition-of-alpha}) induced by each irreducible deterministic RS $\va\in \Vset{\calA}_{\text{RS}}$. 
We achieve this using the convex combination of deterministic RSs from \cref{eq:convex-combination-rs}, where each weight $\lambda_{\va}>0$. 
The convex combination ensures each RS contributes to the final concept remapping distribution. 
The resulting distribution is itself a (nondeterministic) RS \textcolor{teal}{\citep[Lemma 1]{marconato2024bears}}, and hence has perfect label accuracy.

\subsubsection{Building RS-aware NeSy predictors}
How do we achieve such nondeterministic RSs in NeSy predictors? 
This usually requires that the space of learnable concept extractors $\calF$ (\cref{sec:non-identifiability-of-concepts}) is powerful enough to model dependencies between different concepts \citep{van2025independenceRS}. 
However, many implementations of NeSy predictors use concept extractors that assume (when conditioned on the input $\vx$) the different concepts $C_1, \ldots, C_k$ are independent, \ie $p(C_1, \ldots, C_k \mid \vx) = \prod_{i=1}^k p(C_i \mid \vx)$~\citep{van2024independence}. 
NeSy predictors exploit the independence assumption for efficient inference~\citep{darwiche2002knowledge,van2022anesi,choi2025ctsketch,de2023differentiable}. 
However, it also greatly restricts the space of learnable concept extractors $\calF$.

\begin{tcolorbox}[
    colback=gray!2!white,
    colframe=gray,
    arc=0pt,
    outer arc=0pt
]
\begin{example}[Example in~\citep{van2025independenceRS}]
    We consider the \MNISTXOR task where the NeSy predictor should predict the output of the XOR function applied to two \MNIST digit. The XOR function returns 1 if the two input digits are different, and 0 otherwise. 
    \Eg $\vx=\MOne\MZero$ gives $y=1$, while $\vx=\MOne\MOne$ gives $y=0$. 
    This problem contains the shortcut $\MOne\mapsto 0, \MZero\mapsto 1$ (under disentanglement, \cref{sec:mitigation-disentanglement}).
    Given input $\vx=\MOne\MZero$, an RS-aware NeSy predictor should assign 0.5 probability to both the ground-truth concept vector $(1, 0)$ and the RS concept vector $(0, 1)$.
    If we would try to model this distribution with the independence assumption, we find that $p(C_1=1\mid \vx) = p(C_2=1\mid \vx) = 0.5$. 
    But then $p(C_1=1, C_2=1\mid \vx) = p(C_1=1\mid \vx) \cdot p(C_2=1\mid \vx) = 0.25$, \ie it assigns probability to a concept vector that does not give the correct output label. 
    Hence, the resulting distribution is \emph{not} an RS, and might make incorrect label predictions. 
\end{example}
\end{tcolorbox}

In general, \citet{van2025independenceRS} proves that the independence assumption prevents models from being aware of RSs for most inference layers. 
Motivated by these results, we discuss two recent methods that go beyond the independence assumption to build RS-aware NeSy predictors.

\begin{figure}[!t]
    \centering
    \includegraphics[width=0.75\linewidth]{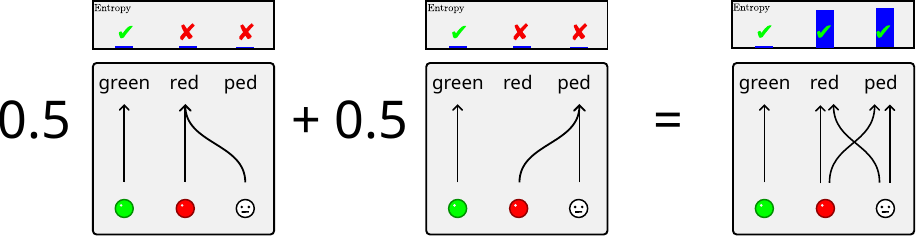}
    \caption{\textbf{Intuition behind \bears.} \bears~\citep{marconato2024bears} constructs ensembles of concept predictors that employ different sets of concepts to solve the task. Concepts that need to be learned correctly have to match across models, while those that cannot will have different groundings. As a result, averaging their predictions increases uncertainty only where disagreement exists. } %
    \label{fig:bears-intuition}
\end{figure}

\subsubsection{RS-Awareness via Ensembles}  \label{sec:bears}  \bears~\citep{marconato2024bears} is a model-agnostic technique for making NeSy predictors RS-aware.
\bears replaces the concept extractor with an \textit{ensemble} of concept extractors, each using the independence assumption. 
\bears constructs this ensemble specifically such that each member captures a different deterministic RS in $\Vset{\calA}_{\text{RS}}$.
This construction has ties with Bayesian deep learning techniques~\citep{wang2020survey, daxberger2021laplace, osawa2019practical, lakshminarayanan2017simple, gal2016dropout} but it is also backed by a separate theoretical setup \citep{marconato2024bears}.
For instance, in \BOIA (\cref{ex:boia-rs}) one extractor might conflate pedestrians with red lights, while another might do the opposite, as in \cref{fig:bears-intuition} (left).
\bears computes the predictive concept distribution by averaging over all extractors in the ensemble.  Intuitively, this works for two reasons.
First, if all extractors have high label accuracy, their average will as well (\cref{eq:convex-combination-rs}).
Second, the concept entropy of the ensemble hinges on the semantics associated by the different extractors to the concepts themselves.
Then, the ensemble has high entropy for concepts that are affected by RSs---like pedestrian and red light---as different extractors each learn different semantics.  
For unaffected concepts---like green light---the extractors agree on their semantics and the ensemble has low entropy.
This intuition is illustrated in \cref{fig:bears-intuition} (right).

In practice, \bears extends deep ensembles~\citep{lakshminarayanan2017simple} with a loss term that encourages ensemble members to model different RSs, and an entropy term (reminiscent of \cref{sec:mitigation-entropy}) that spreads concept predictions.
Compared to other Bayesian deep learning methods, such as MC dropout~\citep{gal2016dropout} and the Laplace approximation~\citep{daxberger2021laplace}, deep ensembles allow learning highly multi-modal distributions, making them an excellent choice for learning multiple, diverse RSs. %
This is supported by experiments, which show that \bears significantly improves RS-awareness of NeSy architectures on different tasks and that it does so better than other Bayesian alternatives \citep{marconato2024bears}.
One downside of \bears is that it requires learning an ensemble of concept extractors.  Fortunately, small ensembles of $5$ to $10$ members were shown to be sufficient in practice.

\subsubsection{RS-Awareness via Diffusion}  \label{sec:nesy-diffusion} 
Motivated by going beyond the independence assumption, \nesydm \citep{van2025neurosymbolic} uses expressive discrete diffusion models as concept extractors. 
Unlike \bears, it trains a single concept extractor $p(\vC \mid \vx)$ via masked diffusion models (MDMs)~\citep{austin2021structured,sahoo2024simple}. 
MDMs start with a concept vector that consists entirely of masks. 
Masks essentially indicate we do not yet know what the value of the concept should be. 
Then, MDMs are tasked with iteratively unmasking the various concepts. 
The main motivation for using MDMs to go beyond independence is scalability: 
each unmasking step can be seen as a concept extractor with the independence assumption. 
Hence, \nesydm can directly reuse existing inference methods that exploit the independence assumption. 

\nesydm consists of three loss components. 
The first loss aims to unmask partially masked concept, the second unmasks partially masked labels, and the final loss encourages high entropy among the concepts. 
One benefit compared to \bears is that \nesydm trains only a single model, eliminating the need to choose the ensemble size. 
Experimentally, \nesydm also matches or improves on calibration and out-of-distribution concept accuracy compared to \bears.

\subsection{Awareness Helps Mitigation}
\label{sec:mitigation-interaction}

To support mitigation strategies, specifically concept supervision (discussed in~\cref{sec:mitigation-supervision}), one can leverage human input to request targeted annotations. This idea is well established in the field of Interactive Machine Learning~\citep{ware2001interactive, fails2003interactive, amerishi2014power, teso2019explanatory}. 
In the context of RSs, the most relevant method explored so far is Active Learning~\citep{settles2012active}, which we briefly describe below.

\paragraph{Active Learning} 
Rather than relying on costly full supervision, Active Learning~\citep{settles2012active} allows a model to interactively query a human (or oracle) for labels on the most informative or uncertain examples. The core idea is to focus annotation efforts where they are most impactful, thereby improving model performance while minimizing human effort. As demonstrated in~\citep{marconato2024bears}, uncertainty estimates over concepts -- provided by \bears (see~\cref{sec:awareness}) or \texttt{NeSyDM}~\citep{van2025neurosymbolic} (see~\cref{sec:nesy-diffusion}) -- can be used to identify and query the most uncertain concepts, effectively addressing RSs with fewer annotations on concepts. This targeted strategy outperforms traditional active learning methods, such as querying concept uncertainty in \PNSP, where the uncertainty signal is often uninformative~\citep{marconato2024bears}.

\subsection{How to choose between mitigations?}
\label{sec:choosing-mitigations}

\changed{
Below, we provide a short guide for chosing among mitigation strategies.
Recall that RSs can be avoided simply by obtaining concept annotations, with the caveat that many may be required in order to avoid all possible RSs.
If doing so is not feasible, the choice of mitigation strategy largely depends on the application.  We summarize common practical issues in \cref{tab:choice-which-mitigations}.
}
\begin{enumerate}
    \item 
    \changed{Not all RSs affect downstream performance equally. This typically occurs in tasks where distinguishing between certain concepts is unnecessary for the final prediction (\eg the type of \texttt{obstacle} on the road). In contrast, other (high-stakes) concepts must be clearly distinguished to ensure correct generalization to new tasks (\eg \texttt{pedestrian} vs \texttt{red\_light}). 
    To enforce such distinctions, one approach is to directly supervise the targeted concepts (\cref{sec:mitigation-supervision}). This can be combined with prototypical representations \citep{andolfi2025right} to reduce annotation costs (\cref{sec:mitigation-prototype}). Alternatively, designing opportune multi-task objectives can promote learning separate concept representations (\cref{sec:mitigation-multitask}).}

    \item 
    \changed{In practice, NeSy models are often prone to learning RSs that collapse concepts into a limited set of combinations, thus impairing generalization and post-hoc identification of the learned concepts \citep{bortolotti2025shortcuts}. 
    Augmenting the learning objective with entropy regularization (\cref{sec:mitigation-entropy}) encourages more diverse concept mappings, mitigating collapse. More resource-intensive alternatives include contrastive learning (\cref{sec:mitigation-contrastive}) and reconstruction penalties (\cref{sec:mitigation-reconstruction}). 
    Preventing collapse also enables post-hoc renaming procedures \citep{fokkema2025sample}, where learned concepts are reassigned meaningful labels. This allows both concept representations and inference layers to be aligned with ground truth \citep{bortolotti2025shortcuts}.
    }

    \item 
    \changed{
    Collapsed representations often lead to overconfident predictions. This issue can be addressed using \bears (\cref{sec:bears}) and \nesydm (\cref{sec:nesy-diffusion}), which explicitly identify concepts affected by RSs. However, these approaches can be computationally expensive. 
    More lightweight alternatives with weaker guarantees include label smoothing (\cref{sec:mitigation-smoothing}) and entropy regularization (\cref{sec:mitigation-entropy}). Label smoothing operates at the output level, while entropy-based methods act at the concept level.
    }

    \item 
    \changed{
    RSs can also create entanglement between concepts. The most effective solution is to use model backbones that enforce disentangled concept representations by design (\cref{sec:mitigation-disentanglement}). 
    When such architectures are not available, weak supervision through multiple views \citep{locatello2020weakly} can help promote disentanglement, although formal guarantees may not be met.}

    \item 
    \changed{
    Eliminating deterministic RSs (i.e., zeroing their count) does not necessarily remove all RSs in practice. This is particularly relevant for NeSy predictors that do not satisfy the extremality condition (\cref{assu:monotonic}). 
    In such cases, it is preferable to adopt inference layers that meet these requirements, such as \PNSP, or to rely on suitable fuzzy logic fragments (\cref{sec:nesy-predictors}). Additionally, weak supervision through \ABL (\cref{sec:mitigation-abduction}) can further help reduce concept-level risk.
    }
\end{enumerate}

\begin{table}[!h]
    
    \caption{
    A list of issues that might arise due to RSs. We outline possible mitigations and the rationale behind each of them.
    }

    \centering
    \scriptsize
    \begin{tabular}{p{0.175\textwidth}p{0.01\textwidth}p{0.3\textwidth}p{0.01\textwidth}p{0.4\textwidth}}
        \toprule
         \textbf{Issue}
         & &
         \textbf{Mitigations}
         & &
         \textbf{Rationale} \\
         \midrule
         Lack of generalization in new tasks
         & &
         Concept Supervision + Prototypes \rechanged{(\textit{if  concept embeddings are clusterizable})}, Multi-task learning (\textit{if available})
         & & 
         Distinguish between high-stakes concepts supervising on them, if available integrate new tasks that can be relevant in novel scenarios. \rechanged{Prototype-based grounding is effective when concept embeddings are clusterizable.} \\
         \midrule
         Collapse onto few concept combinations
         & &
         Entropy, Contrastive Learning, Reconstruction
         & &
         Use additional regularizations which penalize collapse, then apply a post-hoc renaming phase. \\
         \midrule 
         Overconfident predictions 
         & &
         \bears and \nesydm (\textit{higher computational cost}), Smoothing and Entropy (\textit{lesser computational cost})
         & &
         Model uncertainty on concepts affected by RSs, which informs about uncertain predictions. \\
         \midrule
         Entangled concept predictions
         & &
         Disentanglement (\textit{if possible}), 
         Contrastive Learning with Concept Supervision
         & &
         Adopt architectural biases of the network that can help separate between concept predicitons. Contrast different concept values by supervising them. 
         \\
         \midrule
         Non-deterministic concepts
         & &
         Inference layers that satisfy \cref{assu:monotonic}, Weak-supervision with \ABL
         & &
         Avoid methods which might not guarantee full RSs removal when zeroing the count. 
         Use methods which allow for stronger concept risk reduction. \\
         \bottomrule
         
    \end{tabular}
    \label{tab:choice-which-mitigations}
\end{table}

\newpage %
\section{Extensions and Open Problems}
\label{sec:extensions}

In this section, we discuss RSs in contexts beyond those studied in this paper, and outline potential extensions and open research problems.

\subsection{Reasoning Shortcuts in NeSy AI Beyond Predictors}
\label{sec:extension-nesy-ai}

RSs stem from the intrinsic ambiguity of logic inference when the prior knowledge $\BK$ admits inferring the ground-truth label from unintended concepts.
This suggests that RSs arise \textit{\textbf{regardless of how the reasoning step is implemented}}, that is, also for NeSy architectures beyond the four predictors we considered here.\footnote{Models like \DPL and \LTN are full-fledged first-order programming languages, \ie they can handle first-order logic specifications and work with a variable number of logic entities and concepts.  Given that RSs affect them in the propositional case and that FOL is a strict generalization of propositional logic, we can safely assume that RSs exist also in the FOL case.}
\Eg RSs plausibly affect extensions of these models, such as \citep{huang2021scallop, winters2022deepstochlog, desmet2023deepseaproblog, badreddine2023logltn}, as well as predictors that combine probabilistic \textit{and} fuzzy semantics, like {\tt NeuPSL} \citep{pryor2022neupsl},  models that embed symbols into continuous vector spaces, such as neural theorem provers with fuzzy semantics~\citep{rocktaschel2016learning} and probabilistic semantics~\citep{maene2023softunification}, models that implement hard constraints with fuzzy logic, such as~\citep{giunchiglia2020coherent, giunchiglia2024ccn}, and diffusion-based models \citep{van2025neurosymbolic}.
Moreover, RSs have also been identified in {\tt SatNet} \citep{wang2019satnet, simard1991tangent}, an architecture that performs perception and reasoning \textit{entirely in the embedding space}.
This leads us to postulate that---barring implicit mitigation due to the use of reconstruction penalties, see \cref{sec:mitigation-reconstruction}---RSs might also affect \textit{generative} NeSy models, such as \citep{misino2022vael, ferber2024genco}. %

While it is likely that RSs affect a broad class of NeSy architectures, more work is needed to extend the existing theoretical frameworks and mitigation strategies to these models.

\subsection{Reasoning Shortcuts in Concept-based Models}
\label{sec:extension-cbms}

So far, we only considered NeSy predictors where the prior knowledge $\BK$, and therefore the inference layer, is supplied externally and fixed.
Recently, RSs have also been studied in two related families of models whose \textit{\textbf{inference layer is learned}} \citep{bortolotti2025shortcuts}: \textit{i}) NeSy modular predictors that acquire the knowledge from data~\citep{daniele2023deep, tang2023perception, shindo2023thinking, shindo2024learning, wust2024pix2code, debot2024interpretable}, and \textit{ii}) Concept-based Models (CBMs), which pair a concept extractor with an interpretable (typically linear) inference layer \citep{koh2020concept, zarlenga2022concept, marconato2022glancenets, schwalbe2022concept, poeta2023concept, pugnana2026deferring, knab2026whats}.
It was shown that, without the aid of concept supervision, these models suffer from \textit{\textbf{joint reasoning shortcuts}} (JRSs): failure modes involving extracting wrong concepts and/or wrong logical assignments that lead to correct label predictions (see~\cref{ex:mnistsump} for an example).

\begin{figure}[t]
    \centering
    \includegraphics[width=0.9\linewidth]{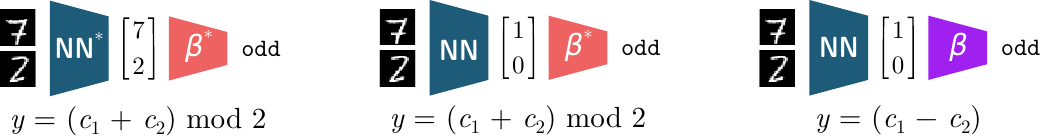}
    \caption{\textbf{Example of Joint Reasoning Shortcuts.}  The task is to predict whether the sum of two \MNIST digits is odd, using training data that cover all (even, even), (odd, odd), and (odd, even) pairs. Red denotes fixed knowledge, purple denotes learned knowledge.  In all the presented cases, the models achieve optimal performance.\textit{Left:} ground-truth concepts with the correct inference layer. \textit{Middle:} NeSy model with prior knowledge may still learn shortcuts by assigning unintended semantics to concepts. \textit{Right:} CBMs, both concepts and the inference layer are misaligned, resulting in \emph{joint reasoning shortcuts}. \textbf{Note:} In both of the previous cases, the OOD setting (even, odd) produces a consistent prediction, whereas the joint reasoning shortcut yields $-1$ as the predicted value. %
}
    \label{fig:jrs}
\end{figure}

\begin{tcolorbox}[
    colback=gray!2!white,
    colframe=gray,
    arc=0pt,
    outer arc=0pt
]
\begin{example}[\MNISTSumParity]
    To provide an example, imagine a variant of \MNISTAdd where the goal is to predict whether the sum of two digits is odd or even, that is, $Y = (C_1 + C_2) \! \mod 2$. Suppose that the training set includes only examples for which $C_2$ is \textit{odd}. 
    A possible JRS amounts to mapping
    all even digits to $0$, \ie $\{\MZero, \MTwo, \MFour, \MSix, \MEight\} \mapsto 0$, and
    all odd digits to $1$, \ie $\{\MOne, \MThree, \MFive, \MSeven, \MNine \} \mapsto 1$. 
    Consistently, the learned knowledge is the difference among the predicted digits $\hat Y = \hat C_1 - \hat C_2$, as it would return the very same labels as using $Y= (C_1 + C_2) \! \mod 2$. In contrast to regular RSs, here both the concepts and the learned knowledge are ``wrong'', with the result that in the OOD setting where $C_2$ includes even digits, the learned model can predict $\hat Y = -1$ (see ~\cref{fig:jrs} for a visual representation).
    \label{ex:mnistsump}
\end{example}
\end{tcolorbox}

All RSs can also be viewed as JRSs, but the opposite is not true: the set of learnable JRSs can vastly outnumber that of regular RSs
\citep{bortolotti2025shortcuts}. For example, permuting concept values does not alter the concept semantics and the learned rules, leading to equivalent solutions upon renaming the extracted concepts.\footnote{Specifically, since concepts are \textit{anonymous} during training, they can be easily aligned to ground-truth concepts at test time by leveraging annotations. This, however, is not possible if a joint-RS is learned by the model \citep{bortolotti2025shortcuts}.
} 
Counting RSs can be adapted to joint-RSs, and an equivalent result to \cref{thm:identifiability-rss} also holds when $\BK$ is learned, showing that dealing with deterministic joint-RSs is, in many cases, sufficient to rule out all joint-RSs.
However, the benefits of many mitigation strategies for RSs discussed in \cref{sec:mitigations} do not transfer to joint-RSs. 
Supervised strategies, like concept supervision and multi-task learning, offer some improvement but are insufficient to address all joint-RSs.
In particular, concept supervision can lead to high concept accuracy, yet it does not necessarily recover the correct knowledge. It is unknown whether more powerful strategies for combating JRSs exist, and a theoretical characterization of JRSs from a statistical learning perspective is currently missing. 

In the context of RSs, investigating the scenario where learned concept vectors ($\vc \in \calC$) and ground-truth concept vectors ($\vg \in \calG$) belong to different spaces, \ie $\calC \neq \calG$, is open and so far unexplored.
Considering two different spaces $\calC$ and $\calG$ results in the NeSy predictor inference layer $\beta$ being structurally different from $\beta^*$, because of the different mapping domain.
This setting naturally draws a connection between RSs and JRSs, where in the latter a concept-based model may learn to use fewer concepts in the bottleneck and fine-tune a $\beta \neq \beta^*$ that still matches label predictions. In this sense, because NeSy predictors $(f, \beta)$ have $\beta \neq \beta^*$, one has to check whether learned concepts and the inference layer possess (a variation of) the intended semantics \citep[Definition 3.3]{bortolotti2025shortcuts}. 
We expect that, in this scenario, RSs can likely increase due to a mismatch between the NeSy predictor's and the ground-truth knowledge.

\subsection{Foundation and Large Language Models}
\label{sec:extension-foundation-models}

An interesting direction for future research is to investigate the connection between RSs and foundation models \citep{bommasani2021foundation, radford2021learning}.
Several recent approaches already leverage information \textit{\textbf{elicited from foundation models}} to prime NeSy~\citep{stammer2024neural, helff2025activation, wust2026synthesizing} and concept-based architectures~\citep{oikarinen2022label, yang2023language, rao2024discover, srivastava2024vlgcbm, yuksekgonul2023post}, \eg using pre-trained representations or weak concept annotations obtained by prompting visual-language models. %
Clearly, the benefit is that foundation models are trained on vast amounts of data, which allows covering a large support of (potentially new) inputs, and are multimodal---allowing the use of textual concept descriptions to estimate the concepts in the data.    %
Moreover, we expect that pre-training implicitly implements some of the RS mitigation strategies we outlined: concept supervision (\cref{sec:mitigation-supervision}) and multi-task learning (\cref{sec:mitigation-multitask}). 
In fact, it is quite likely that some elementary concepts, like the color or shape of objects, need to be separated in model representations to allow them to match a huge amount of image-caption pairs (\eg from the Web Image Text dataset \citep{srnivasan2021witdataset}). 
For these reasons, it is plausible that, depending on the domain, foundation models may produce useful, high-quality concepts.

Empirical evidence, however, calls for caution.  %
Recent findings~\citep{debole2025concept, wust2025bongard} in the context of VLM-augmented CBMs indicate that foundation models can fail to identify some basic visual concepts. %
In addition, they may fall short due to issues such as hallucinations~\citep{huang2024survey}, shortcuts~\citep{yuan2024llms}, and limited logical~\citep{calanzone2025logically} and conceptual consistency~\citep{sahu2022unpacking}.
In a direction closely related to RSs, \citet{stein2025neuro} have shown that language models are affected by \textit{\textbf{symbol hallucinations}}, a phenomenon where models prompted to generate concepts and executable programs to solve a task end up hallucinating poorly grounded symbols that still manage to achieve high accuracy.  Intuitively, symbol hallucinations can be thought of as the analog of RSs for prompting, \rechanged{although they originate not because of ambiguities related to a fixed prior knowledge in NeSy. This, in turn, complicate their characterization.} 
An important direction is to determine whether foundation models themselves are affected by (joint) RSs proper, but we expect that the current theory will need to be extended to also treat models that cannot be described as NeSy predictors. 
Attempts to provide identifiability guarantees for concept learning in model representations (among which those of foundation models) have recently surfaced \citep{zheng2025nonparametric, liu2025predict,rajendran2024causal}, but it remains to be understood how this relates to shortcuts when these guarantees are not given. 

Reasoning shortcuts have also been studied in language modelling, though with a different scope from the one of this paper. 
In question answering, multi-hop reasoning is the model's ability to answer a question by combining information from multiple pieces of evidence rather than relying on a single fact~\citep{yang2018hotpotqa}. As an example of two-hop reasoning, we would expect that to answer the question \textit{``In which city of the Orange Free State was the author of The Lord of the Rings born?''}, the model first has to identify J.~R.~R.~Tolkien as the author of The Lord of the Rings, and second has to determine that Tolkien was born in Bloemfontein, which is in the Orange Free State.
\citet{jiang2019avoiding} found that when models are trained on certain multi-hop QA datasets, such as HotpotQA~\citep{yang2018hotpotqa}, they often exploit shortcuts that allow them to return correct answers by matching keywords or surface patterns in the question and context, \emph{without performing the intended multi-hop reasoning steps}. 
In this context, the authors refer to it as a \textit{reasoning shortcut} for the model.
For instance, in the example above, an incorrect reasoning step consists of predicting Bloemfontein just because of the presence of ``the Orange Free State''.
Although these unintended solutions differ from the reasoning shortcuts discussed in our manuscript, they can be related to joint reasoning shortcuts, whereby correct concept assignments are given but the learned knowledge fails to capture the ground truth. Making this link concrete is open and is an important research direction.

\subsection{Reasoning Shortcuts in Reinforcement Learning}
\label{sec:extention-rl}

NeSy reinforcement learning (RL)~\citep{NeSyRL, veronese2025learning} is a new field in which logical knowledge is integrated into the reinforcement learning workflow for various purposes, including guiding exploration~\citep{veriied_exploration}, ensuring safe trajectories~\citep{yang2023safe}, improving explainability~\citep{Baugh2025, deane_nesy25}, performance~\citep{NRM, detect_understand_act}, or generalization to new environments~\citep{garnelo_2016, badreddine_2019}.
Most of the works consider environments with symbolic observations, where grounding symbols or concepts is not required. 
A smaller set tackles more realistic environments by discretizing or clustering the neural features extracted from high-dimensional inputs~\citep{umili_antonioni, garnelo_2016, dreamerv2}. In such cases, the extracted concepts lack a predefined meaning, potentially limiting the human interpretability of the agent’s behavior.
Only a few works have investigated embedding prior symbolic knowledge into RL policy learning ~\citep{badreddine_2019, NRM, symDQN_2025}.
While reasoning shortcuts are still relatively underexplored in NeSy RL, first studies in this direction have already proven valuable for:
(i) certifying the quality of learned concepts and its reusability in single-task vs. multitask settings ~\citep{NRM}, and (ii) improving the explainability of NeSy-RL agents when knowledge is learned rather than injected~\citep{deepDFA_images}.

\begin{figure}[t]
    \centering
    \begin{tabular}{cc|ccc}
        \includegraphics[height=9em]{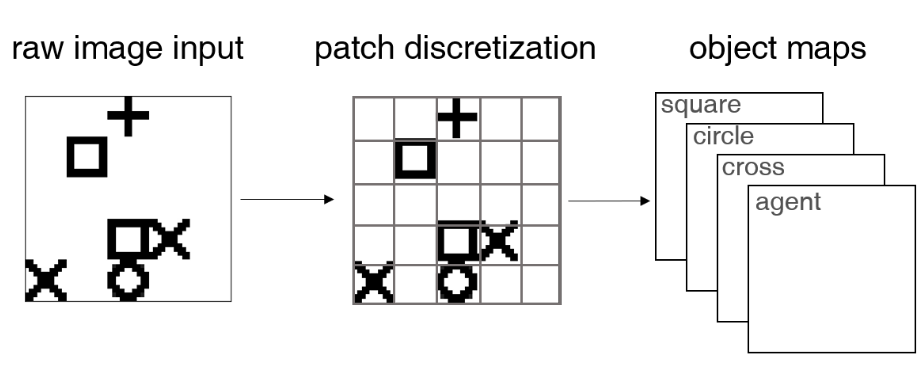}
         & 
         &
         &
         \includegraphics[height=9em]{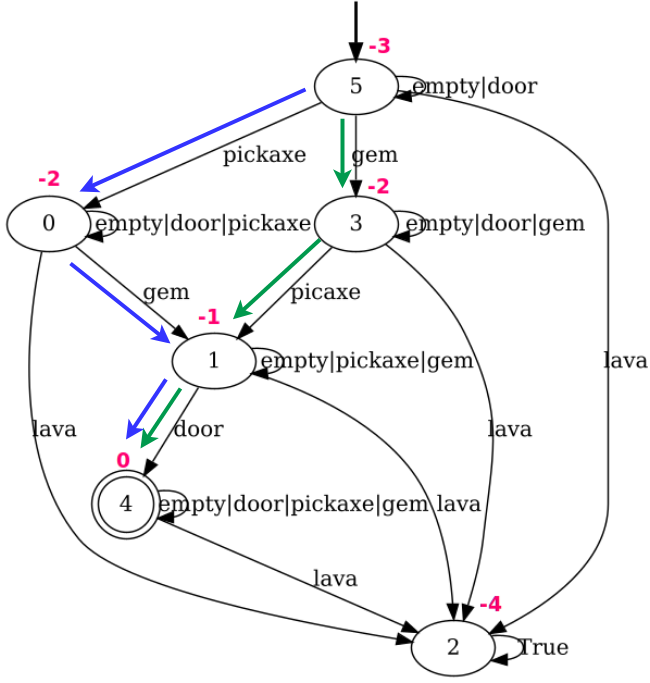}
         &
        \includegraphics[height=9em]{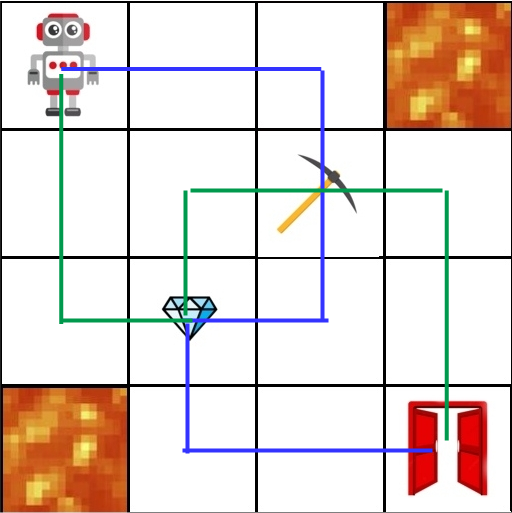}
    \end{tabular}

    \caption{(Left) Environment introduced in ~\citep{garnelo_2016} and mainly used in ~\citep{badreddine_2019,symDQN_2025}. 
    The agent (+ sign) has to maximize the sum of rewards by collecting shapes, each shape gives a different immediate reward. 
    In both works, they use a sort of soft knowledge injection that is biased for this specific environment: they divide the input image into 25 patches (one per grid cell), and the shape recognizer classifies each patch singularly.
    (Right) Environment introduced in ~\citep{NRM} for temporally extended visual tasks. The agent has to maximize the reward on the automaton. It changes automaton state by reaching specific items in the environment. On the left side, the automaton for the task "reach the pickaxe and the gem before the door, while always avoiding the "\texttt{lava}", is displayed.%
    On the rightmost side, a visualization of the environment is portrayed. In green and blue two optimal trajectories causing the indistinguishability of symbols "\texttt{pickaxe}" and "\texttt{gem}" due to the structure of the task.}
    \label{fig:nesy_env}
\end{figure}

\citet{badreddine_2019} and \citet{symDQN_2025} focus on solving the RL environment introduced in \citep{garnelo_2016} and shown in \cref{fig:nesy_env} (Left). They both employ a NeSy shape recognizer exploiting knowledge encoded in \LTN. 
In these works, RSs do not appear because they rely on mitigation strategies specifically tailored to that particular environment, such as the use of pretrained shape detectors in \citep{badreddine_2019}, the additional knowledge of the environment model in \citep{symDQN_2025}, and patch discretization in both works.

A more recent work in NeSy-RL~\citep{NRM} explicitly investigates reasoning shortcuts through neural reward machines (NRMs).
There, the RL agent learns to solve temporally extended tasks expressed in linear temporal logic over finite traces (LTLf) \citep{LTLf} formulas. \cref{fig:nesy_env} (Right) illustrates an example of such a task.
NRMs encode temporal knowledge through a probabilistic relaxation of the deterministic finite automaton (DFA) equivalent to LTLf task. %
Here, the only bias %
is that rewards are shaped according to the automaton: reward is maximum at the automaton’s accepting states (reachable only by satisfying trajectories) and reduced elsewhere. This is a minimal bias that allows applying the framework to diverse environments and tasks without any change in the framework.
In the context of the example of \cref{fig:nesy_env}, the green and blue trajectories correspond to equivalent solutions for the agent, which may induce the agent to swap the concepts \texttt{``gem''} and \texttt{``pickaxe''}, resulting in an RS.
Moreover, in ~\citep{NRM} an efficient algorithm to identify all reasoning shortcuts that depend on the temporal logical specification is introduced, allowing to count RSs over three order of magnitude faster than a brute-force baseline.\footnote{While the brute-force baseline constructs all possible $\alpha$'s in $\Vset{\calA}$, the algorithm introduced in \citep{NRM} exploits common properties of RL tasks, such as terminal states and self-loop transitions in automata, to reduce the space of the search.} Applying this algorithm to the LTLf formulas commonly used in non-Markovian RL showed an extremely high number of RSs.
Interestingly, in single-task RL, these RSs do not harm learning performance, since they involve only concept renamings and do not induce any concept collapse. Thus, all task-relevant concepts remain distinguished, but they may be swapped. However, this swap can harm the transfer of learned concepts to new multitask settings~\citep{kuo_2021}.
A follow-up work~\citep{FLNRM} extends NRMs to the cases where the task logical knowledge is not provided a priori: both the knowledge and the concepts are learned directly from exploration data~\citep{deepDFA_images}. 
It investigates an algorithm to jointly minimize the automaton states and concepts learned via RS detection, yielding knowledge that is significantly more compact and explainable.

Future research can likely explore RSs in multitask generalization. In this context, prior knowledge is increasingly used to automatically generate new environments and tasks~\citep{openended_1, openended_2}, making it timely to exploit (parts of) this knowledge during training to boost the RL agent's performance. 
Another important open question concerns the role of temporal dependencies in NeSy RL. 
Unlike static tasks (\eg classification or regression), RL data consists of exploration traces, where each observation depends on the previous ones, raising the challenge of how best to exploit this interdependence for better RS detection~\citep{van2025independenceRS}.

\subsection{Imbalanced Learning}
\label{sec:extension-partial-learning}

\citet{tsamoura2025imbalances} identified another phenomenon that is inherent to NeSy: learning imbalances, which are major differences in errors that occur when classifying instances of different classes (aka \emph{class-specific risks}) and can lead to poor learning of concept extractors. 
Existing research on machine learning \citep{LA, Label-Distribution-Aware-Margin-Loss,solar}  has studied imbalances, but this was only under the prism of \textit{long-tailed} (aka \textit{imbalanced}) data%
, \changed{where instances from different classes} occur with very different frequencies, \citep{Learning-from-Imbalanced-Data,devil-in-tails}.  
However, these results cannot fully characterize learning imbalances in all NeSy tasks. This is because \textit{the background knowledge may cause learning imbalances \changed{in the labels} even when the \changed{concept} data is uniformly distributed\changed{, or vice versa}}. We illustrate the above phenomenon by the following example:

\begin{tcolorbox}[
    colback=gray!2!white,
    colframe=gray,
    arc=0pt,
    outer arc=0pt
]
\begin{example}[Learning Imbalances in \MNISTMax~\citep{tsamoura2025imbalances}]
    \label{ex:mnist-max-imbalances}
    Let us consider a variant of the classical \MNISTAdd scenario, referred to as \MNISTMax, in which instead of predicting the sum, we predict the maximum of two MNIST digits\changed{, \eg $(\MFive\MNine) \mapsto 9$}.
    We will adopt the notation in \citet{wang2023learning} and denote each training sample by $(\vx_1, \vx_2, \vy)$, where $\vy$ corresponds to the target maximum. 
    We distinguish the following two cases: 
    \begin{enumerate}[leftmargin=1.5em]
        \item
        The marginal of $\vY$ is uniform.
        \changed{In other words,} the number of training samples %
        \changed{for which the maximum is 0 ($\vy = 0$)} is equal to the number of training samples where \changed{the maximum} $\vy$ %
        \changed{takes} any of the remaining nine digits. %
        \changed{In this settings,} we expect that learning the digit zero \changed{(concept $\vc = 0$)} %
        \changed{to} be easier than learning the digit nine \changed{(concept $\vc = 9$)}, \ie the classifier will incur fewer errors in recognizing the digit zero than %
        \changed{when recognizing} any other digit. This is because training samples \changed{in which the maximum is zero, \ie $(\vx_1,\vx_2, 0)$, can only occur when both digits are zero and the label $\vy = 0$ appears uniformly in the training set. As a result, such samples force the NeSy model to learn the concept zero, since it is the only one satisfying the prior knowledge.}%
        
        \item
        The marginal of the %
        \changed{ground-truth concepts $\vG$}, %
        \changed{\ie the values of \MNIST digits,} %
        is uniform. %
        \changed{However, when we create training samples} $(\vx_1, \vx_2, \vy)$ %
        by sampling %
        \changed{two} \MNIST 
        \changed{digits independently, the resulting label $\vy = \max(\vx_1, \vx_2)$} is not uniform. %
        \changed{As a result,} learning the digit nine %
        \changed{(concept $\vc = 9$) is} easier than learning the digit zero \changed{(concept $\vc = 0$) as $\vy = 9$ appears much more frequently than $\vy = 0$,} %
        \changed{even though samples of the form $(\vx_1, \vx_2, 0)$ provide a direct and unambiguous way to learn the concept zero.}
    \end{enumerate}
\end{example}
\end{tcolorbox}

\citet{tsamoura2025imbalances} theoretically characterized this phenomenon in NeSy, extending previous results \citep{cour2011}. The characterization involved solving a non-linear program whose solutions are upper bounds to class-specific risks \changed{of the concepts}. In addition, the authors proposed algorithms to mitigate imbalances during training and testing time. The former algorithm assigns pseudolabels to training data based on a novel formulation of NeSy as an integer linear program \citep{ILP-cookbook}. The latter algorithm constrains the model's predictions on test data using robust semi-constrained optimal transport \citep{robust-ot}. Both mitigation algorithms rely on a common idea in long-tailed learning: enforcing the class priors to the predictions of a concept extractor. The intuition is that the concept extractor will tend to predict the labels that appear more often in the training data. Hence, enforcing the priors gives more importance to the minority concepts at training time and encourages the model to predict minority concepts at testing time.

\subsection{Additional Open problems}
\label{sec:open-problems}

Here, we list a few more directions which were not treated in previous subsections:

\begin{enumerate}[leftmargin=1.5em]

    \item
    Although several mitigation strategies have been proposed (some of them quite powerful, such as relying on concept supervision~\citep{koh2020concept}), the most effective ones are costly in terms of human annotations. 
    {Related to this, multitask learning (\cref{sec:mitigation-multitask}) can be employed to obtain such annotations by designing combinations of NeSy tasks that promote the removal of reasoning shortcuts. However, it remains unclear how to construct such tasks effectively. A naive approach might involve combining learned programs~\citep{muggleton1994inductive, de2008probabilistic, wust2024pix2code} with \texttt{rs-count}~\citep{bortolotti2024benchmark} to evaluate how the newly generated programs affect the number of reasoning shortcuts. This, however, requires access to ground-truth concepts, which is typically infeasible in real-world applications, and can also be highly inefficient in terms of computational cost. Developing efficient methods for artificially constructing such NeSy tasks therefore remains an open research problem.}
    On the other hand, inexpensive methods such as entropy regularization~\citep{manhaeve2021neural} do not offer {guarantees regarding their effectiveness in reducing the reasoning shortcuts count}. There is thus a need for {mitigation strategies that are both theoretically grounded and cost-efficient, requiring minimal human annotation effort while still providing provable guarantees of effectiveness}. %

    \item \changed{The mitigation strategies discussed in \cref{sec:mitigations} focus on avoiding \textit{deterministic} RSs~\cref{eq:causes} and as such provide explicit reduction of their number, see \cref{eq:count-rss}.  Implicitly, doing so equates to assuming that all RSs are equally impactful.
    However, this is not necessarily true: mistaking red lights for pedestrians is often worse than confusing red lights for stop signs.
    An alternative is to design mitigation strategies that take this into consideration. By prioritizing more costly RSs, this could reduce the need for expensive concept supervision and, more generally, reduce the down-stream cost of RSs.
    One option is to design mitigation strategies by building on the statistical learning perspective we discussed in \cref{sec:rss-as-statistical-learning}, which natively enables modelling the  risk~\cref{eq:rss-risk-expected} of RSs in a cost-sensitive fashion.
    How to properly do so remains an open question.}

    \item Another open direction concerns the development of stronger concept detection pipelines, which could improve the overall identification process~\citep{locatello2020object}. 
    For example, in computer vision, RSs are closely related to visual grounding~\citep{xiao2024towards}, where much work has focused, for instance, on object detection; however, a concrete link between object detection and RSs has yet to be established. 
    Moreover, incorporating temporal dependencies (\eg in reinforcement learning or in textual data) may further improve early detection and mitigation of RSs \citep{lippe2022citris,lippe2023biscuit}.

    \item %
    To better advance the study of RSs, it is also extremely useful to  extend the available experimental resources. 
    While \texttt{rsbench}~\citep{bortolotti2024benchmark} represents the first benchmark suite {for a standardized evaluation of the impact of RSs in NeSy predictors and the efficacy of mitigation strategies, the field lacks benchmarks to examine how RSs affect model performance in highly challenging settings}.
    {An unexplored direction is to evaluate RSs on large, real-world NeSy datasets such as \texttt{ROAD-R}~\citep{giunchiglia2023road}}.
    Furthermore, 
    to facilitate deploying NeSy predictors, 
    it is beneficial to develop modular implementations of NeSy models that are both easy to use and freely available. Recent initiatives, like \texttt{ULLER}~\citep{vankrieken2024uller} and  \texttt{DeepLog}~\citep{derkinderen2025deeplogneurosymbolicmachine, manhaeve2026deeplog}, are promising steps towards these implementations. %

\end{enumerate}

\section{Related Work}
\label{sec:related-work}

\paragraph{Symbol Grounding.}  
The symbol grounding problem concerns how symbols acquire meaning independently of human interpretation~\citep{harnard1990grounding}. 
Reasoning shortcuts correspond to a faulty attribution of meaning to symbols (\textit{w.r.t.}\ their human-interpretable semantics), which can hinder interpretability and out-of-distribution generalization.
~\citet{harnard1990grounding} pioneered that symbol grounding should chiefly depend on the \textit{identification} of the symbols. 
He argues that \textit{discrimination} is the ability of systems to assign inputs representing different concepts with different symbols (\eg images of horses and cats are associated to different symbols), but that this is not sufficient: \emph{identification} further requires that these symbols are exactly the concepts we have in mind for them (\eg for an image of a horse, the symbol must be exactly the concept of horse). 
Identification is precisely what underlies the grounding of the symbols. 
Similarly, an RS (\cref{sec:rss-theory}) is when we \emph{discriminate} for the purposes of the task, but fail to completely \emph{identify} concepts. 

The symbol grounding problem has inspired challenging datasets for visual reasoning like CLEVR \citep{johnson2017clevr} and the visual abstractions benchmark~\citep{hsu2025makes}.
The former requires scene understanding based on multiple 3D objects annotated with textual descriptions, whereas the latter focuses on grounding highly varied instantiations of the same concept (\eg mazes). 
They argue that \textit{schemas}, collections of connected lower-level concepts, together form the basis for grounding higher-level abstract concepts like a maze. 

Symbol grounding is also closely related to binding \citep{greff2020binding} in neural networks, that is, how models derive a compositional understanding of the world in terms of symbolic entities, like objects or concepts. 
Recently, researchers asked to what degree foundation models require --- or even have already achieved --- symbol grounding \citep{hsu2023s,jiang2024multimodal}. 
Some authors argue that the problem remains for models that are trained purely on language~\citep{pavlick2023symbols,levine2025language}, which are symbolic in nature. 
Others argue that forming meaningful concepts requires embodied experiences~\citep{barsalou2020challenges,Barsalou_1999}, which current models fundamentally lack \citep{dove2024symbol,levine2025language}. 
\citet{steels2008symbol} argues the focus should be on AI systems that autonomously create and maintain the grounding of concepts, and that supervised machine learning is insufficient as the semantics still comes from humans. 
Part of the focus has shifted to what forms of grounding ---if linguistic, multimodal, or embodied--- are necessary for genuine conceptual understanding \citep{barsalou2020challenges, gubelmann2024pragmatic} and to whether this should be more framed specifically for distributed representations \citep{mollo2023vector}.
While NeSy predictors go beyond supervised machine learning and reduce reliance on direct concept supervision, they still require human-provided knowledge to be effective and currently do not support autonomous concept acquisition.

\paragraph{Inductive Logic Programming} Inductive logic programming (ILP)~\citep{de2008probabilistic, muggleton1994inductive} is a branch of machine learning that induces logic programs from examples and background knowledge. Hypotheses are typically represented as sets of first-order Horn clauses, and the goal is to find a symbolic program that entails all positive examples while excluding negative ones. One important technique in ILP is \emph{predicate invention}, where the system autonomously generates new predicates to capture latent relationships in the data~\citep{stahl1993predicate}. By creating intermediate concepts, predicate invention allows ILP models to abstract complex relational patterns, simplify rule sets, and improve generalization. For example, in an animal classification task, an ILP system might invent a \texttt{mammalian(X)} predicate to encode the mammal concept. However, the grounding of invented predicates can misalign with the one intended by the human, which makes predicate invention an interesting research direction to study connections with reasoning shortcuts and human interpretability.

A recurring challenge in ILP, both with and without predicate invention, is the presence of \emph{symmetries} in the hypothesis space, which often leads to redundant search~\citep{tarzariol2022lifting}. For instance, two clauses such as \texttt{parent(X,Y) :- mother(X,Y).} and \texttt{parent(A,B) :- mother(A,B).} are equivalent as they differ only in variable naming. Similarly, two clauses that differ only in the order of their literals are identical: for example, \texttt{sibling(X,Y) :- parent(Z,X), parent(Z,Y).} is equivalent to \texttt{sibling(X,Y) :- parent(Z,Y), parent(Z,X).} Exploring the hypothesis space in ILP without addressing symmetries can lead to the generation of many redundant, equivalent clauses. To mitigate this issue, various ILP systems employ \emph{symmetry-breaking} techniques that constrain the search space and avoid unnecessary symmetries. For example, FOIL~\citep{quinlan1990learning} addressed this by enforcing a consistent left-to-right ordering of literals, while Progol~\citep{muggleton1995inverse} and Aleph~\citep{srinivasan2001aleph} used mode declarations, which restrict the possible forms that clauses can take, and Metagol~\citep{cropper2016metagol}, instead, provided metarules, which define general clause schemata that implicitly prevent the generation of equivalent variants. 

In ILP, symmetries arise from structural redundancies in the hypothesis space, such as clauses that are equivalent up to variable renaming, literal order, or clause permutation. These symmetries may not permit identification: there can be multiple programs that fit the data equally well, yet differ in how they capture the intended abstraction. In particular, recurring relational patterns in the optimal hypothesis space can create multiple equivalent representations, each providing the same explanatory power.
An analogous problem also occurs in satisfiability (SAT), where variables or clauses that play equivalent roles introduce redundancy in the search space~\citep{sakallah2021symmetry}. Just as in ILP, SAT solvers employ symmetry-breaking techniques to eliminate redundant solutions~\citep{anders2024satsuma, ulyantsev2016symmetry, bogaerts2022certified}.
Studying the connection between symmetries in the search space and reasoning shortcuts can be of interest, as it may offer insights into both improving learning efficiency and understanding how learned hypotheses relate to intended abstractions.

\paragraph{Shortcuts in Machine Learning.} \label{par:shortcuts} %

Machine learning models often rely on shortcuts, spurious correlations between input features and target labels that are not causally related, leading to poor performance, particularly in OOD settings~\citep{geirhos2020shortcut, teso2023leveraging, ye2024spurious, steinmann2024navigating}.
Prior work related to concepts has examined how spurious correlations affect both CBMs and NeSy models~\citep{margeloiu2021concept, mahinpei2021promises, havasi2022addressing, raman2023concept, stammer2021right}. These studies primarily focus on understanding and correcting concept quality in the presence of spurious confounders in the data. To address this problem, strategies such as Explanatory Interactive Learning~\citep{teso2019explanatory, teso2023leveraging, bontempelli2021toward, teso2021interactive, bontempelli2023concept} have been developed, aiming to identify such spurious correlation and allow humans to correct them during the model debugging process. However, this line of work does not consider the high-level abstraction required to connect symbols to input features, a challenge central to the symbol grounding problem.
In contrast, this paper focuses on reasoning shortcuts~\citep{li2023learning, marconato2023not, wang2023learning, NRM, marconato2024bears, bortolotti2024benchmark, delong2024mars, bortolotti2025shortcuts}, which arise not from low-level feature correlations but from the misuse of concepts during decision-making, as their semantics differ from the intended ones.

\paragraph{Identifiability} 
Identifiability has been largely studied in representation learning, especially within independent component analysis (ICA) \citep{Hyvarinen2009ICA, hyvarinen2024identifiability} and causal representation learning (CRL) \citep{scholkopf2021toward}.
In ICA, the goal is to determine the independent underlying components from observed data. The central question of identifiability is: under what conditions can the same independent components (up to tolerable ambiguities) be consistently recovered \citep{buchholz2022function, gresele2023learning}? 
If components are not identifiable, re-estimating them at different times may yield conflicting results, meaning there is no unique way to determine the same independent factors. 
The same lack of uniqueness also appears in reasoning shortcuts.
Seminal works have explored conditions under which non-linear independent components can be uniquely recovered, including using auxiliary variables \citep{hyvarinen2019nonlinear, khemakhem2020variational}, sparsity \citep{lachapelle2022disentanglement}, anchor points \citep{moran2021identifiable}, weak supervision \citep{locatello2020weakly}, multiple views \citep{gresele2020incomplete}, and constraints to the model family \citep{gresele2021independent}.
These ideas have also been extended to CRL, where identifiability is necessary not only to discover the underlying components but also to reconstruct the causal relationships between them \citep{von2021self, lippe2022citris, buchholz2023learning, ahuja2023interventional, lippe2023biscuit, von2024nonparametric, fokkema2025sample}.
We foresee that many techniques and learning setups that provide guarantees for identification can also help in mitigating RSs, but so far the connection to ICA and CRL has not been investigated. 

Identifiability is also studied in supervised classification \citep{reizinger2025cross}, multi-task learning \citep{lachapelle2023synergies, fumero2023leveraging}, contrastive learning \citep{von2021self, zimmermann2021contrastive}, and language models \citep{roeder2021linear, marconato2025all}. 
In these settings, models may achieve close predictive performance while learning dissimilar internal representations \citep{nielsen2024challenges, nielsen2025doesclosenessdistributionimply}. This situation is analogous to reasoning shortcuts, where correct outputs are obtained through misaligned or unintended representations.
Revealing the precise connection between these models and RSs may help extend beyond the current paradigm of NeSy predictors.

\section{Conclusion}
\label{sec:discussion}

Symbol grounding aims to link humans' and machines' high-level abstractions of the world. %
Well-grounded symbols, or concepts, contribute to generalization and interpretability. 
NeSy AI holds the great promise of facilitating this alignment in hybrid systems that are human interpretable and understandable.
Our work characterizes the past, present and future challenges of symbol grounding in NeSy AI systems through the lens of RSs.
 The presence of RSs prevents identification of the correct concepts, and thus possibly compromises the benefits of NeSy AI.
Our theoretical analysis in \cref{sec:rss-theory} indicates that, in general, all NeSy predictors we consider---and likely most others---are naturally susceptible to RSs.  That is, concepts are non-identifiable in general.
However, the theory also characterizes the mechanisms behind the occurrence of RSs, and therefore paves the way for designing a new generation of NeSy AI systems that are more reliable and easier to align with humans' expectations through effective mitigation strategies.
More work is, however, needed to articulate what conditions guarantee provable retrieval of the ground-truth concepts without requiring dense concept annotations. We expect that this can be done in a similar spirit and with similar strategies as those used to guarantee identifiability of representations in independent component analysis and causal representation learning (see \cref{sec:related-work}).

\section*{Acknowledgements}

We thank Samy Badreddine for useful input while writing this paper.
Part of this work was initiated before Efthymia Tsamoura joined Huawei Labs.
Funded by the European Union. The views and opinions expressed are however those of the author(s) only and do not necessarily reflect those of the European Union, the European Health and Digital Executive Agency (HaDEA) or the European Research Executive Agency. Neither the European Union nor the granting authority can be held responsible for them. Grant Agreement no. 101120763 - TANGO.
Emile van Krieken is funded by the NWO AiNed project ``Human-Centric AI Agents with Common Sense'' under contract number NGF.1607.22.044. 
Paolo Morettin is supported by the MSCA project “Probabilistic Formal Verification for Provably Trustworthy AI - PFV-4-PTAI” under GA no. 101110960.
Elena Umili is supported by the PNRR MUR project PE0000013-FAIR.
AV is supported by the ``UNREAL: Unified Reasoning Layer for Trustworthy ML'' project (EP/Y023838/1) selected by the ERC and funded by UKRI EPSRC.
Stefano Teso was partially supported by the Flemish Government under the
Flemish research foundation (FWO) project ``Neurosymbolic AI for Constraint Learning'' (Project G047124N).

\bibliographystyle{ACM-Reference-Format}
\bibliography{references, explanatory-supervision}

\appendix

\end{document}